\documentclass[runningheads]{llncs}
\usepackage{graphicx}
\newcommand{\bz}{\mathbf{z}}

\newcommand{\bv}{\mathbf{v}}
\newcommand{\be}{\mathbf{e}}
\newcommand{\bx}{\mathbf{x}}
\newcommand{\by}{\mathbf{y}}

\newcommand{\bh}{\mathbf{h}}
\newcommand{\bn}{\mathbf{n}}

\newcommand{\bTheta}{\mathbf{\Theta}}

\newcommand{\bmu}{\boldsymbol{\mu}}

\newcommand{\bg}{\mathbf{g}}

\usepackage{amsfonts}
\usepackage{booktabs}
\usepackage{subcaption}
\usepackage{amsmath}
\usepackage[figuresright]{rotating}
\usepackage{xcolor}
\usepackage[all,cmtip]{xy}
\usepackage{algorithmic}
\usepackage{algorithm}
\usepackage{hyperref}
\usepackage{tabularx}
\usepackage{enumerate}
\usepackage{circuitikz}
\usepackage{makecell}
\usepackage{multirow}
\usepackage{xcolor}
\usepackage{array}
\usepackage{breakcites}

\graphicspath{ {Images/} }

\begin{document}
\title{{Bridging Data and Physics: A Graph Neural Network–Based Hybrid Twin Framework}}

\titlerunning{A GNN–Based Hybrid Twin Framework}

\author{M.~Gorpinich\inst{1,2},
B.~Moya\inst{2},
S.~Rodriguez\inst{2},
F.~Meraghni\inst{2},
Y.~Jaafra\inst{1},
A.~Briot\inst{1},
M.~Henner\inst{1},
R.~Leon\inst{1},
F.~Chinesta\inst{2,3} 
}

\authorrunning{M.~Gorpinich et al.}

\institute{Valeo, Creteil, France  \and PIMM Lab. ENSAM Institute of Technology. Paris, France\and CNRS@CREATE LTD. Singapore}

\maketitle

\begin{abstract}
{Simulating complex unsteady physical phenomena relies on detailed mathematical models, simulated for instance by using the Finite Element Method (FEM). However, these models often exhibit discrepancies from the reality due to unmodeled effects or simplifying assumptions. We refer to this gap as the ignorance model. While purely data-driven approaches attempt to learn full system behavior from measurements, they require large amounts of high-quality data across the entire spatial and temporal domain. In many real-world scenarios, such information is unavailable as physical quantities are measurable only at limited points, making full data-driven modeling unreliable. To overcome this limitation, a hybrid twin approach is required. Instead of simulating phenomena from scratch, here we focus on the modeling of the ignorance component. Since physics-based models approximate the overall behavior of the phenomena, the remaining ignorance is typically lower in complexity than the full physical response, therefore, it can be learned with significantly fewer data. A key difficulty, however, is that spatial measurements are sparse, and even more, obtaining data measuring the same phenomenon for different spatial configurations --- system’s domain, external load, etc. --- is challenging in practice.  Our contribution is to overcome this limitation by using Graph Neural Networks (GNNs) to represent the ignorance model. GNNs inherently handle geometric structures, making them compatible with any approximation technique, for instance, FEM. They learn the spatial pattern of the missing physics even when the number of measurement locations is limited. This allows us to enrich the physics-based model with data-driven corrections without requiring dense spatial, temporal and parametric data. To assess the performance of the proposed methodology, we evaluate this GNN-based hybrid twin on nonlinear heat transfer problems across different meshes, geometries, and load positions. The results indicate that the GNN captures model discrepancies and generalizes corrections across varying domain geometries and mesh discretizations. This leads to improvements in simulation accuracy and interpretability, while also reducing data requirements.}

\end{abstract}

\section{Introduction}

High-fidelity simulations of complex systems, such as those that exhibit a strongly nonlinear behaviour --- turbulent fluid flows, structural deformations under large loads, or architectured materials, to name a few --- are crucial for design, safety assessment, and performance optimization.
    
The representation of their behaviour leads to the resolution of complex and expensive models for their approximation. However, in most cases, the considered models does not accurately represent reality. This is due to physical biases or simplifications in the models, or to inherent errors when discretizing the corresponding Partial Differential Equations (PDEs) using, for instance, the Finite Elements Method \cite{taylor2013finite}. 

Therefore, in these scenarios, a gap between the prediction and the reality is perceived \cite{oberkampf2010verification}. In order to address this limitation, an extension of classical Digital twins was proposed, called hybrid twins \cite{chinesta2020virtual}. The main idea consists in learning this ignorance by a data-driven model, usually constructed using model-order reduction \cite{sancarlos2021learning,rodriguez2023hybridscarse,rodriguez2023hybrid} as the sparse Proper Generalized Decomposition or deep learning techniques \cite{moya2022digital,moya50post,ghnatios2024hybrid,holt2024automatically}. In the field of hybrid modeling, the main challenges lie in the ability to generalize, adapt, and integrate seamlessly with the underlying physics-based model \cite{willard2020integrating}. The main advantage of combining a physics-based with a data-based model for the ignorance is reducing the need for experimental data, as long as the physical model represents reality up to some degree. This is an appealing property at the industrial level, where experimental data is limited due to its deployment costs.
    
Several deep learning techniques have been used to build this ignorance model. Among them, one can mention the use of LSTMs (Long short-term memory) or ResNet (Residual Network) architectures \cite{ghnatios2023learning}, which have a major impact in learning dynamical systems. This is possible because, although measurements can only be taken at a few spatial locations, temporal sampling can be performed without limitation, allowing the acquisition of rich and detailed time-series data. However, constructing an ignorance model capable of providing corrections across the entire spatial domain remains challenging, because the sparsity of spatial measurements makes it difficult for these models to infer accurate corrections in unobserved regions of the domain.

An attractive alternative to overcome this spatial correction limitation is the use of Graph Neural (GNNs) \cite{sanchez2020learning}. The main advantage of these networks consists in their capability of learning inherent spatial geometric relations of the dataset, being able to produce predictions for different geometric domains.

By operating directly on graph structures, GNNs can encode the topology of a physical system, capturing both local interactions and long-range dependencies. 
Another important benefit is its flexibility with respect to domain shape and geometry.  
GNNs are also capable of processing data defined on meshes with non-uniform node density across the domain. This is particularly advantageous in simulations where mesh refinement is applied selectively in regions of high gradients or physical interest \cite{pfaff2020learning}, allowing the model to preserve accuracy without enforcing uniform discretization.
They explicitly incorporate relational information through edges and graph connectivity \cite{battaglia2018relational}. This structured representation allows the network to better model conservation laws, interaction forces, and other physically meaningful constraints.
This technique has proven effective for learning full models of the prediction of complex systems, given known initial conditions \cite{pfaff2020learning}.

GNNs coupled with FEM proved to be effective for predicting rigid body dynamics \cite{bhattoo2022learning}, elasticity simulations \cite{firouzi2025graph, zhou2024graph}, solids \cite{zhai2025stress}, fluid dynamics \cite{toshev2023learning, dupuy2023modeling, quattromini2023active, quattromini2025mean}, and other practical applications. Nevertheless, these models have some drawbacks when the dataset is not expressive enough for the phenomenon to be learnt \cite{han2022predicting}. However, in the field of hybrid twinning, since the main objective corresponds to the construction of the ignorance model, the use of GNNs for the best adapted ignorance spatial construction seems appealing. Therefore, the novelty of this work is the  implementation of a hybrid twin using GNNs to achieve a spatial correction, given by the ignorance model, when considering sparse data points for its construction.

Importantly, in this work the adoption of a GNN architecture is motivated by the irregular, graph-structured nature of our simulation mesh data, where node number and connectivity vary across samples. Baseline architectures like Multi-Layer Perceptrons (MLPs) and Convolutional Neural Networks (CNNs) are poorly suited for this task, as they require data to be forcibly flattened to a fixed size or interpolated onto regular grids \cite{zaheer2017deep, bronstein2017geometric, kipf2016semi}. These ad hoc transformations introduce discretization artifacts, disrupt structural information, and do not take into account permutation invariance. Because our primary goal is to evaluate generalization across different graph structures \cite{xu2018powerful, battaglia2018relational}, MLPs and CNNs do not naturally support this problem setting.

To illustrate this idea, we consider synthetic data to present the methodology. The reference problem is the enrichment of a linear heat transfer simulation in order to predict a nonlinear counterpart response, unknown. These simulations are obtained using the FEM. Although the authors are aware about possible differences between real world problems and the use cases under consideration, the current work focuses on the proof of concept and the use cases with a more realistic problem setting will be a subject for future works.

Here, the ignorance model is learnt using GNNs with a graph representation derived from the FEM discretization. GNNs offer the ability to capture a generalized representation of corrections across different scenarios (domain geometry or load position variations) and discretization levels, while requiring less data due to the imposition of geometric biases, thereby enhancing the efficiency of learning the ignorance model.
    
We show proof of the efficiency of the hybrid framework under different domains and conditions, showing  stability, error reduction, and generalization capability compared to baseline, fully data-driven GNNs.

The paper is organised as follows. Section \ref{rel_works} exposes the state-of-the-art and makes an overview of the common approaches applied to the problem under consideration. Section \ref{method} describes the model architecture details. Following, Section \ref{experiment_setup} presents the configurations of the test use cases. Section \ref{results} analyses the  results, and finally Section \ref{disc}  concludes the paper,  explores possible perspectives of the research and summarizes the findings.
    
\section{Related works} \label{rel_works}

Physics-based models have been traditionally used for the approximation of physical phenomena of interest. These ensure both interpretability and expressivity, as they have been developed in accordance with physical laws \cite{quarteroni2025combining}. However, they are sometimes limited by poor modeling and mathematical assumptions that do not capture the whole complexity of the problem \cite{chinesta2020virtual}. This issue can arise from a wrongly chosen time or space discretization \cite{strikwerda2004finite}, or possibly due to missing or incorrect information in the mathematical approximation \cite{moya2022digital}, such as, for instance, assuming an incorrect non-linearity term \cite{gonzalez2019learning} or simplified physics contributions. Hence, representations of the physics problem deviate from reality, compromising their use in real applications \cite{oberkampf2010verification}. These gaps in the representation are identified through measurements performed on the real system. In the era of AI, one may consider adopting a purely data-driven approach to learn directly from data approximations that fit the observed reality. Nonetheless, such approaches lack interpretability and generalization, making them unreliable during testing phases \cite{montans2019data}.

Hybrid modeling establishes a bridge between data and knowledge to work in an informed and consistent framework for model discovery \cite{chinesta2020virtual,kurz2022hybrid}. Hybrid modeling is a traditional and extensive field of research in engineering, which has converged with Theory-guided Data Science (TGDS) \cite{karpatne2017theory}. Both establish how knowledge, and particularly physics models and known epistemic laws, can enrich model learning.

In this context, hybrid modeling can be applied from different perspectives depending on how information and data are combined. The most explored topic is currently the development of physics-guided machine learning, where physics can be part of the algorithm architectures (physics-embedded) or enforced during the learning process (physics-informed) \cite{wu2024physics}. One of the most well-known and applied techniques in the field is the so-called Physics-Informed Neural Networks (PINNs) \cite{raissi2017physics,hennigh2021nvidia,haghighat2021sciann}, which have gained attention as a method to incorporate physical laws directly into the learning process, notably PDEs. Studies demonstrated the effectiveness of PINNs in solving a wide range of PDEs in which we highlight contributions in fluid dynamics and thermal problems based on the focus of this work \cite{hennigh2021nvidia,raissi2017physics,wang2021learning,eivazi2024physics,xu2023physics,amini2022physics}. However, learning an incorrect hypothesis, i.e. PDEs that incorrectly reflect the physics of the phenomenon, will guide to the learning of incorrect approximations \cite{zou2024correcting}. More general solutions rely on physics-embedded learning approaches. These include structure-preserving methods that enforce conservation or symplectic properties \cite{greydanus2019hamiltonian,tong2021symplectic}, as well as models incorporating epistemic biases derived from thermodynamic principles \cite{tierz2025feasibility,idrissi2024multiscale}. Still, the accuracy of these approaches can be jeopardized by the quality and quantity of data, the complexity of the underlying system, and the way physical biases are imposed. Moreover, these models remain grey-boxes, where generalization, and particularly interpretability, are often limited \cite{cuomo2022scientific}.

However, a particularly recent focus on neural network architectures that adapt to irregular and unstructured domains is worth our attention. Graph Neural Networks (GNNs) \cite{kipf2016semi,hamilton2017inductive,zhang2020deep,hoang2023graph,velivckovic2017graph} have shown remarkable potential in various domains, particularly in handling complex relational data \cite{pfaff2020learning,shao2023pignn,barwey2023multiscale,feng2024graph}. This is due to the imposition of so-called geometric biases, which enable the model to achieve higher generalization during learning \cite{bronstein2021geometric}. By leveraging graph structures, they are capable of capturing and learning the physics of a system based on the pairwise node interactions, which can also be combined with inductive biases, if known \cite{thangamuthu2022unravelling,bermejo2025meshgraphnets}.

Another approach to hybrid modeling is the one present in hybrid twins, the methodology followed in this work, which consists in learning an ignorance model that approximates solely the error detected between the physics model and the reality \cite{champaney2022engineering,liang2025harnessing}. As a result, only a few data measurements will be sufficient to characterize the model of ignorance.

In this case, several examples can be found of physics-based models corrected by an ignorance model built upon sparse-PGD \cite{sancarlos2020rom,moya2022digital,rodriguez2023hybrid} and more general machine learning approaches \cite{bonavita2020machine,gonzalez2019learning,holt2024automatically}. The work in \cite{sun2022physinet} combines a physics-based and an ignorance models built upon a fully connected neural network that learns the discrepancy, improving the predictions along the system's life-cycle. Also, the work of Daby-Seesaram et al. proposes a correction model between an approximation of the physics model using the Proper Generalized Decomposition and the ground truth \cite{daby2025finite}. Similarly, authors in \cite{ghnatios2024hybrid} apply this correction framework to the design of magnetic bearings. Also, the work of \cite{yun2022novel} employs this framework for disaster management, particularly in the prediction of wildfires. Another field with a strong impact of hybrid twins is structural health monitoring, where the ignorance appears when there is some damage and failure present in the system \cite{luo2023hybrid,di2022data,rodriguez2023hybrid,di2024damage}.

The integration of the FEM principles with GNNs \cite{mitusch2021hybrid,li2023finite,eslamlou2025hybrid} has opened new avenues for enhancing the simulation workflow. While these works do not use a hybrid twin paradigm, combining both data-driven (GNN) and physics-driven (FEM) approaches in a single framework allows for accurate and scalable simulations. The authors in \cite{li2023finite} demonstrate improved accuracy in simulating fluid flows by integrating physical constraints based on the FEM into the loss function and twice-message aggregation. In the work presented in \cite{eslamlou2025hybrid}, the authors utilize a hybrid FEM and GNN approach to create an efficient and accurate tool for structural damage detection. Their hybrid modeling framework has several modules. The first module employs a GNN trained on modal data from FEM simulations to estimate the location and severity of structural damage. In the second module, a conformal prediction technique quantifies  uncertainty in the GNN's predictions. These uncertainty-aware predictions initialize a warm-started FEM model updating workflow.

In contrast, in this work, we propose the development of a hybrid twin, using an existing physical model as an input to learn the gap between it and the experimental data that includes nonlinearities present in the thermodynamic phenomena under study.

\section{Method}\label{method}

Graph Neural Networks (GNNs) have gained prominence as an effective means to model and learn from graph-structured data. A graph $ G = (V, E) $ consists of nodes $V = \{1, \dots, n\}$, where $n$ is the number of nodes, connected by edges $ E $, where $E \subseteq V \times V$ and $m$ is the number of edges. Nodes represent spatial discretization points in a domain, and edges correspond to the relationships or connections between these points.

GNNs are well-suited for physical domains because they can inherently consider the spatial and topological properties of the data. A typical GNN architecture includes three primary components: the encoder, the processor, and the decoder. The encoder transforms the original node and edge features into a latent, called \textit{hidden}, representation for the graph-based learning that combines structural information (neighbors, connectivity, or role) with contextual information (abstract features that summarize how a node relates to its surroundings, enabling the model to generalize beyond the initial representation). The processor is the main step of the GNN, since it performs message passing across the graph structure to update node and edge representations based on information from their neighbors. Multiple layers of this message-passing process are incorporated to capture both local and global relational patterns. After this processing step, the decoder translates the learned hidden features into the desired output, which is problem and application dependent.

\subsection{Encoder}

In each node of the graph we have the following features:

\begin{equation}\label{eq:node_feat}
    \bv_{i} = [\bz_i;\bn_i]
\end{equation}

\noindent
and in every edge:

\begin{equation}\label{eq:edge_feat}
    \be_{ij} = [\bx_{i} - \bx_{j}; ||\bx_{i} - \bx_{j}||_2],
\end{equation}

\noindent
where $\bv_{i}$ is the $i$-th node features in a graph, $\bz_i$ represents physical quantities that characterize the $i$-th node (temperature, velocity, etc.), $\bn_i$ is the one-hot encoded vector that represents different groups of nodes in a graph (interior, border, heat source, boundary condition, etc.),  $\be_{ij}$ are the features of an edge between $i$-th and $j$-th nodes, and $\bx_{i}$ are the coordinates of $i$-th node. 

In general, boundary nodes are treated in the same way as interior nodes. The only difference is that boundary nodes may be subject to boundary conditions or other local physical constraints. In our case, a Dirichlet boundary condition with a fixed temperature is applied. For these nodes, the vector $\mathbf{n}_i$ is adjusted accordingly. In addition, the network output at these nodes is set to zero to satisfy the imposed boundary condition.

The encoder transforms node and edge input features \eqref{eq:node_feat} and \eqref{eq:edge_feat} into a hidden representation separately. Mathematically, this can be expressed as:

\begin{equation*}
    \bh_{v_i}^{(0)} = \varepsilon_{V}(\bv_i; \bTheta_{\varepsilon_{V}}),
\quad
    \bh_{e_{ij}}^{(0)} = \varepsilon_{E}(\be_{ij}; \bTheta_{\varepsilon_{E}}),
\end{equation*}

\noindent
where $\bh_{v_i}^{(0)}$, $\bh_{e_{ij}}^{(0)}$ are the initial hidden states of node $v_i$ and edge $e_{ij}$ respectively; $\varepsilon_{V}$, $\varepsilon_{E}$ are the corresponding encoders represented by fully-connected neural networks, $\bTheta_{\varepsilon_{V}}$, $\bTheta_{\varepsilon_{E}}$ represent the learnable parameters of the encoders respectively.

It is also worth noting that the current architecture is defined by the dimensions of node and edge features, and not the number of edges and vertices in the graph.

\subsection{Processor}

The processor iteratively updates the node hidden features by aggregating information from neighboring nodes. This component captures the interactions between different parts of the domain. At each message passing layer $k$, the hidden states of nodes and edges are updated sequentially. First, the edge embeddings are updated:

\begin{equation*}
\bh_{e_{ij}}^{(k+1)} = \bh_{e_{ij}}^{(k)} + f_E^{(k)} \left( \bh_{e_{ij}}^{(k)}, \bh_{v_i}^{(k)}, \bh_{v_j}^{(k)}; \bTheta_{f_E^{(k)}} \right),
\end{equation*}

\noindent
and then node embeddings are updated using:

\begin{equation*}
\bh_{v_i}^{(k+1)} = \bh_{v_i}^{(k)} + f_V^{(k)} \left(\bh_{v_i}^{(k)},\bg\left(\left\{ \bh_{e_{ij}}^{(k+1)} \mid j \in \mathcal{N}(i)\right\} \right) ; \bTheta_{f_V^{(k)}}    \right)
\end{equation*}

\noindent
where $\mathcal{N}(i)$ is the set of neighboring nodes of $v_i$, $\bh_{v_i}^{(k)}$ is a hidden representation of the features of a node $v_i$ on a message passing layer $k$, $\bh_{e_{ij}}^{(k)}$ is a hidden representation of the features of the edge $e_{ij}$, $\bg$ is  permutation-invariant aggregation function; $f_V^{(k)}$ and $f_E^{(k)}$ are the processing functions for node and edge updates on the layer $k$ with parameters $\bTheta_{f_V^{(k)}}$ and $\bTheta_{f_E^{(k)}}$ respectively.

It is worth noting that the equations used to represent the node and edge updates are phrased in residual form \cite{he2016deep}. The aggregation function $\bg$ that combines the messages from all neighboring nodes corresponds to the aggregation function in this case. The processing functions $f_V^{(k)}$, $f_E^{(k)}$ are represented by fully-connected neural networks. They are used to update the nodes and edges hidden states based on their current state and the aggregated information.

\subsection{Decoder}

The decoder converts the final hidden states back to the original feature space to make a prediction. The decoder function $\delta_V$ with parameters $\bTheta_{\delta_V}$ maps the processed node features to the desired output features, which are problem dependent:

\begin{equation*}
    \hat\by_i = \delta_V(\bh_{v_i}^{(K)}; \bTheta_{\delta_V}),
\end{equation*}

\noindent
where $\hat\by_i$ is the predicted output for node $v_i$, $\bTheta_{\delta_V}$ represents the learnable parameters of the decoder, $K$ is a total number of message-passing layers.

Here, since we are following the hybrid twin paradigm \cite{chinesta2020virtual}, we propose to learn the gap between the FEM approximation and observed ground truth:

\begin{equation}\label{eq:gap_model}
    f^\text{GT}(t; \bmu) = f^\text{FEM}(t; \bmu) + f^\Delta(f^\text{FEM}(t; \bmu)).
\end{equation}    

\noindent
where $f^\text{FEM}(t; \bmu)$ is a FEM approximation that depends on parameters $\bmu$ and a timestep $t$, $f^\text{GT}(t; \bmu)$ is a corresponding ground truth (GT),  $f^\Delta(f^\text{FEM}(t; \bmu))$ is a deviation model represented by a GNN that takes FEM approximation as input.

According to the equation \eqref{eq:gap_model} for the thermal problem the input node features include:
\begin{equation*}
    \bz_{i} = T^\text{FEM}_i.
\end{equation*}
\noindent
where $T_i$ is a temperature in the node $i$.

From the same equation, we can conclude that:
\begin{equation*}
    y_i(t) = T^\text{GT}_i(t)-T^\text{FEM}_i(t),
\end{equation*}
\noindent
and respectively:

\begin{equation*}
    \hat y_i(t) = \hat T_i(t)-T^\text{FEM}_i(t),
\end{equation*}

\noindent
where $\hat T_i(t)$ is a corrected temperature in the node $i$.

The ground truth simulation frame is obtained by adding the input of the model to the output.

\begin{figure}[ht!]
    \centering
    \includegraphics[width=\textwidth]{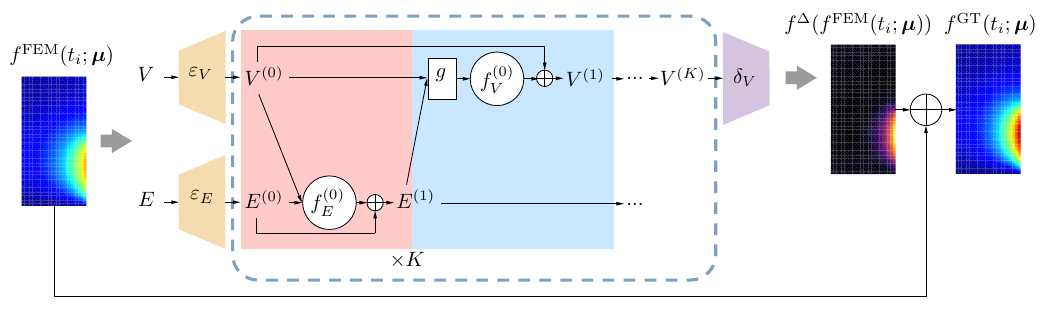}

\caption{Scheme of the proposed approach of the heat transfer model. The model takes a linear simulation frame as an input. Node and edge features are passed through the encoder that projects them into the hidden space. The resulting features are then passed through $K$ message passing layers that update node features in the hidden space. The decoder outputs the difference of temperature between linear and nonlinear simulations. The output is added to the input to obtain the nonlinear simulation frame.}
\label{fig:method_schema}
\end{figure}

Figure \ref{fig:method_schema} demonstrates the detailed description of the model architecture.

\section{Numerical setup}\label{experiment_setup}

We test the performance of the proposed approach on a series of use cases. For this purpose we are using the data of the linear and the corresponding nonlinear heat transfer simulation in a metal plate (iron). The task of the GNN is to be able to enrich the linear simulation and to learn the gap between it and the nonlinear one. 

The linear approximation of a physics-based model is governed by the following isotropic heat equation:

\begin{align}
    & \rho c_P \frac{\partial T}{\partial t} - k \nabla^2 T = \dot{q}_V\\
    & T(x = 0) = 298\;\text{K}\\
    & T(t = 0) = 298\;\text{K}.
\end{align}    

\noindent
where $\rho$ is the density of the metal plate, $c_P$ is the specific heat capacity at constant pressure, $k$ is the thermal conductivity of the material (at $298\;\text{K}$ for linear case), $\dot{q}_V$ is the volumetric heat source. The temperature is given in Kelvin ($\text{K}$).

On the other hand, the nonlinear ground truth is computed by considering a nonlinear conductivity coefficient:

\begin{align}
    & \rho c_P \frac{\partial T}{\partial t} - \nabla\cdot (k(T) \nabla T) = \dot{q}_V\\
    & T(x = 0) = 298\;\text{K}\\
    & T(t = 0) = 298\;\text{K}.
\end{align}    

\noindent
where $k(T) = \frac{k_0}{(1+\beta(T-T_0))}, \beta = 1.2 \cdot 10^{-3}\;\text{K}^{-1}$ is the nonlinear thermal conductivity.

Both linear and nonlinear simulations were generated using FEM. The nonlinear simulations are used to build the synthetic dataset of the ground truth that leads the learning of the ignorance model to be coupled with the simulation that does not account for the nonlinear terms. The time duration of each simulation is 10 seconds.

All the dataset configurations used in this work are described and listed in Table \ref{tab:data_descr}.

\begin{table}[ht!]
\centering
\begin{tabular}{p{1.3cm}p{2.3cm}p{2.0cm}p{2.8cm}p{1.3cm}p{1.3cm}}
\toprule
    Dataset name & Domain shape & HS type & Load position & HS power \text{$P, W/m^3$} & Mesh \\
\midrule
A1 & Rectangle & Linear & Half top boundary & 15000 & Regular \\
A2 & Rectangle & Linear & Full top boundary & 15000 & Regular \\
A3 & Rectangle & Linear & Half top boundary & 15000 & Irregular (coarse)\\
A4 & Rectangle & Linear & Full top boundary & 15000 & Irregular (coarse)\\
\midrule
A5 & Rectangle & Linear & Half top boundary & 15000 & Irregular (fine)\\
A6 & Rectangle & Linear & Full top boundary & 15000 & Irregular (fine)\\
A7 & Rectangle & Linear & Half top boundary & 15000 & Submesh \\
A8 & Rectangle & Linear & Full top boundary & 15000 & Submesh \\
\midrule
B1 & Rectangle & Gaussian\linebreak distribution & --- & --- & Regular \\
B2 & L-shape & Linear & Full top boundary & 6000 & Irregular \\
\bottomrule
\end{tabular}
\vspace{0.5cm}
\caption{The description of all the datasets used in the current work (HS is the heat source).}
\label{tab:data_descr}
\end{table}

The number of frames for each simulation is 4000, and $\Delta t$ is $2.5 \cdot 10^{-3} \;\text{s}$ except for the simulations for dataset B1, where the number of frames is 200 and $\Delta t$ is $5 \cdot 10^{-2}\;\text{s}$. For each case, we apply a Dirichlet boundary condition with a constant temperature equal to $298\;\text{K}$. The initial temperature of the simulation is $298\;\text{K}$. For each dataset, we specify vector $\bn_i$ by labeling differently the nodes on the heat source, on the boundary condition, and the rest of the nodes. 

\subsection{Learning the gap with restricted number of timesteps}

\begin{figure}[ht!]
 \begin{subfigure}{0.24\textwidth}
     \includegraphics[width=\textwidth]{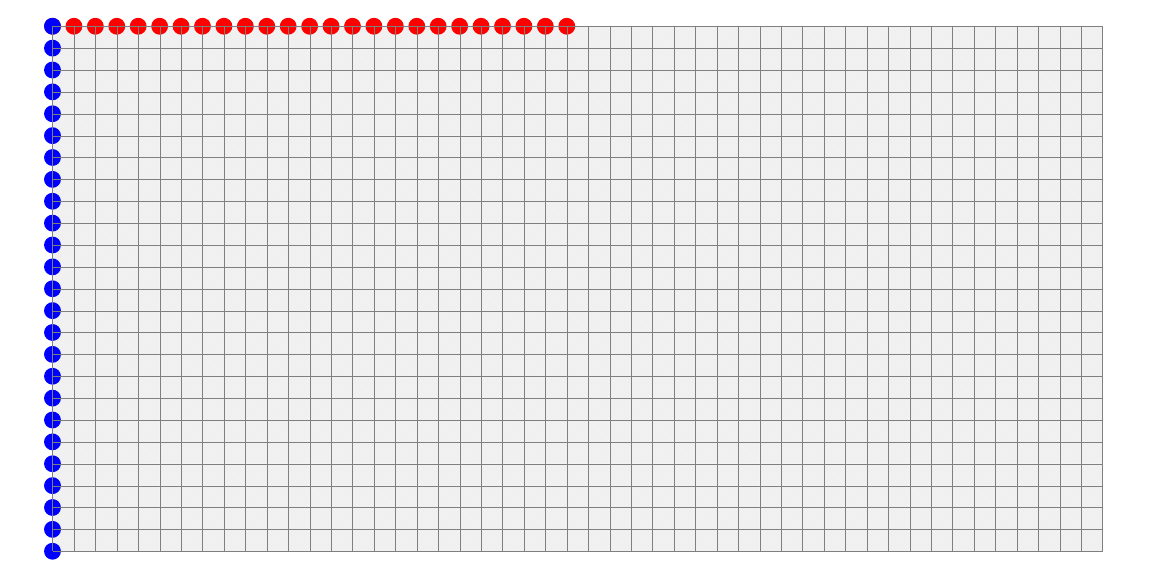}
     \caption{}
     \label{fig:schema_a1}
 \end{subfigure}
 \hfill
 \begin{subfigure}{0.24\textwidth}
     \includegraphics[width=\textwidth]{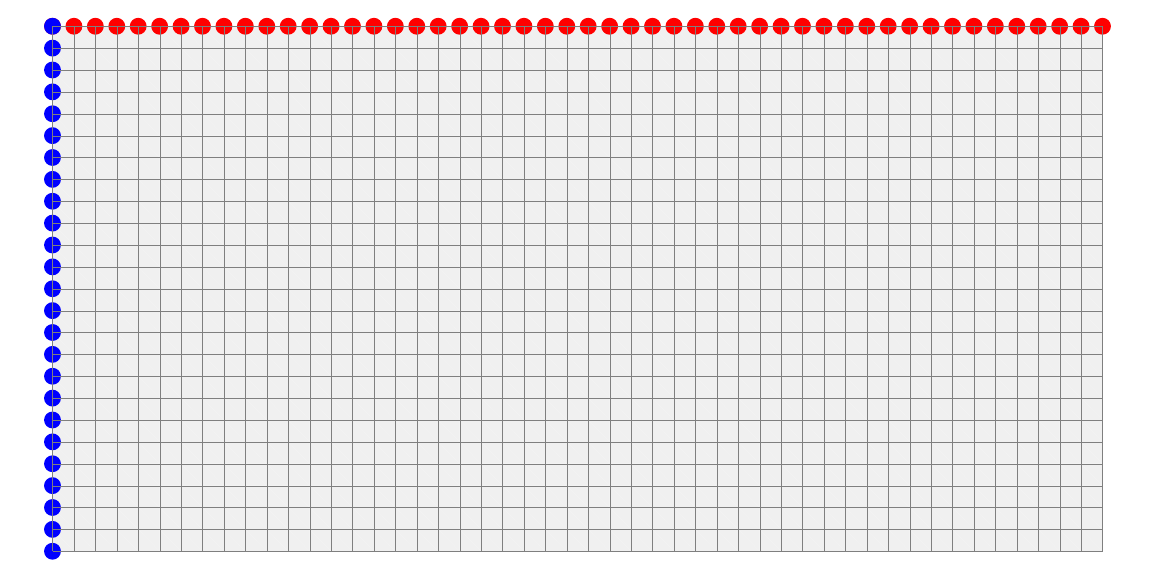}
     \caption{}
     \label{fig:schema_a2}
 \end{subfigure} 
 \hfill
 \begin{subfigure}{0.24\textwidth}
     \includegraphics[width=\textwidth]{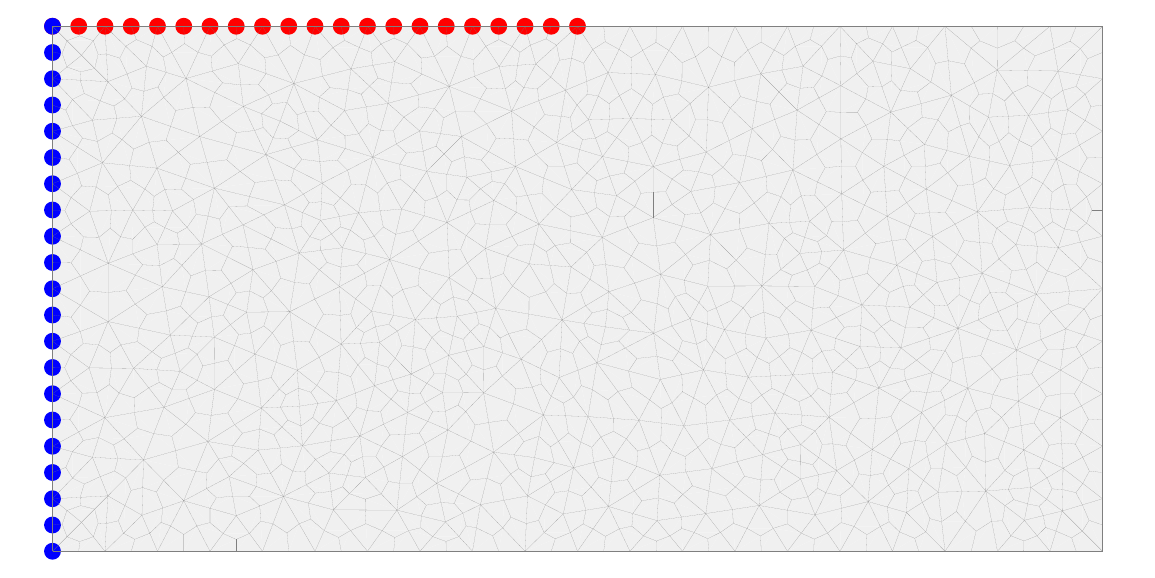}
     \caption{}
     \label{fig:schema_a3}
 \end{subfigure} 
 \hfill
 \begin{subfigure}{0.24\textwidth}
     \includegraphics[width=\textwidth]{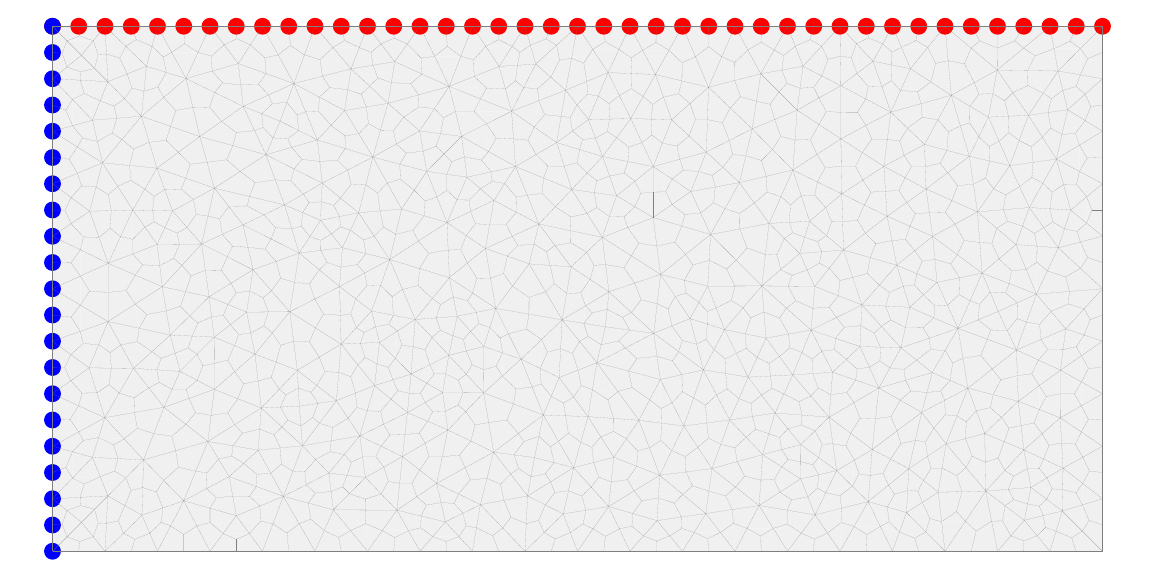}
     \caption{}
     \label{fig:schema_a4}
\end{subfigure}
\medskip
\begin{subfigure}{0.24\textwidth}
     \includegraphics[width=\textwidth]{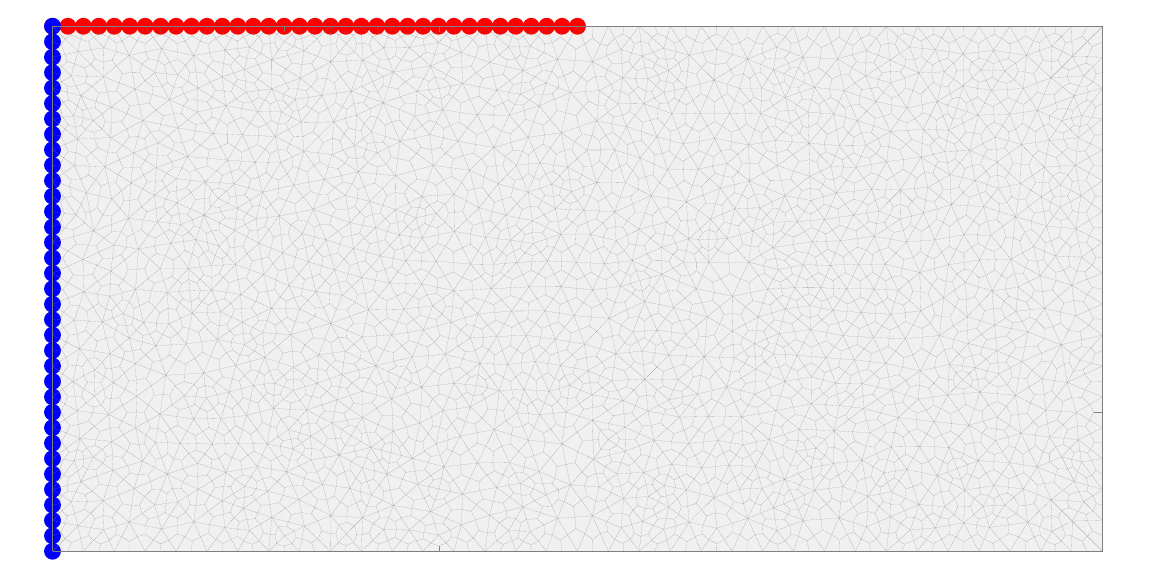}
     \caption{}
     \label{fig:schema_a5}
 \end{subfigure}
 \hfill
 \begin{subfigure}{0.24\textwidth}
     \includegraphics[width=\textwidth]{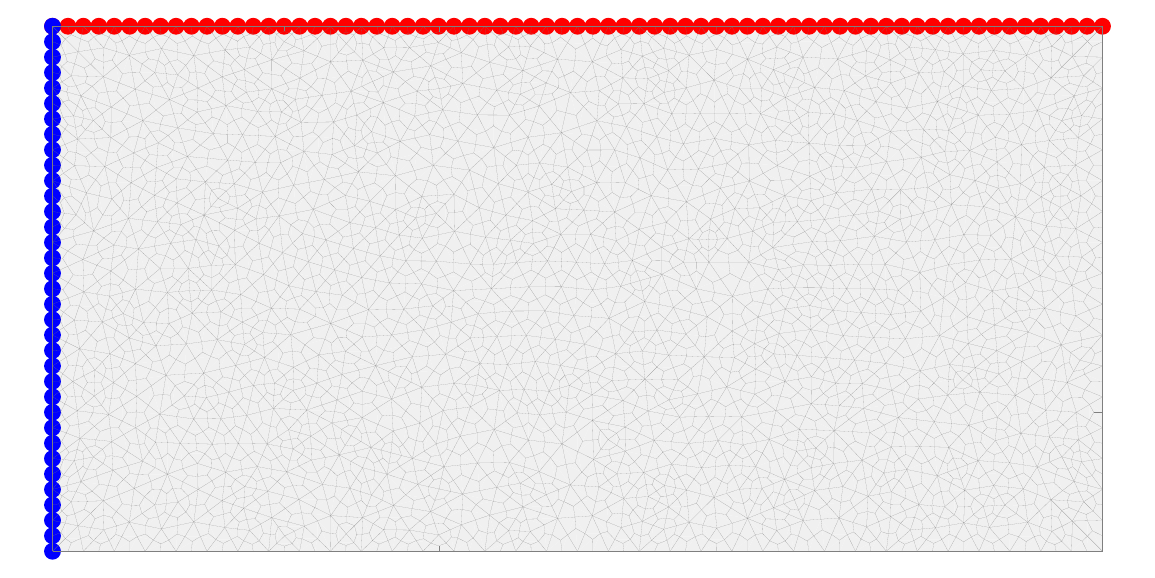}
     \caption{}
     \label{fig:schema_a6}
 \end{subfigure} 
 \hfill
 \begin{subfigure}{0.24\textwidth}
     \includegraphics[width=\textwidth]{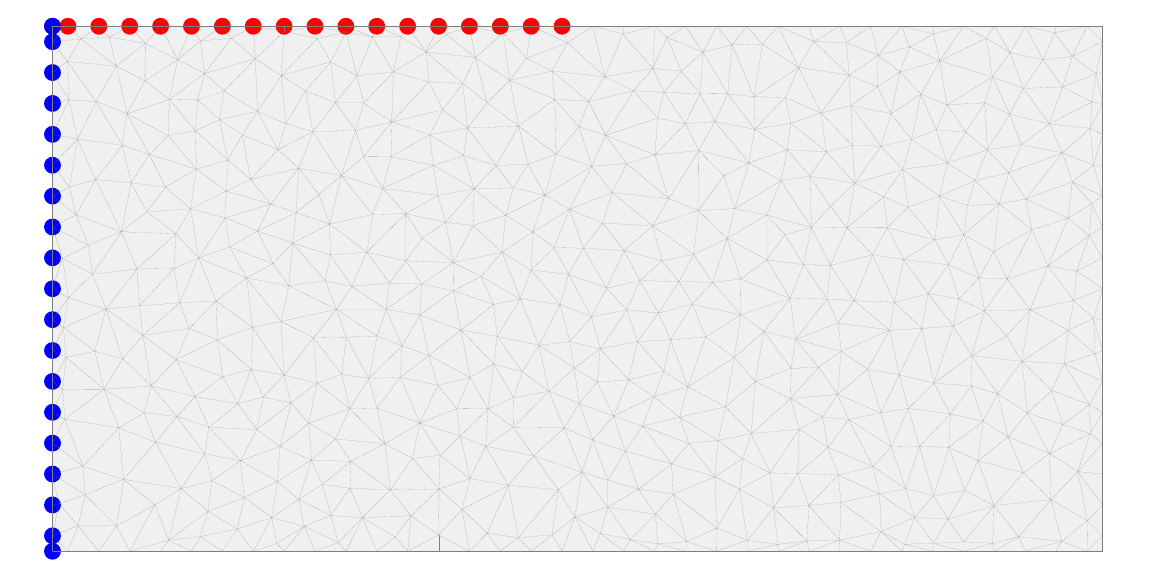}
     \caption{}
     \label{fig:schema_a7}
 \end{subfigure} 
 \hfill
 \begin{subfigure}{0.24\textwidth}
     \includegraphics[width=\textwidth]{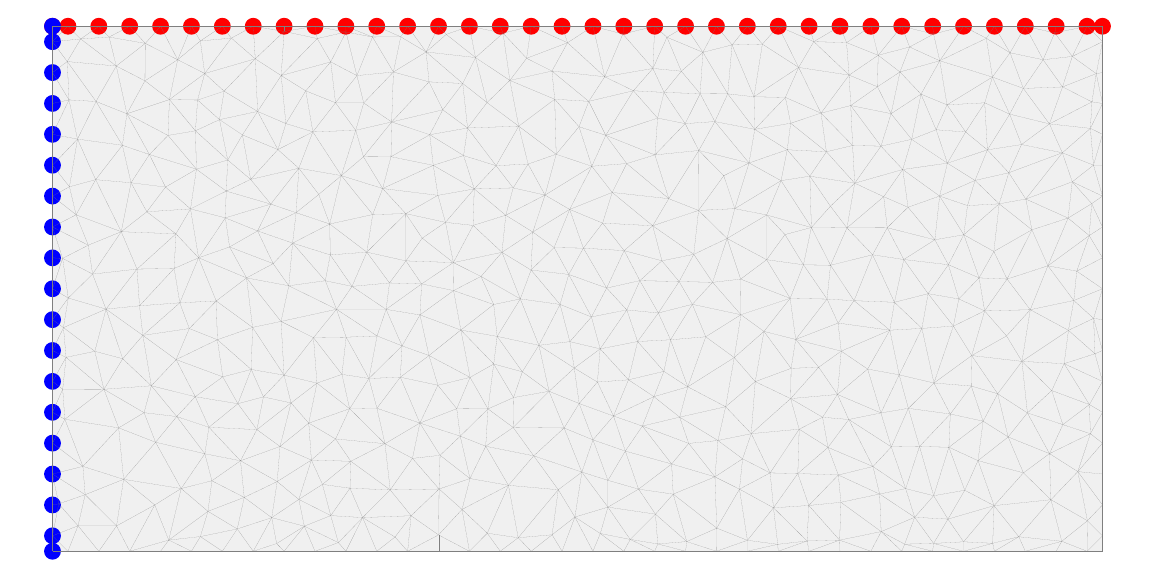}
     \caption{}
     \label{fig:schema_a8}
 \end{subfigure}
 \caption{The schematic domain representation of the data used for the use cases: (a) dataset A1; (b) dataset A2; (c) dataset A3; (d) dataset A4; (e) dataset A5; (f) dataset A6; (g) dataset A7; (h) dataset A8. The mesh nodes on the heat source are in red, the nodes of the Dirichlet BC are in blue.}
 \label{fig:schema_hybgnn_datasets}
\end{figure}

In the first use case, we test the capability of the model to learn the gap of the specified simulation with a restricted amount of data. The Figures \ref{fig:schema_a1}, \ref{fig:schema_a2} demonstrate the schematic representation of the data used for these use cases. For each of these datasets, we compare the performance of our hybrid twin and the baseline MeshGraphNet (MGN) \cite{pfaff2020learning} model without adaptive remeshing. MGN by design learns the nonlinear simulation from scratch (the details of the baseline architecture are described in \ref{appendix:base_arch}). In both cases, the training is performed on 10\% of the samples. After that, the model is evaluated on the full dataset.

\subsection{Mesh generalization}

The next use case is considered for testing the generalization capability of the model to perform accurately on different meshes. The Figures \ref{fig:schema_a3}, \ref{fig:schema_a4} display the corresponding domains and mesh configurations used for evaluation. We take the model trained on a random 50\% of frames from the datasets A1 and A2, and evaluate the performance of this model on full simulations A3 and A4 respectively.

\subsection{Training with a scarce number of spatial nodes}

The goal of the next use case is to test if the model is able to learn the gap in a setting when the number of nodes in a mesh is relatively scarce. First, we generate the data for the cases A5 and A6 (Figures \ref{fig:schema_a5} and \ref{fig:schema_a6}). After that, we select 40\% of the nodes in an original mesh and reconnect them using the Delaunay triangulation algorithm to generate a submesh. The obtained datasets A7 and A8 (Figures \ref{fig:schema_a7} and \ref{fig:schema_a8}) are used for the model training. After that, the models are evaluated on the simulations with original mesh density A5 and A6 respectively.

\subsection{Generalization over load positions and domain shapes}

\begin{figure}[ht!]
    \centering
    \includegraphics[width=0.5\textwidth]{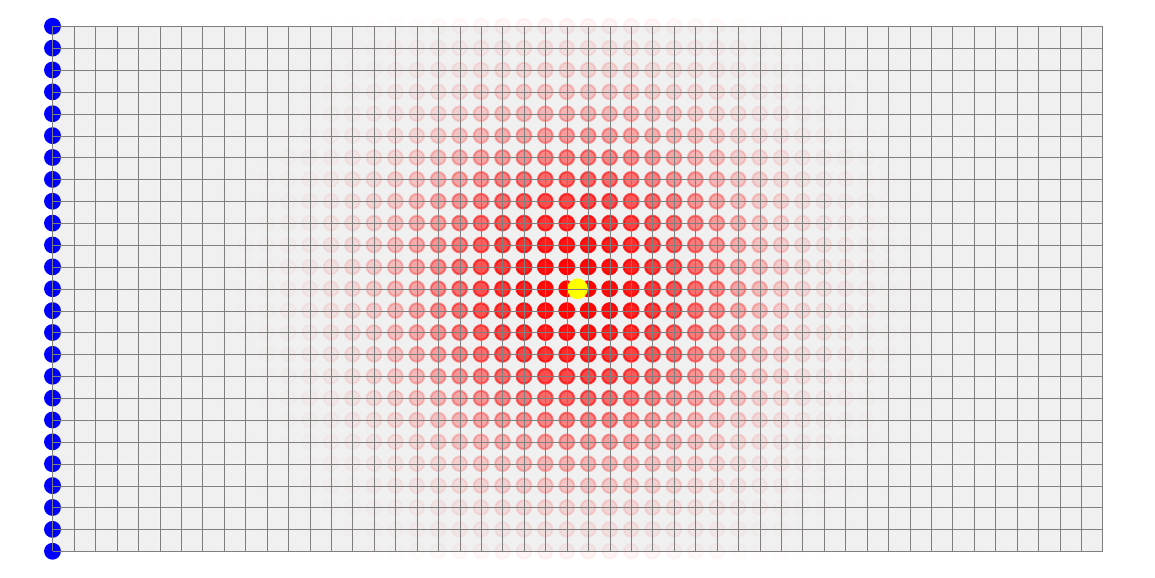}
    \caption{Schematic representation of the domain in dataset B1. The heat source is normally distributed on a plate (red), each design has a different placement of the heat source center (yellow). The Dirichlet BC is on the left side of the plate (blue).}
    \label{fig:schema_gaussian}
\end{figure}

The other use cases test the generalization capabilities of the model to different design geometries. The dataset B1 consists of 50 simulations of a heat transfer. The heat source load is normally distributed over the entire plate. Each simulation has a different position of the center of the Gaussian (Figure~\ref{fig:schema_gaussian}). The positions of the centers of the Gaussians were chosen using a Latin Hypercube Sampling (LHS) to ensure an even coverage of the design parameter space. The maximum power applied in the node is $60\; \text{W}/\text{m}^3$. Here we apply power to all the nodes on a plate, so in order to build a one-hot vector $\bn_i$, we specify not the nodes on the heat source but the node in the center of the Gaussian. 40 simulations were used for training and  10 for evaluation.

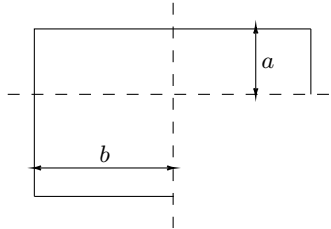
\begin{figure}[ht!]
    \centering
\begin{tikzpicture}[x=0.75pt,y=0.75pt,yscale=-0.7,xscale=0.7]

\draw   (21.57,142.2) -- (21.57,21.43) -- (220.8,21.43) ;
\draw  [dash pattern={on 4.5pt off 4.5pt}]  (2.4,68.6) -- (241.2,68.6) ;
\draw  [dash pattern={on 4.5pt off 4.5pt}]  (121.57,2.57) -- (121.57,165) ;
\draw    (21.57,142.2) -- (122,142.2) ;
\draw    (220.8,21.43) -- (220.8,68.6) ;
\draw    (24,121.69) -- (119.6,121.69) ;
\draw [shift={(121.6,121.69)}, rotate = 180] [color={rgb, 255:red, 0; green, 0; blue, 0 }  ][line width=0.75]    (4.37,-1.32) .. controls (2.78,-0.56) and (1.32,-0.12) .. (0,0) .. controls (1.32,0.12) and (2.78,0.56) .. (4.37,1.32)   ;
\draw [shift={(22,121.69)}, rotate = 0] [color={rgb, 255:red, 0; green, 0; blue, 0 }  ][line width=0.75]    (4.37,-1.32) .. controls (2.78,-0.56) and (1.32,-0.12) .. (0,0) .. controls (1.32,0.12) and (2.78,0.56) .. (4.37,1.32)   ;
\draw    (181.09,22.36) -- (181.09,66.6) ;
\draw [shift={(181.09,68.6)}, rotate = 270] [color={rgb, 255:red, 0; green, 0; blue, 0 }  ][line width=0.75]    (4.37,-1.32) .. controls (2.78,-0.56) and (1.32,-0.12) .. (0,0) .. controls (1.32,0.12) and (2.78,0.56) .. (4.37,1.32)   ;
\draw [shift={(181.09,20.36)}, rotate = 90] [color={rgb, 255:red, 0; green, 0; blue, 0 }  ][line width=0.75]    (4.37,-1.32) .. controls (2.78,-0.56) and (1.32,-0.12) .. (0,0) .. controls (1.32,0.12) and (2.78,0.56) .. (4.37,1.32)   ;

\draw (183.41,40.3) node [anchor=north west][inner sep=0.75pt]    {$a$};
\draw (66.7,103.35) node [anchor=north west][inner sep=0.75pt]    {$b$};
\end{tikzpicture}
    \caption{Parameterization of the domain shape in B2 dataset.}
    \label{fig:b2_param}
\end{figure}

The B2 dataset is used to test the generalization to different domain shapes and different meshes. It includes simulations with L-shapes. The shape of the domain is parameterized with two parameters $a$ and $b$ (Figure \ref{fig:b2_param}).

\begin{figure}
\begin{subfigure}{0.24\textwidth}
     \includegraphics[width=\textwidth]{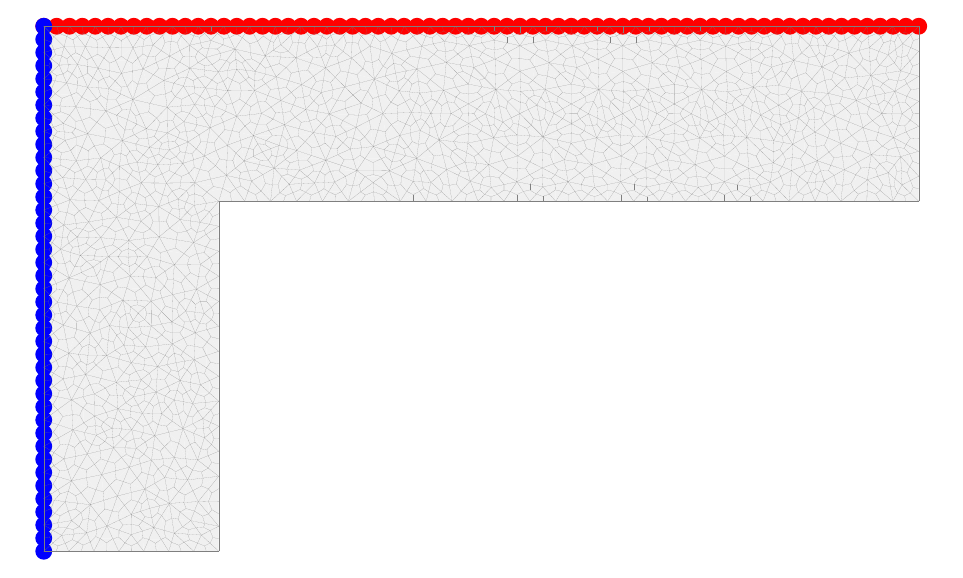}
     \caption{$a = b = 0.4$}
     \label{fig:d1}
 \end{subfigure}
 \hfill
 \begin{subfigure}{0.24\textwidth}
     \includegraphics[width=\textwidth]{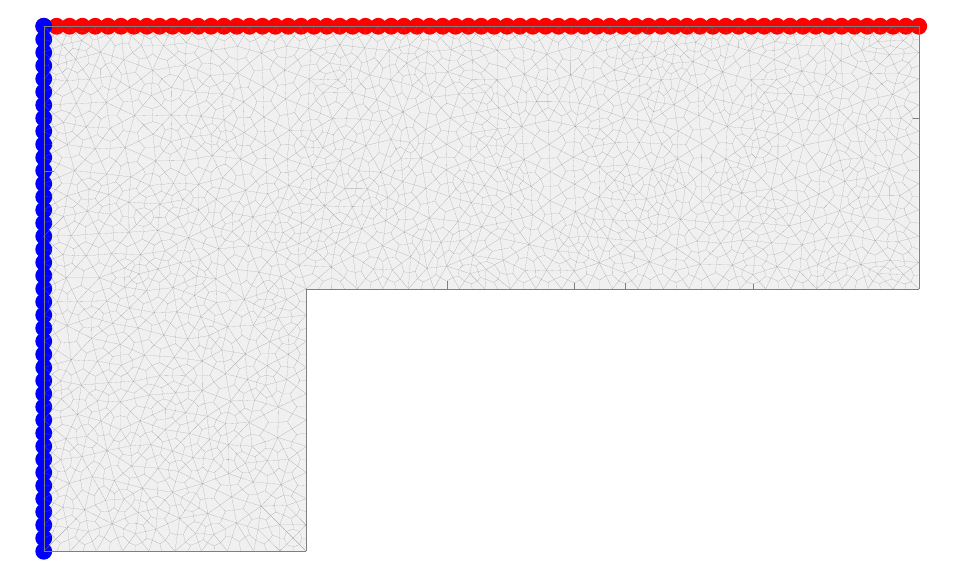}
     \caption{$a = b = 0.6$}
     \label{fig:d2}
 \end{subfigure}
  \hfill
 \begin{subfigure}{0.24\textwidth}
     \includegraphics[width=\textwidth]{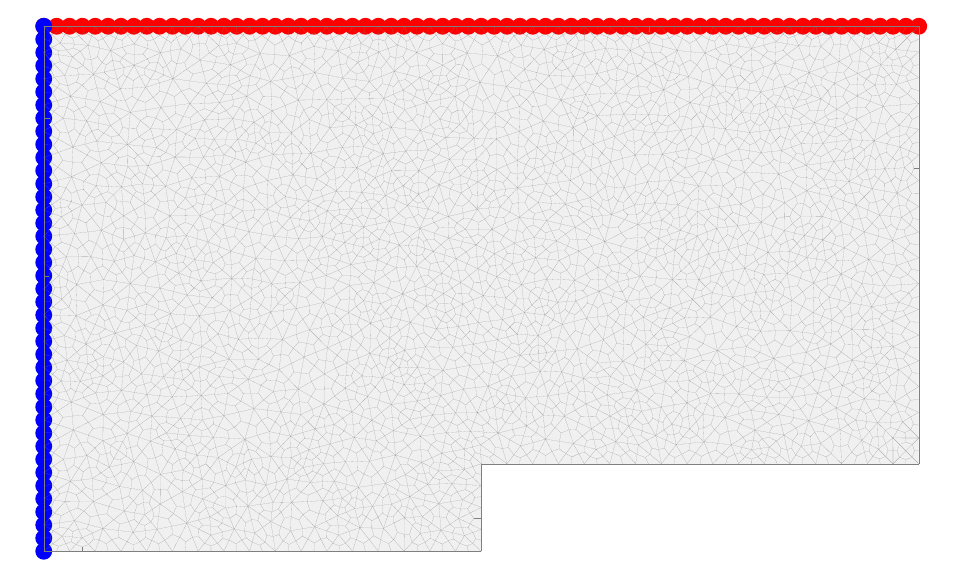}
     \caption{$a = b = 1.0$}
     \label{fig:d4}
 \end{subfigure}
  \hfill
 \begin{subfigure}{0.24\textwidth}
     \includegraphics[width=\textwidth]{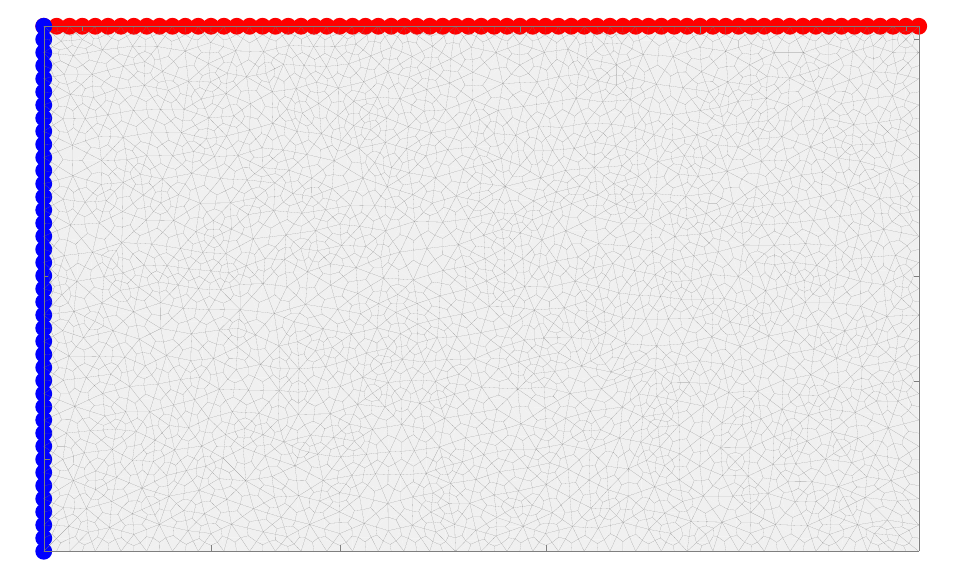}
     \caption{$a = b = 1.2$}
     \label{fig:d5}
 \end{subfigure}
 \medskip
 \begin{subfigure}{0.24\textwidth}
     \includegraphics[width=\textwidth]{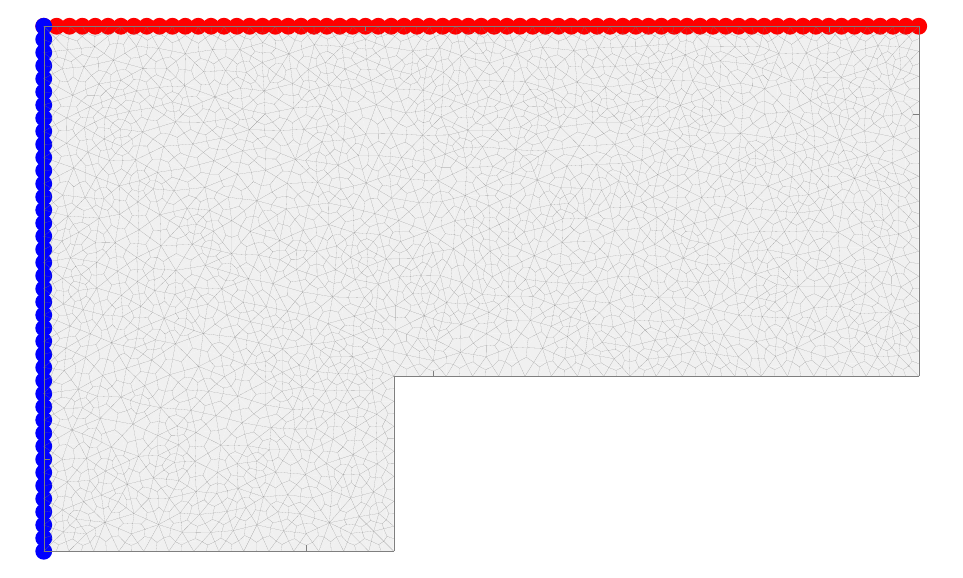}
     \caption{$a = b = 0.8$}
     \label{fig:d3}
 \end{subfigure}
  \hfill
 \begin{subfigure}{0.24\textwidth}
     \includegraphics[width=\textwidth]{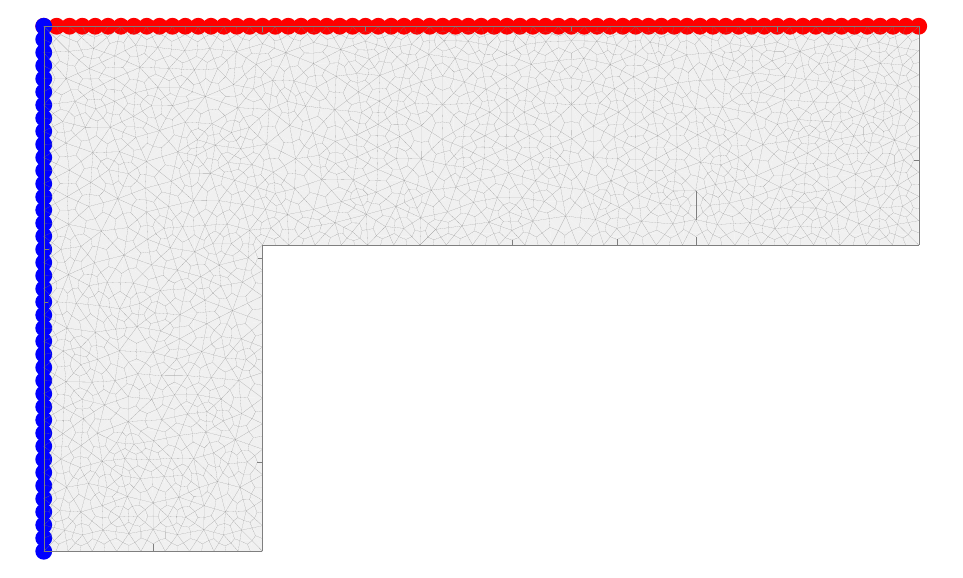}
     \caption{$a = b = 0.5$}
     \label{fig:d4.5}
 \end{subfigure}
  \hfill
 \begin{subfigure}{0.24\textwidth}
     \includegraphics[width=\textwidth]{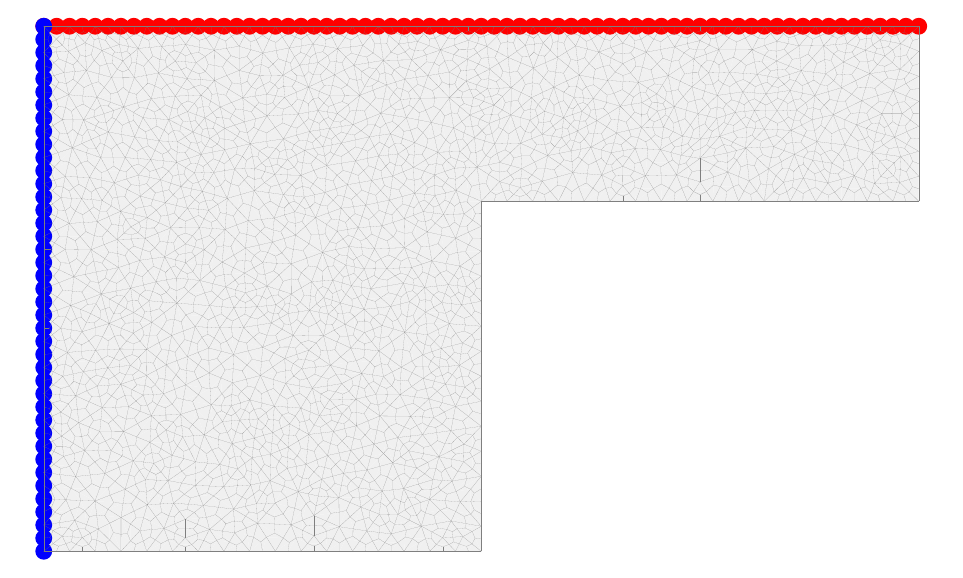}
     \caption{$a = 0.4, b = 1.2$}
     \label{fig:aneqb1}
 \end{subfigure}
  \hfill
 \begin{subfigure}{0.24\textwidth}
     \includegraphics[width=\textwidth]{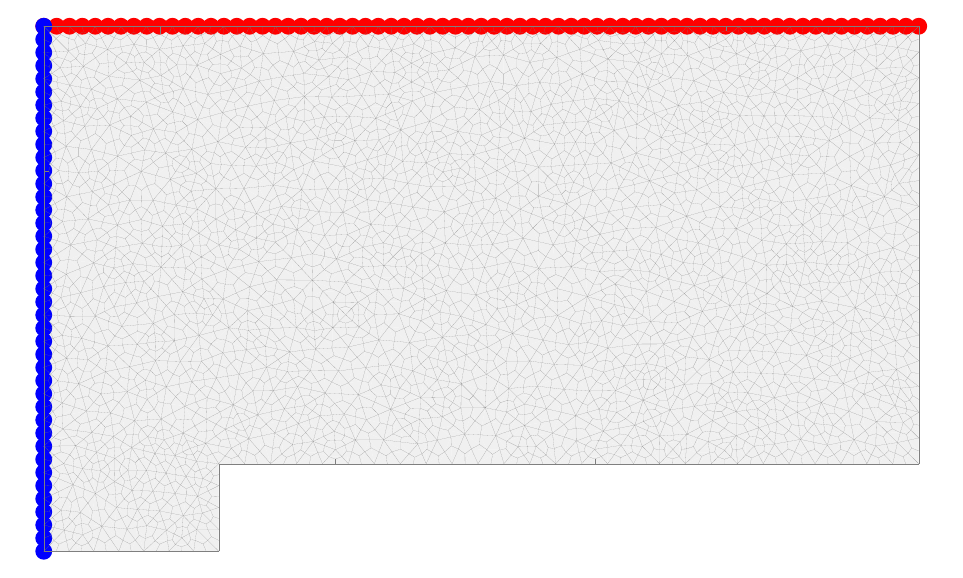}
     \caption{$a = 1.0, b=0.4$}
     \label{fig:aneqb2}
 \end{subfigure}
 \caption{Schematic representation of the simulation domains used in dataset B2. The mesh nodes on the heat source are in red, the nodes of the Dirichlet BC are in blue.}
    \label{fig:schema_b2}
\end{figure}

Figures \ref{fig:d1}, \ref{fig:d2}, \ref{fig:d4}, and \ref{fig:d5} demonstrate the schematic domain representation of the simulations used for training. The following shapes have two parameters $a$ and $b$, $a = b$, $a, b \in [0.4, 1]$.  The domain shapes in Figures \ref{fig:d3}, \ref{fig:d4.5}, \ref{fig:aneqb1} and \ref{fig:aneqb2} are used for the evaluation.

In all use cases, we use minmax normalization of the data before performing training and inference of the model. The accuracy of the prediction of the temperature field is evaluated with mean absolute error:

\begin{equation*}
    \text{MAE} = \frac{1}{n} \sum_{i=1}^{n} \left| T_i^{\text{GT}} - \hat{T}_i \right| = \frac{1}{n} \sum_{i=1}^{n} \left| T_i^{\text{GT}} - T^{\text{FEM}}_i - \hat y_i \right|,
\end{equation*}

and mean absolute percentage error:

\begin{equation*}
    \text{MAPE} = \frac{100\%}{n} \sum_{i=1}^{n} \left| \frac{T_i^{\text{GT}} - \hat{T}_i}{T_i^{\text{GT}}} \right| = \frac{100\%}{n} \sum_{i=1}^{n} \left| \frac{T_i^{\text{GT}} - T^{\text{FEM}}_i - \hat y_i}{T_i^{\text{GT}}} \right|.
\end{equation*}

During the training of each model, we use gradient clipping. To assess the training robustness, we perform 3 runs for each use case with different random seeds.

\section{Results}\label{results}

Here, we present a detailed evaluation of the proposed hybrid twin across all use cases introduced in Section 4. We analyze its accuracy, robustness, and generalization capabilities under variations in mesh type, spatial discretization, load configurations, and domain geometries.

\subsection{Learning the gap with restricted number of timesteps}

In the first setting, we test the capacity of training with limited data. We use just 10\% of the dataset samples containing the linear FEM approximation and the corresponding synthetic experimental data replicating the ground truth, on two independent problems, A1 and A2, learning an ignorance model for each configuration.

\begin{figure}[ht!]
\begin{tabular}{cccccc}
 $t$ & $1$ & $500$ & $1500$ & $3000$ & $4001$ \\ 
 \rotatebox[origin=c]{90}{\parbox[c]{-2cm}{\centering Linear}} & {\includegraphics[width=0.19\textwidth]{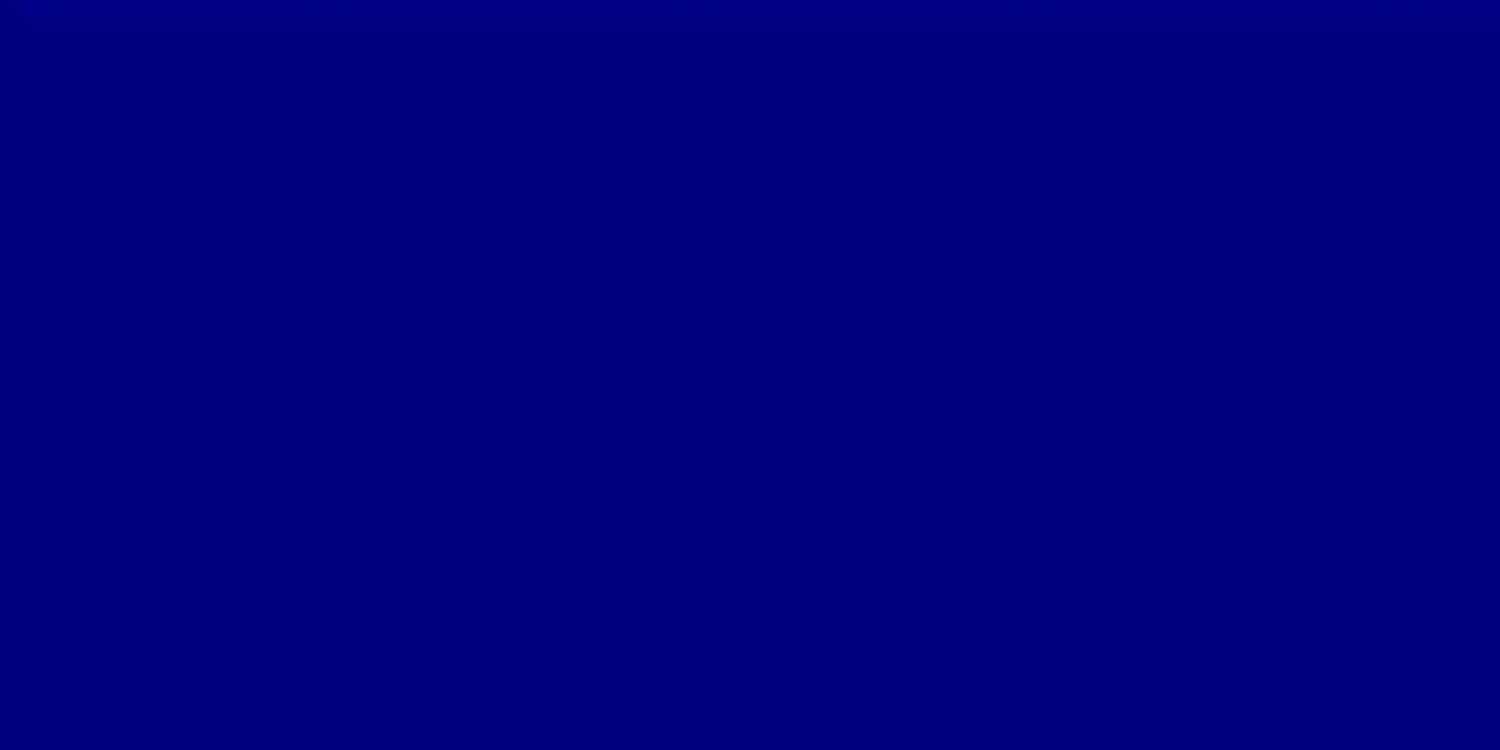}} & {\includegraphics[width=0.19\textwidth]{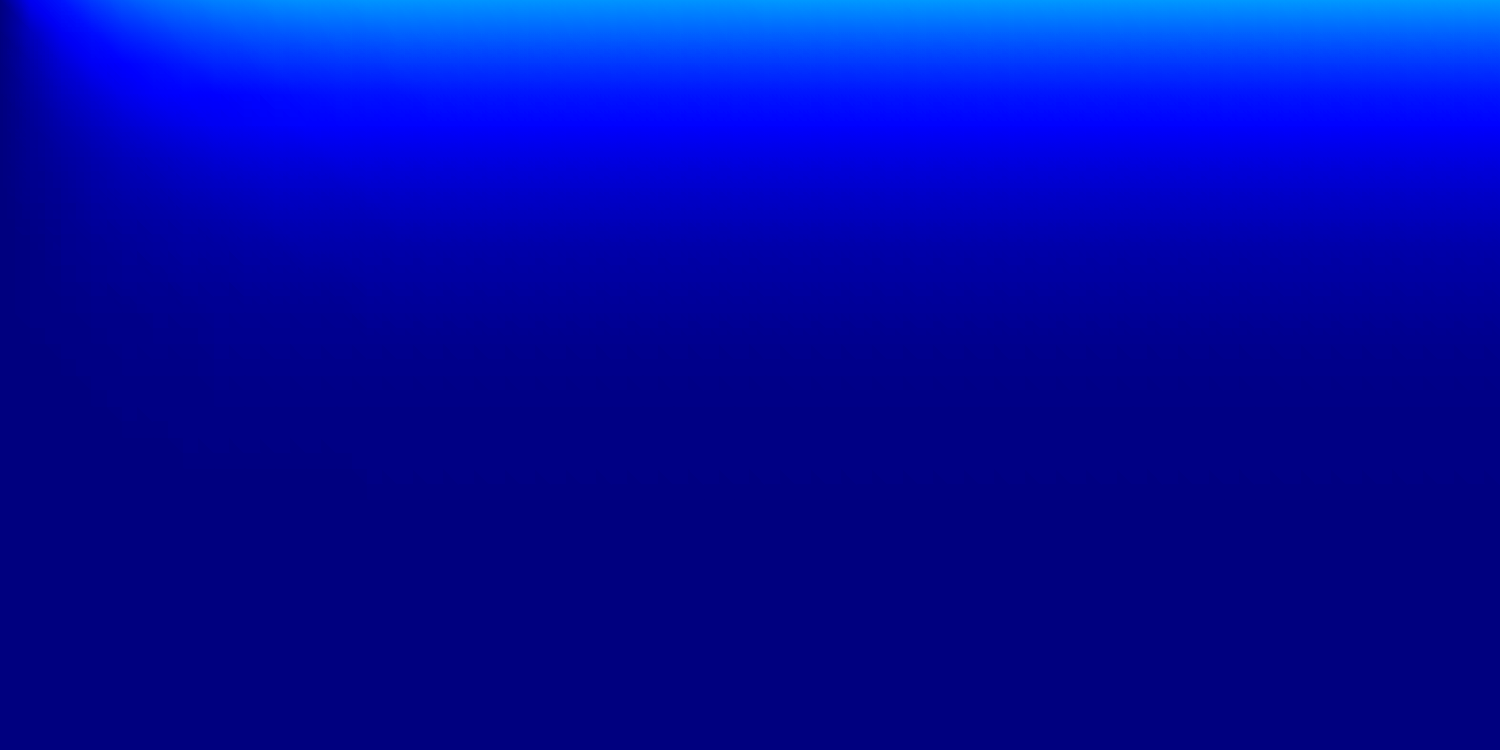}} & {\includegraphics[width=0.19\textwidth]{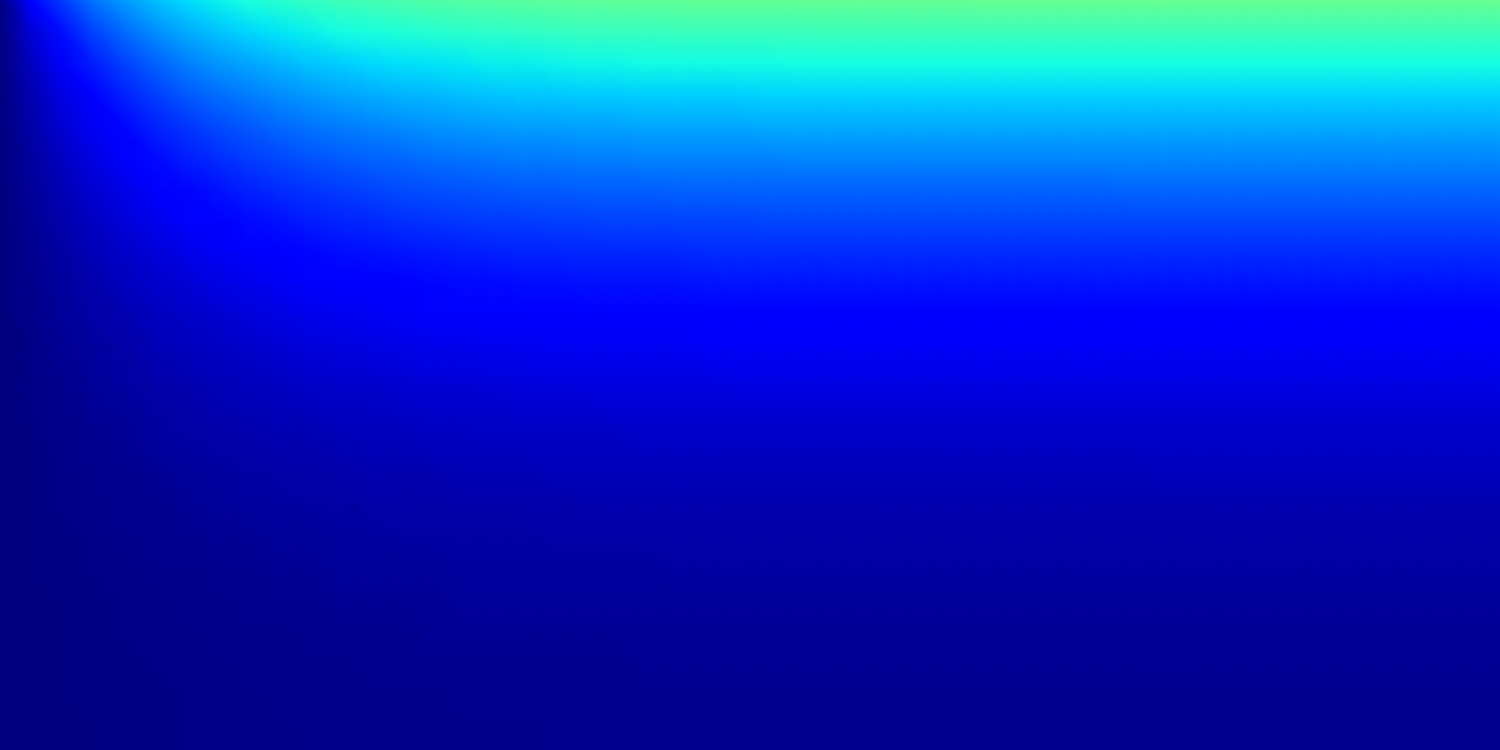}} & {\includegraphics[width=0.19\textwidth]{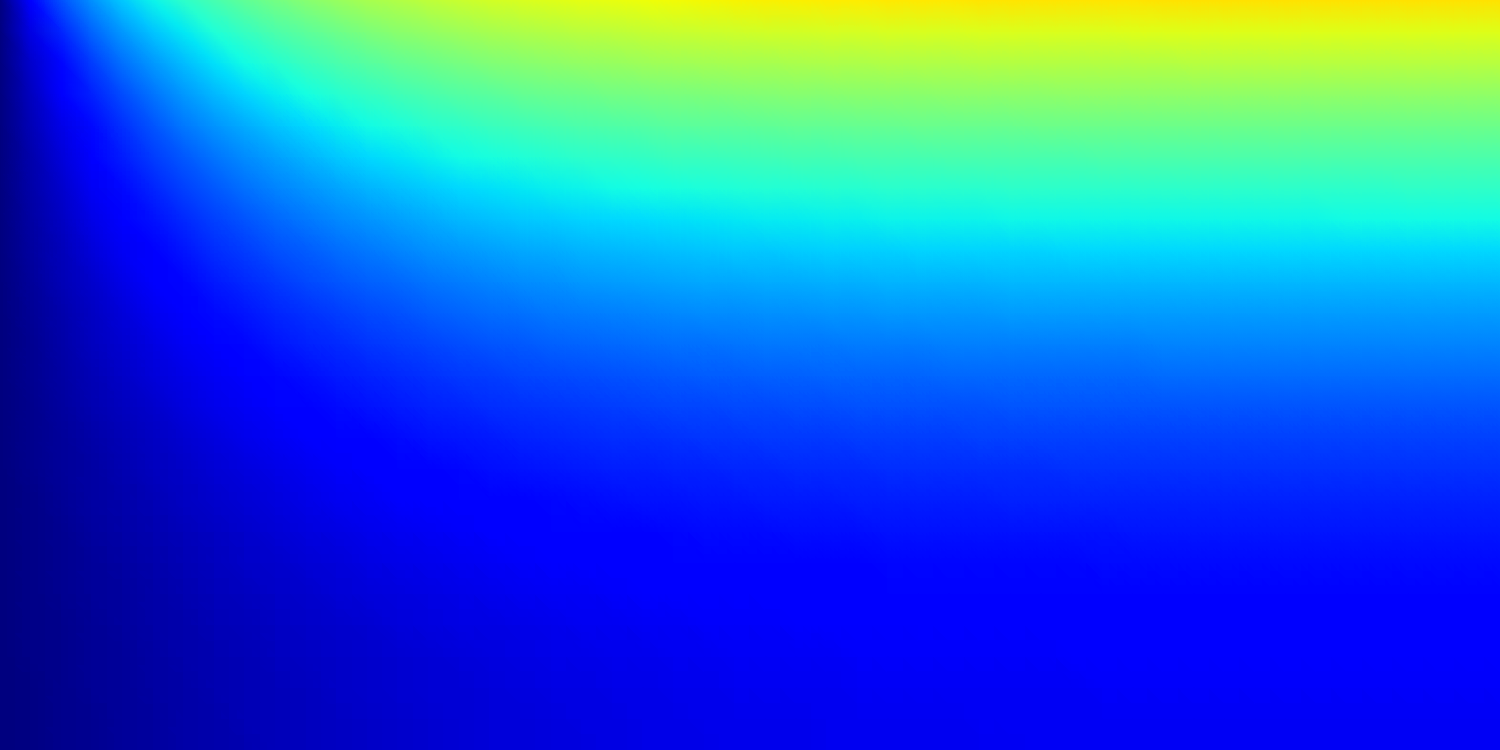}} & {\includegraphics[width=0.19\textwidth]{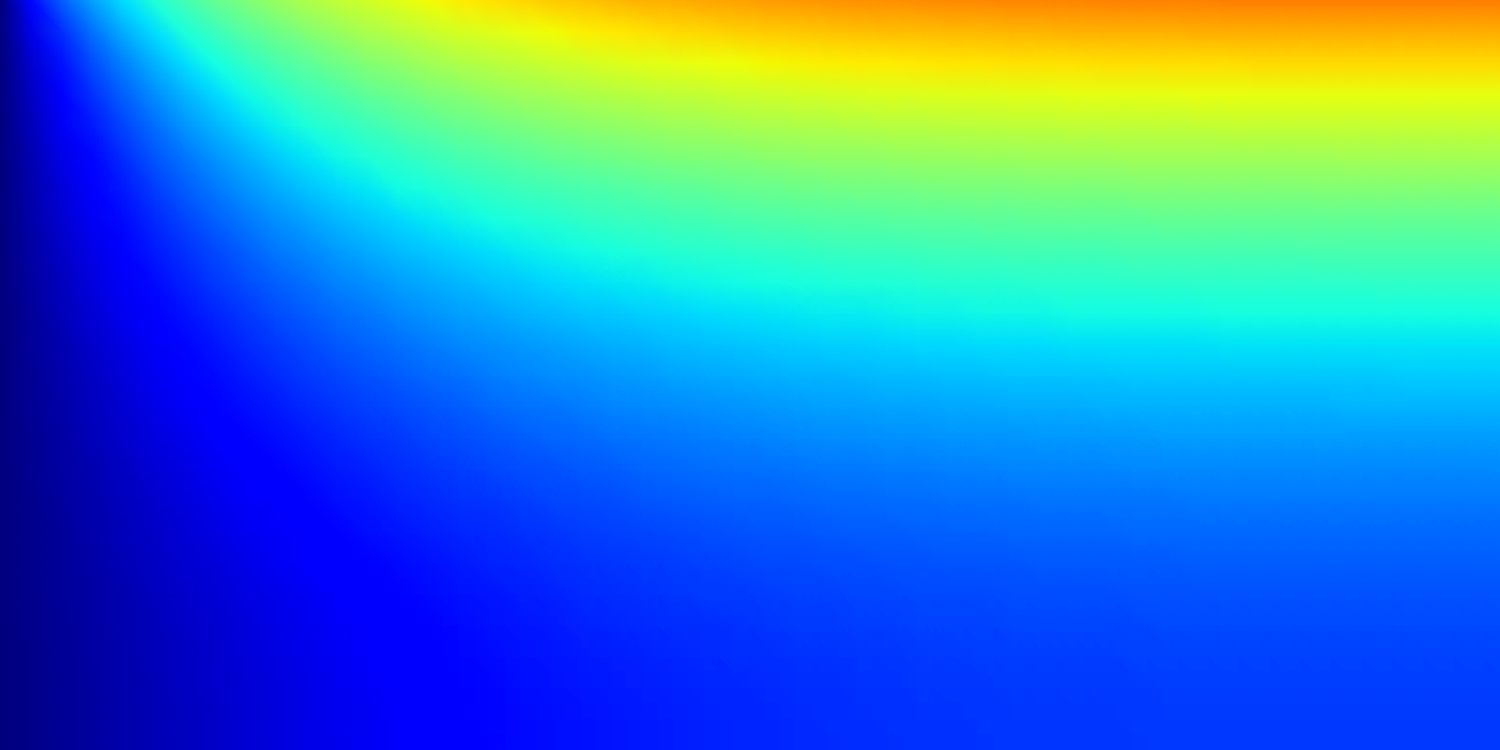}} \\
\rotatebox[origin=c]{90}{\parbox[c]{-2cm}{HT}} & {\includegraphics[width=0.19\textwidth]{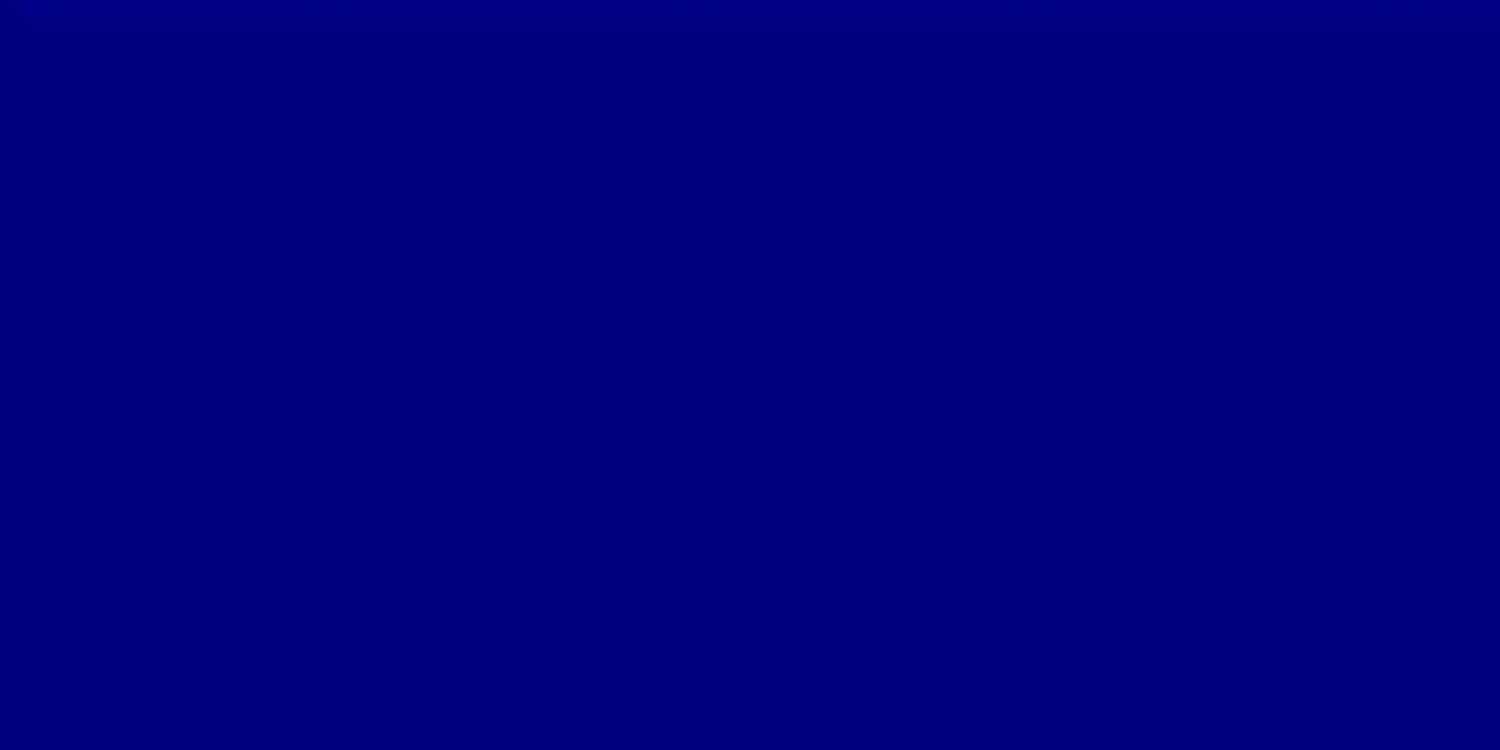}} & {\includegraphics[width=0.19\textwidth]{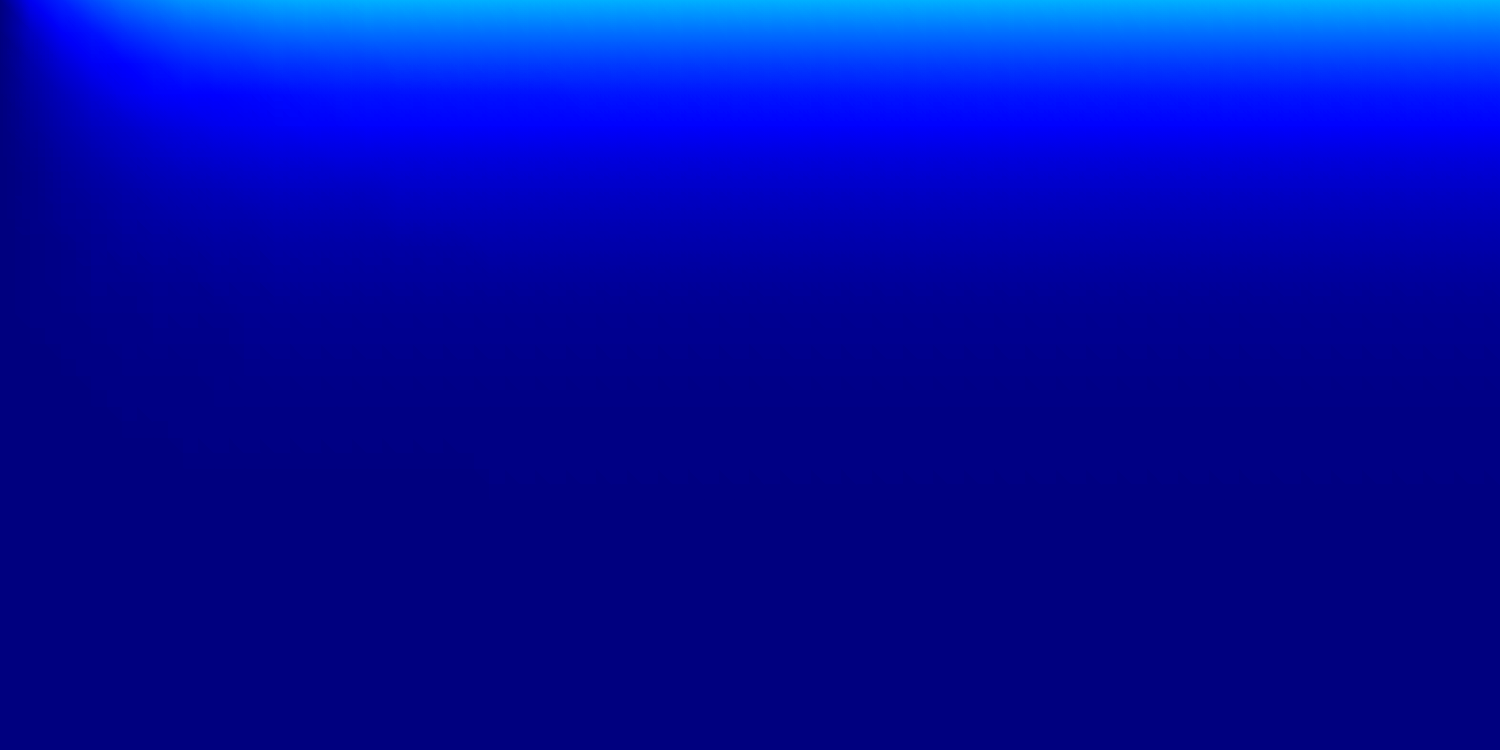}} & {\includegraphics[width=0.19\textwidth]{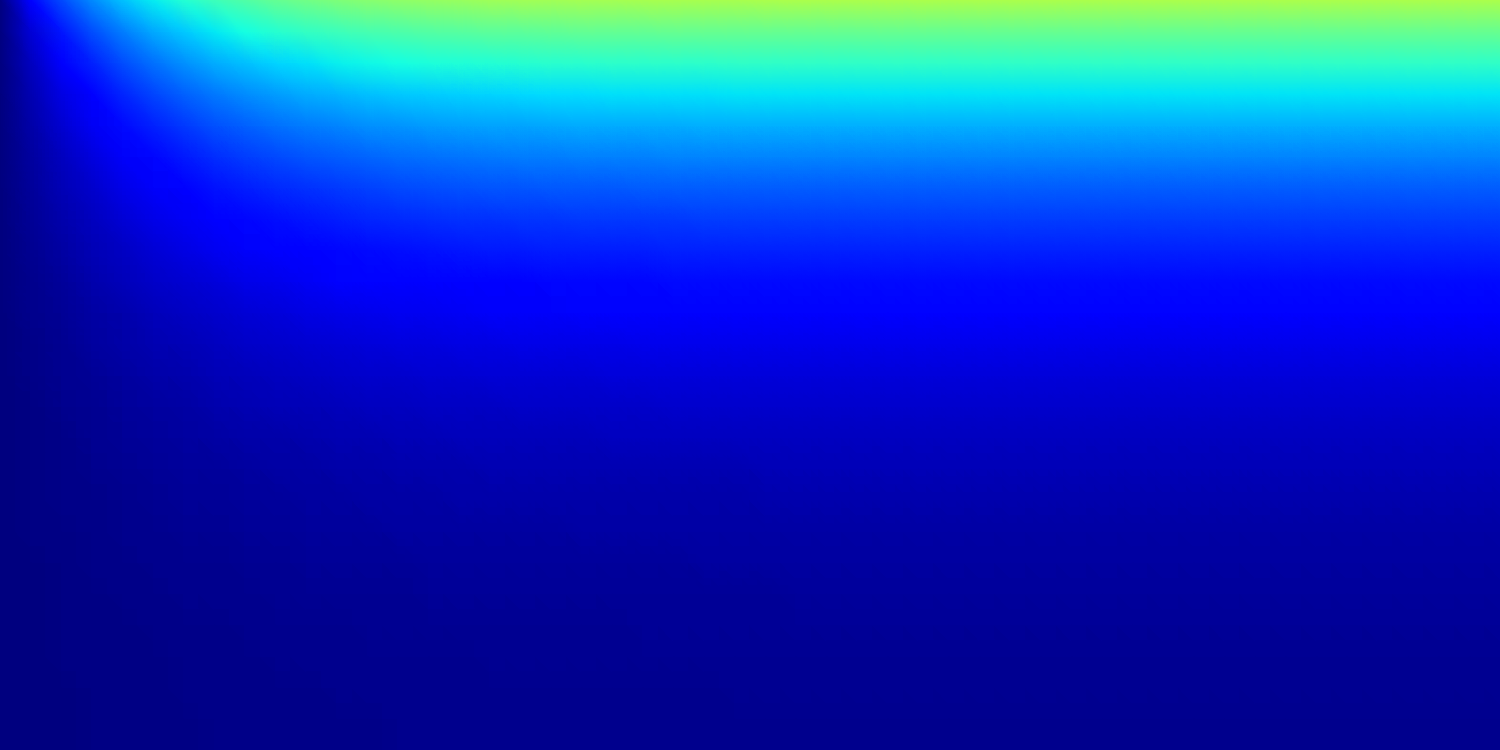}} & {\includegraphics[width=0.19\textwidth]{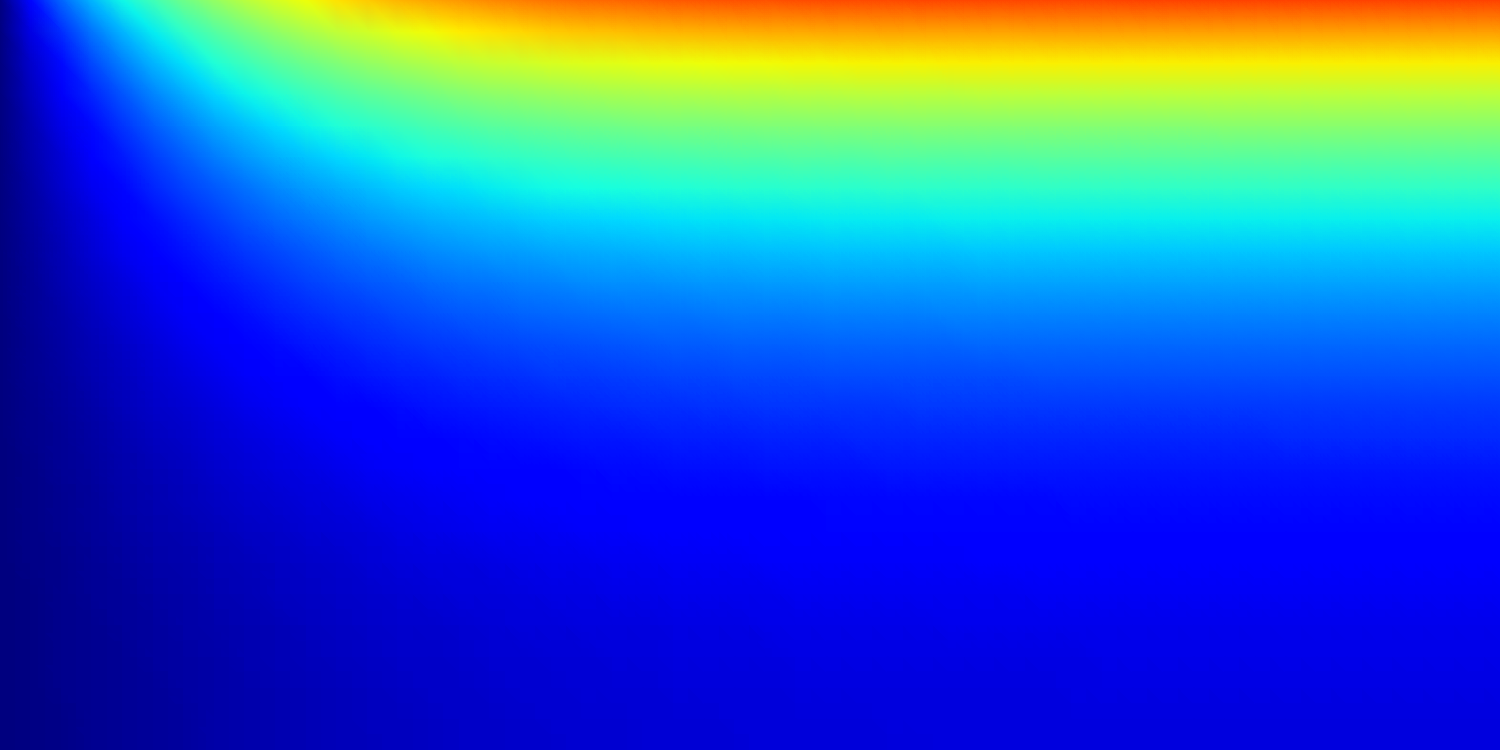}} & {\includegraphics[width=0.19\textwidth]{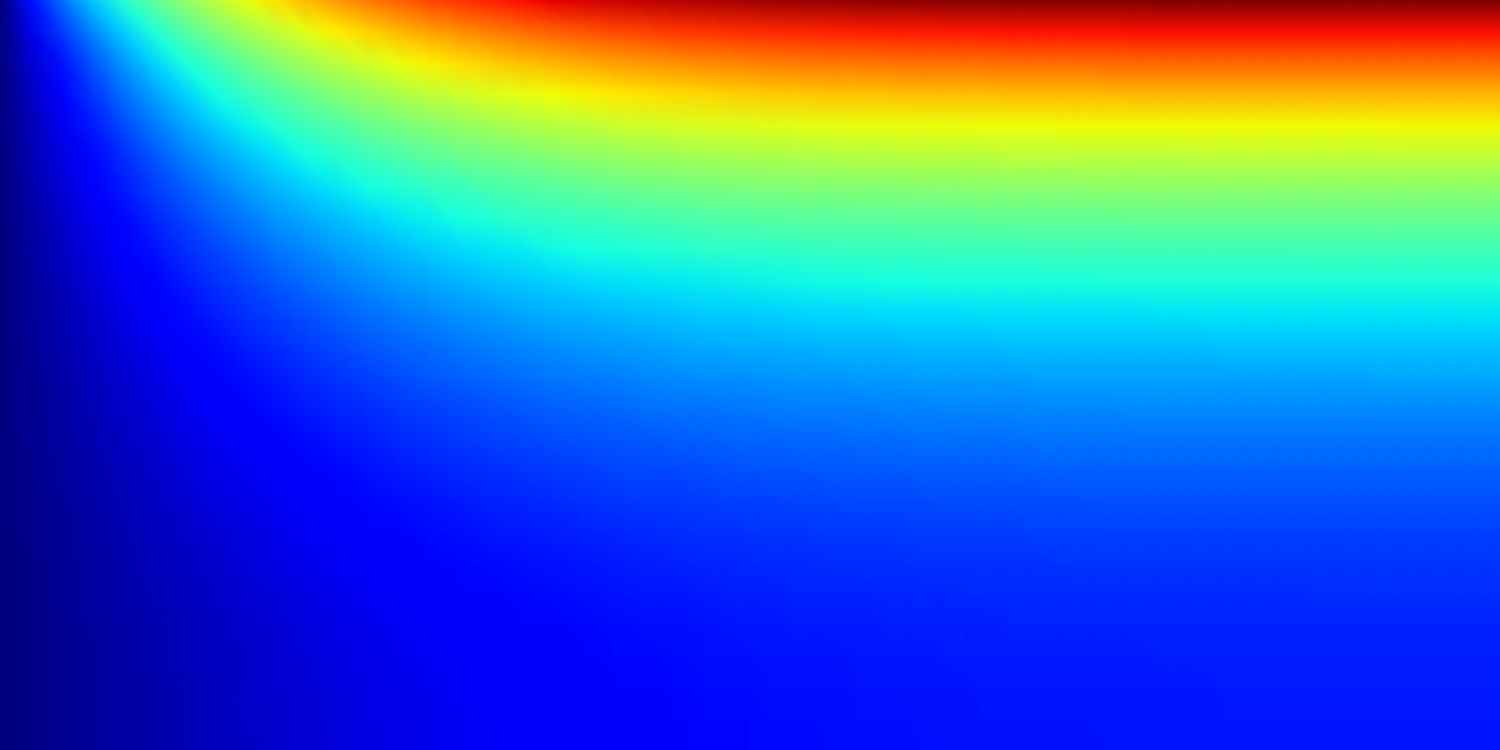}} \\
\rotatebox[origin=c]{90}{\parbox[c]{-20cm}{GT}} & {\includegraphics[width=0.19\textwidth]{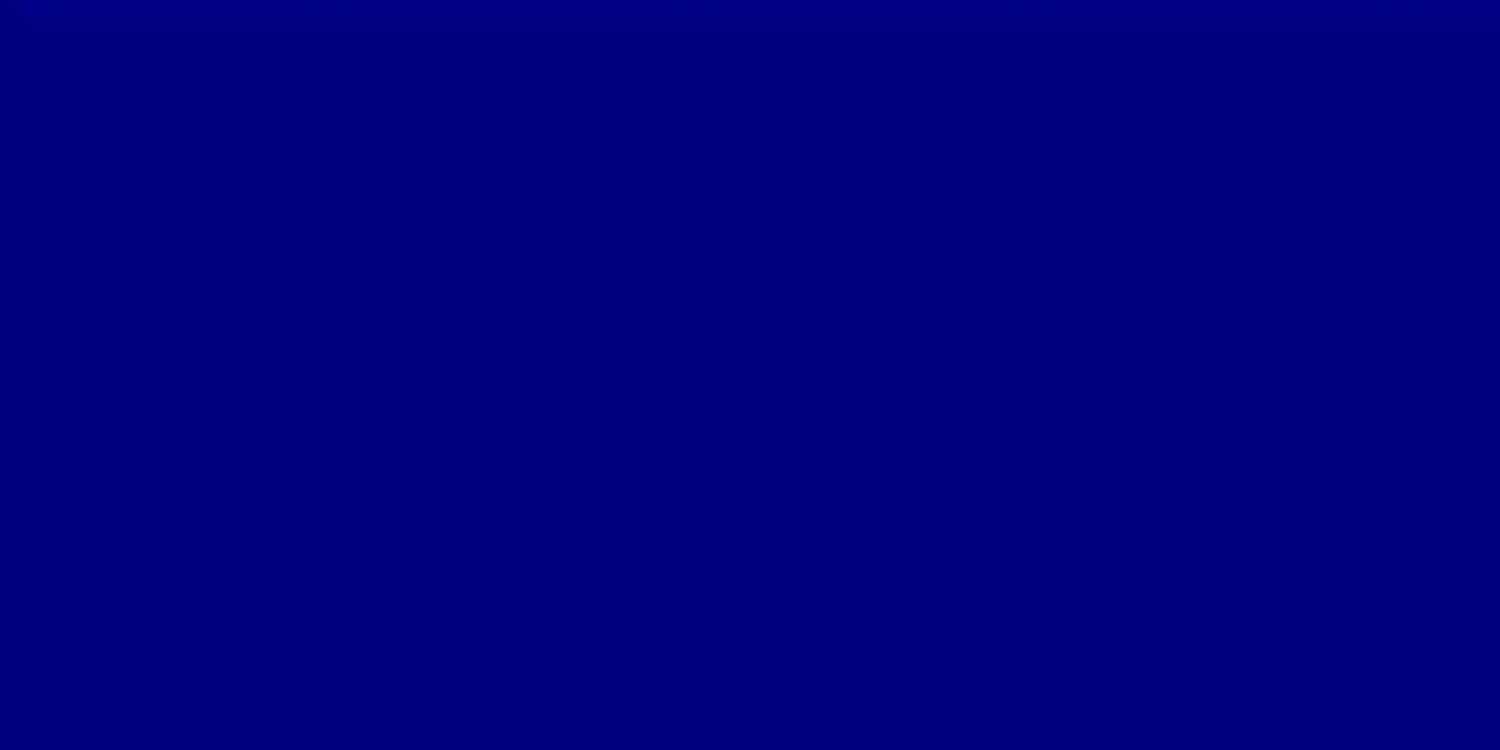}} & {\includegraphics[width=0.19\textwidth]{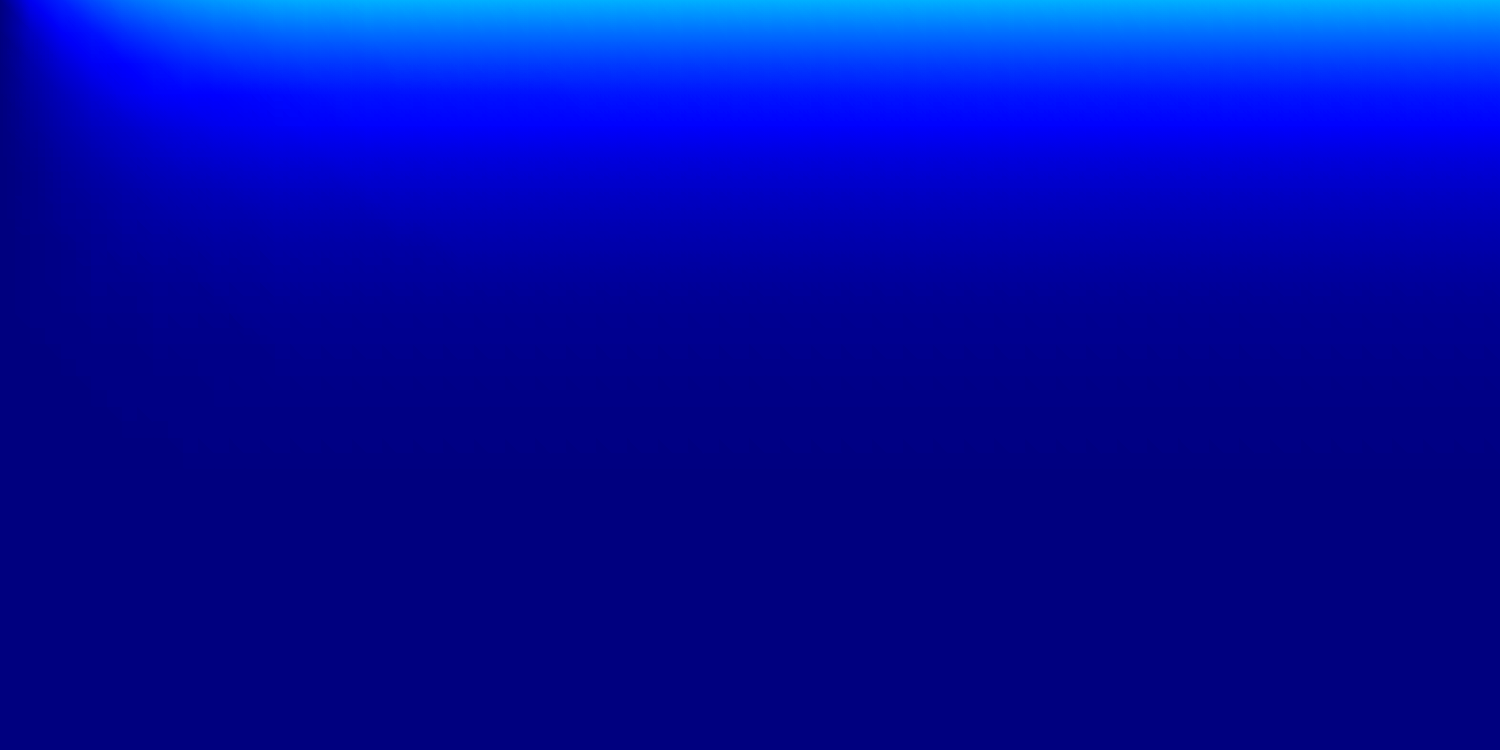}} & {\includegraphics[width=0.19\textwidth]{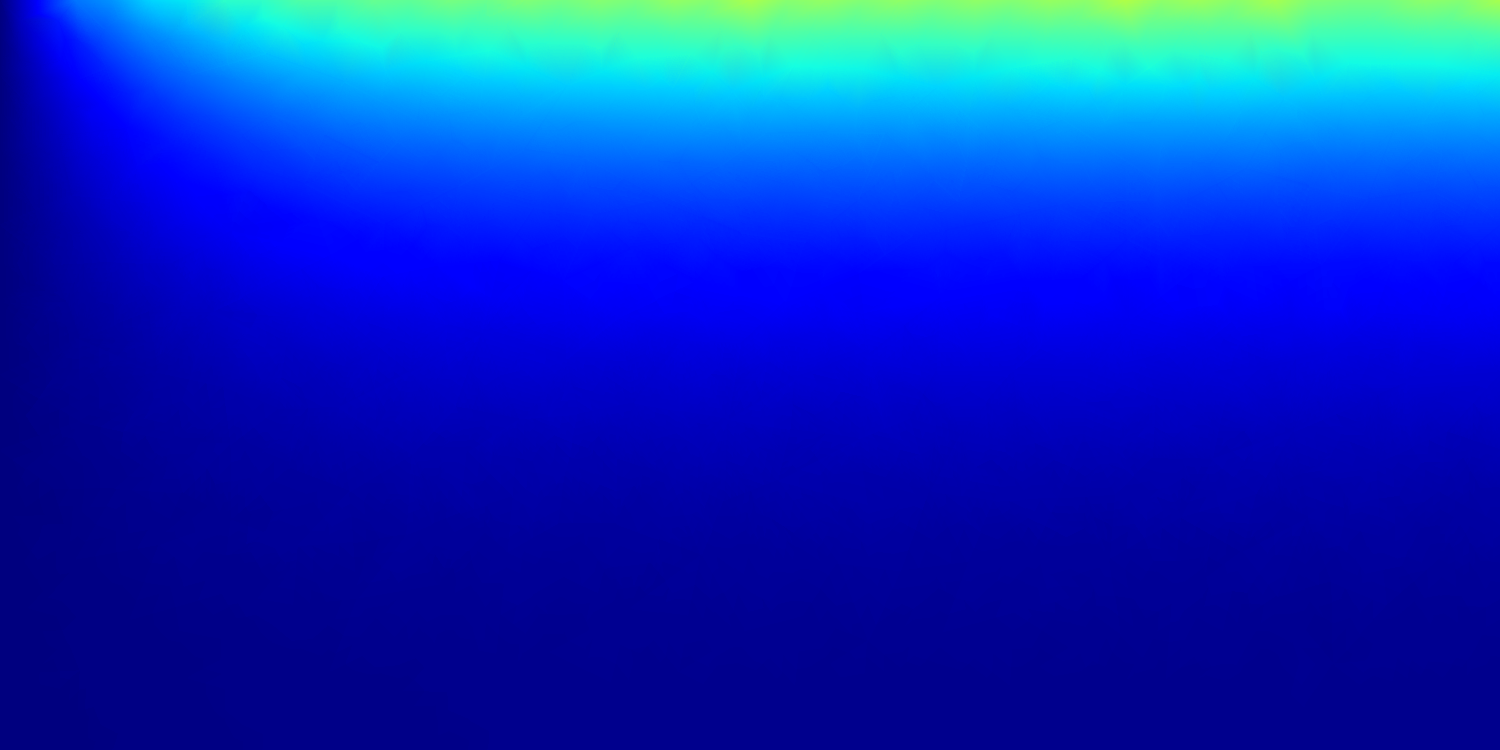}} & {\includegraphics[width=0.19\textwidth]{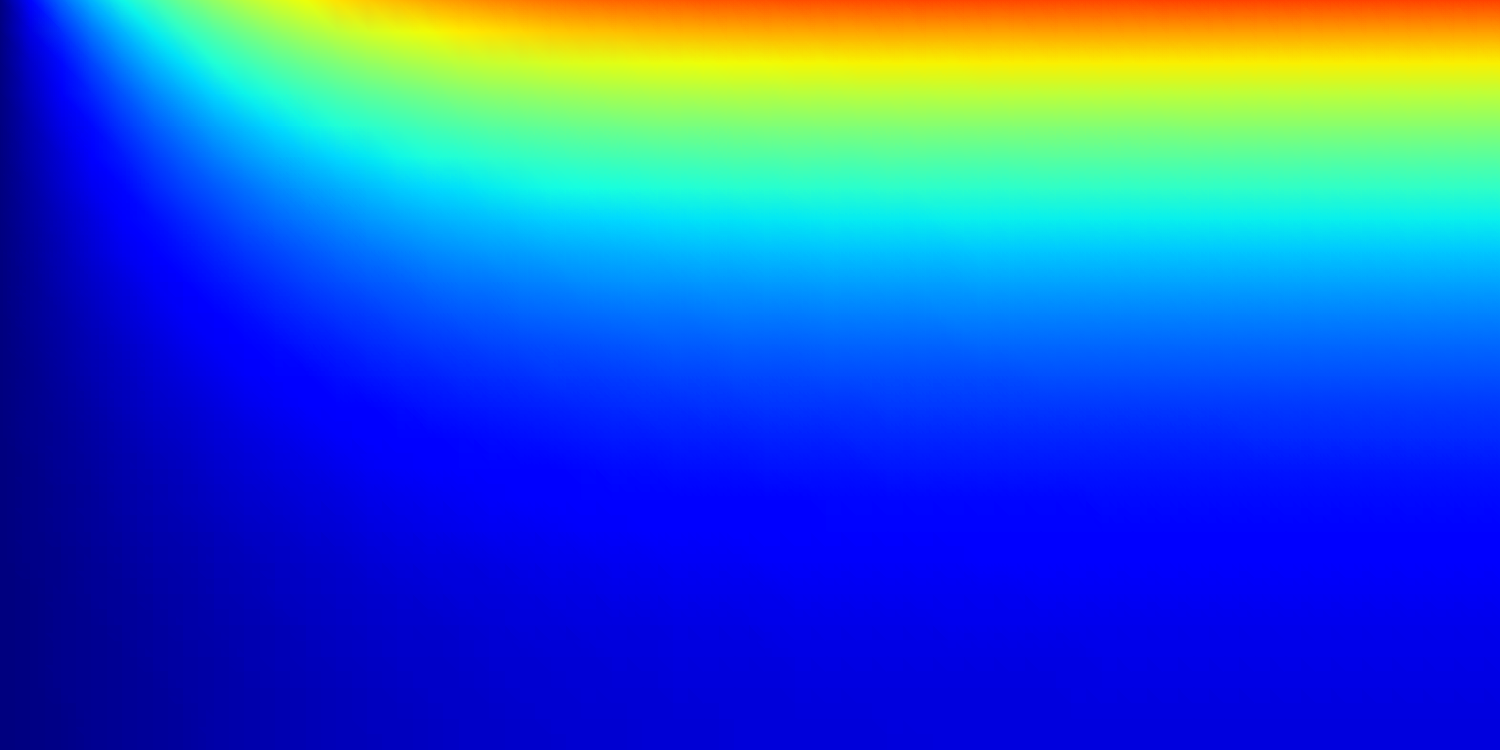}} & {\includegraphics[width=0.19\textwidth]{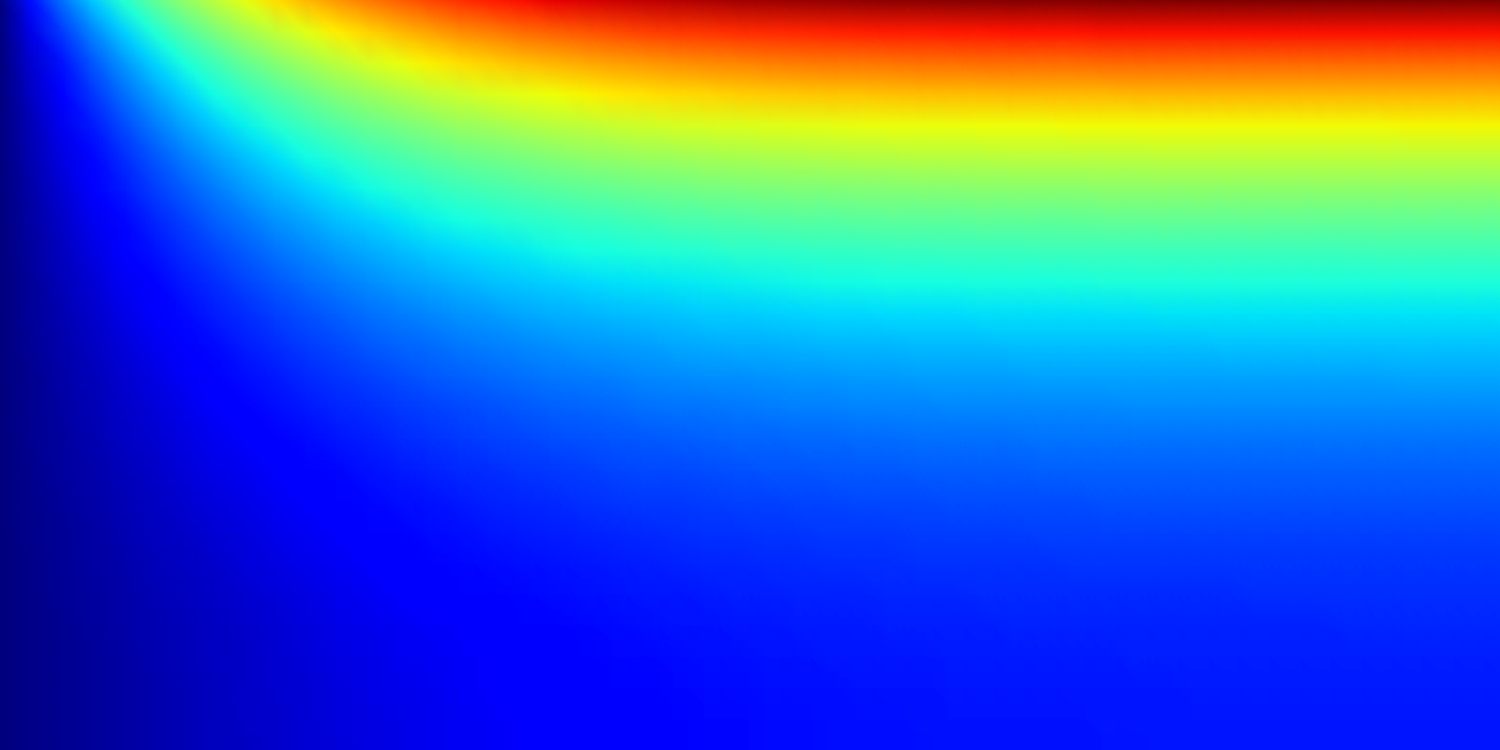}}
\end{tabular}
\begin{subfigure}{\textwidth}
     \includegraphics[width=1.03\textwidth]{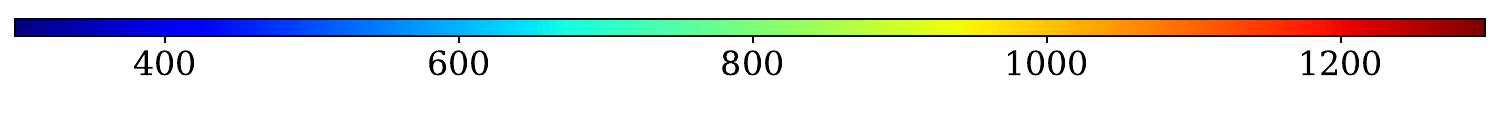}
     \caption*{}
     \label{fig:cbar_all_10}
 \end{subfigure} 
\vspace{-1cm}
\caption{Predicted temperature field for A2 dataset with the training performed on 10\% of data. Our hybrid twin is able to correctly predict the temperature gap for long frame sequences.}
\label{fig:sim_10}
\end{figure}

In Figure \ref{fig:sim_10} we display the changes in the temperature field through time for the model trained on the A2 dataset. We can see that in the simulation, the temperature monotonically increases through time. The hybrid twin correctly predicts the gap throughout the entire simulation. We can see that the original gap between linear and nonlinear simulations increases through time, which makes the prediction of the gap for the last frame in the simulation the most challenging. As a result, and considering that in all the simulations we use for the training the temperature values also monotonically increase, in the following figures we display only the result for the last frame showing the greatest gap correction. The displayed frames were unseen during the training. 

\begin{figure}[ht!]
 \begin{subfigure}{0.5\textwidth}
     \includegraphics[width=\textwidth]{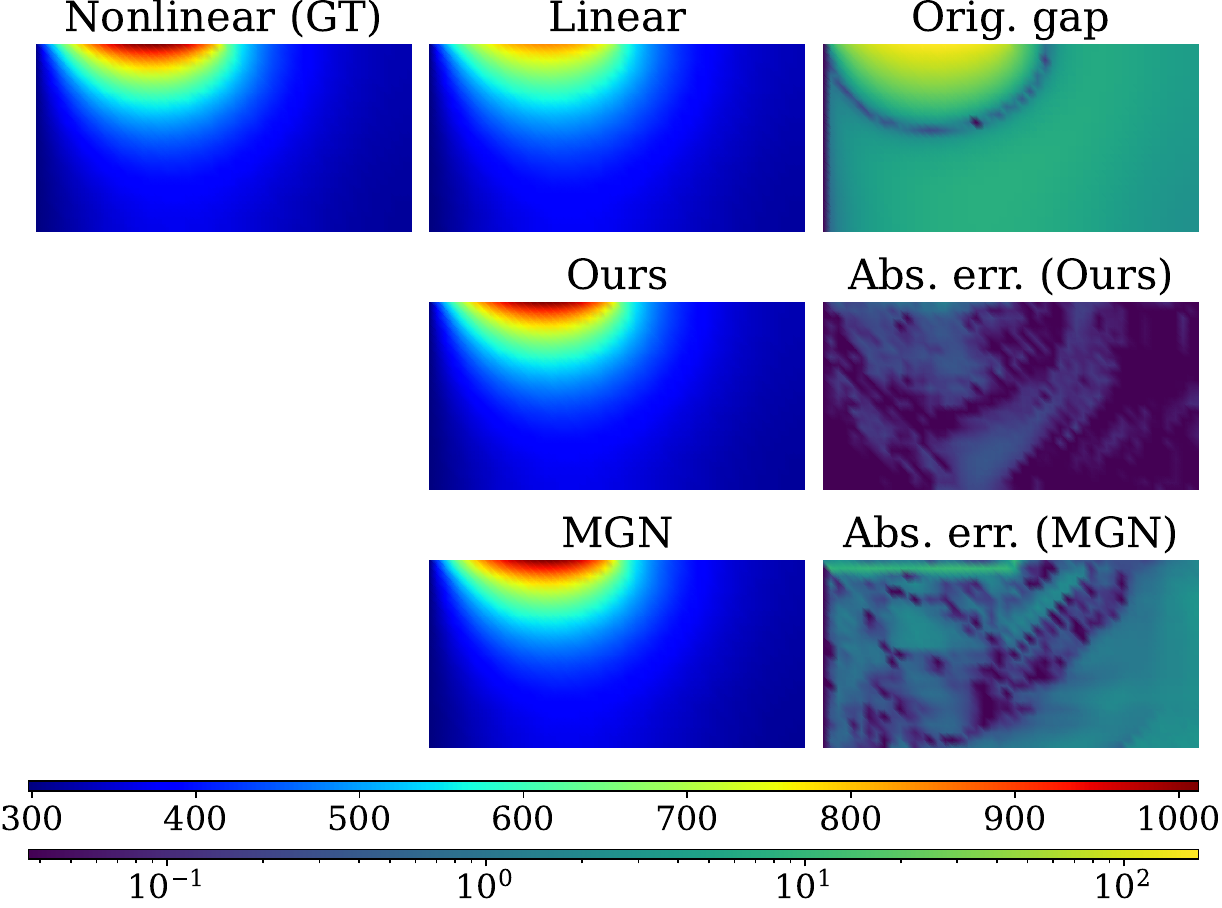}
     \caption{}
     \label{fig:half}
 \end{subfigure}
 \hfill
 \begin{subfigure}{0.48\textwidth}
     \includegraphics[width=\textwidth]{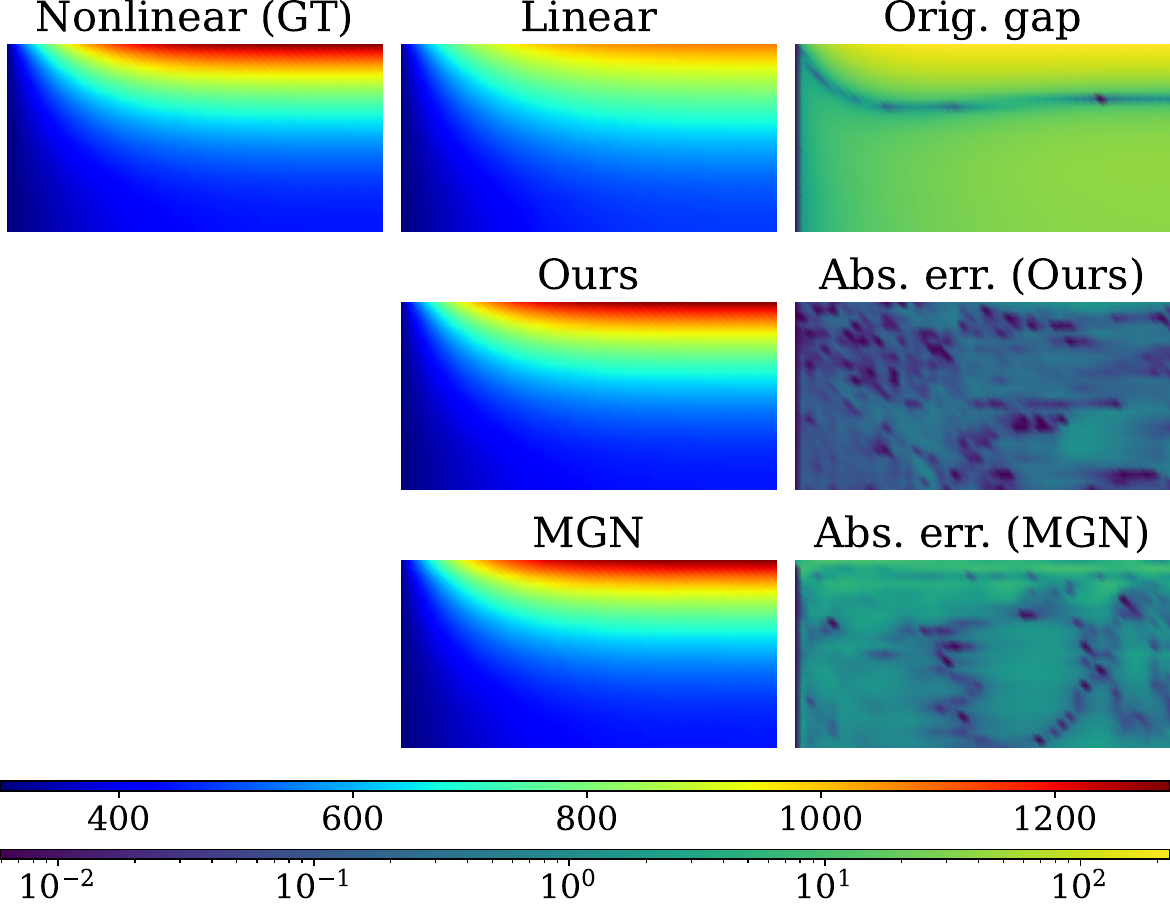}
     \caption{}
     \label{fig:all}
 \end{subfigure} 
\caption{Predicted temperature field for A1 and A2 datasets with the training performed on 10\% of data.}
\label{fig:main_10}
\end{figure}

Figure \ref{fig:main_10} shows the performance of the hybrid twins learnt for problems A1 and A2 respectively.

The error in this and all the subsequent cases is displayed in log scale due to the difference in the magnitude of error for different cases. To eliminate error values that are equal to zero, we replaced them with the lowest nonzero error in the node. The ignorance models are able to predict both seen and unseen frames with the maximum absolute error in the node of a few degrees, which is less than 1\% relative error, while the original maximum error in the node between linear and nonlinear simulations is equal to 16\% for both A1 and A2 datasets. We also compare the performance of the proposed model with the baseline approach, which is an autoregressive MeshGraphNet (MGN) model \cite{pfaff2020learning}. 
An autoregressive neural network predicts the next step frame using the previous step prediction as  the next input. This causes rapid error accumulation over the whole length of the simulation because of the possible noise in each prediction. During the training process, this performance issue is tackled by adding noise to the input of the network at each timestep. We use Gaussian noise with zero mean and the standard deviation that is equal to the RMSE value of an autoregressive model during the training. This technique is usually referred to as noise injection (NI) \cite{sanchez2020learning}.

The MGN model was trained with noise injection (NI) to compensate for error accumulation in the rollout. We can see that the performance of the baseline is lower than the performance of our hybrid twin.

The results are summarized in Table \ref{tab:summary_1}. For each training run of a corresponding use case, the MAE and MAPE results were averaged across all nodes in each frame and across all frames in the simulation. Then the results for all runs were averaged across different seeds.

We can clearly see that the performance of our hybrid twin is higher than that of MGN. 

\begin{table}[ht!]
\centering
\begin{tabular}{p{2.1cm}p{1.5cm}p{3.4cm}p{3.1cm}}
\toprule
Model & Dataset & MAE, $\text{K}$ & MAPE, \% \\
\midrule
Hybrid twin & \multirow{2}{*}{A1 (10\%)} & $\mathbf{(4.86 \pm 0.68) \cdot 10^{-2}}$ & $\mathbf{(1.24 \pm 0.23) \cdot 10^{-2}}$  \\
MGN+NI &  & $19.19 \pm 31.63$ & $4.64 \pm 7.66$  \\
Hybrid twin & \multirow{2}{*}{A2 (10\%)} & $\mathbf{(10.34 \pm 0.76) \cdot 10^{-2}}$ & $\mathbf{(22.93 \pm 0.86) \cdot 10^{-3}}$  \\
MGN+NI &  & $1.01 \pm 0.15$ & $0.20 \pm 0.02$  \\
\bottomrule
\end{tabular}
\vspace{0.2cm}
\caption{Results of the use cases with a reduced number of frames for training.}
\label{tab:summary_1}
\end{table}

\subsection{Mesh generalization evaluation}

The goal of the next test case is to assess the mesh generalization capabilities of our method. We take the model that was trained on A1 and A2 datasets with the regular mesh on 50\% of randomly selected samples, and evaluate the performance of this model on the datasets A3 and A4 with an irregular mesh. 

\begin{figure}[ht!]
 \begin{subfigure}{0.5\textwidth}
     \includegraphics[width=\textwidth]{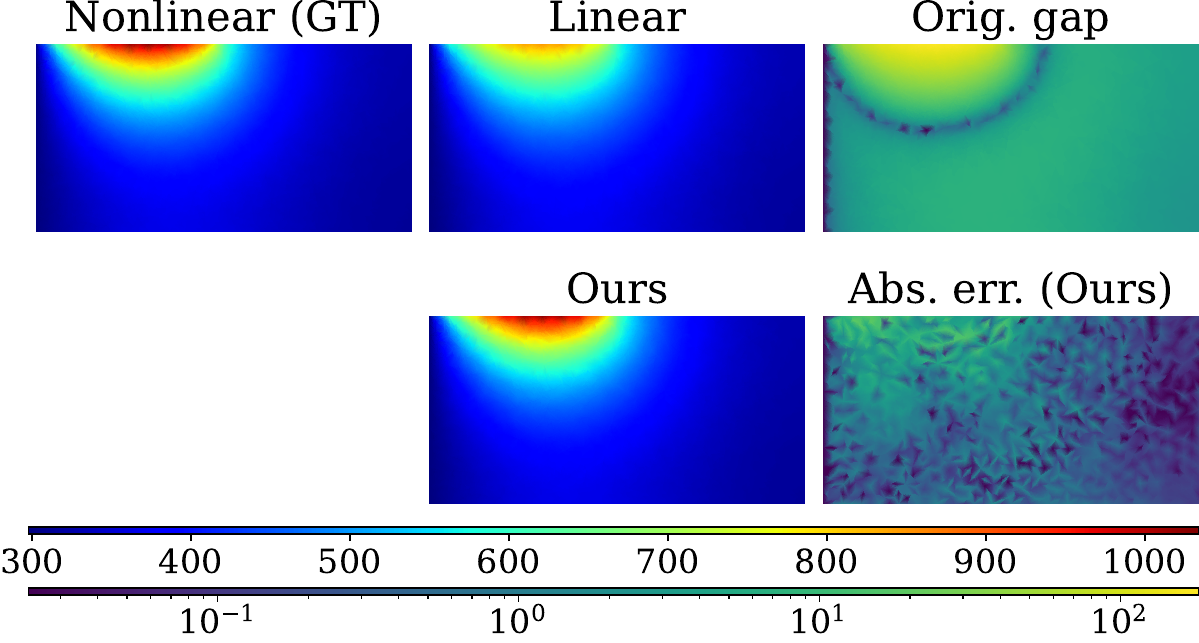}
     \caption{}
     \label{fig:half_irreg}
 \end{subfigure}
 \hfill
 \begin{subfigure}{0.49\textwidth}
     \includegraphics[width=\textwidth]{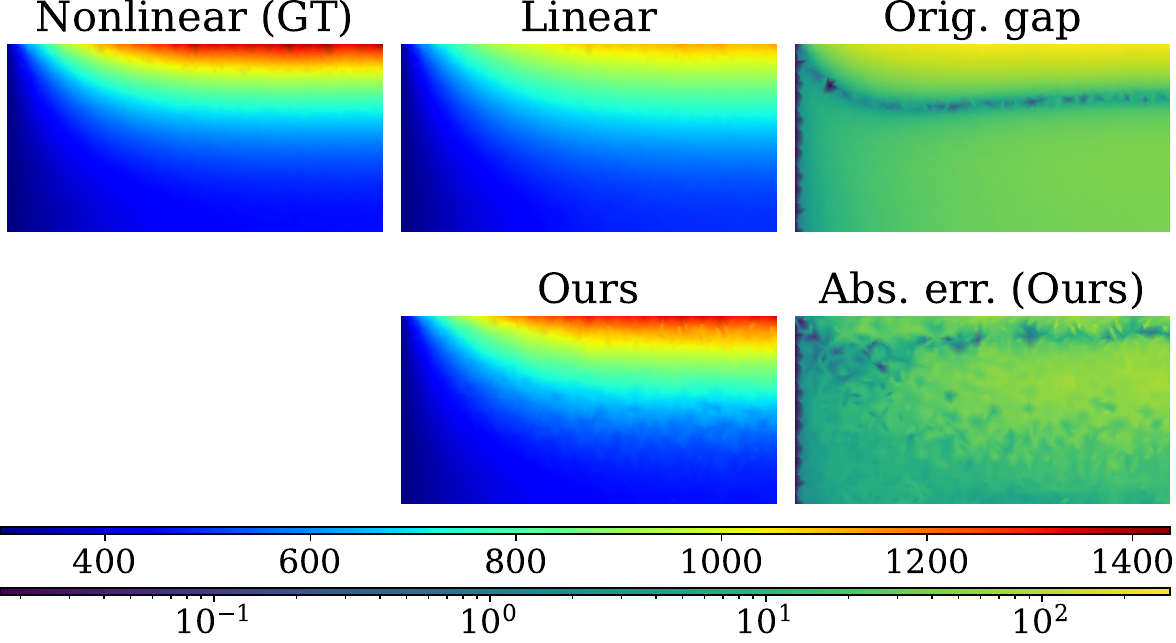}
     \caption{}
     \label{fig:all_irreg}
 \end{subfigure} 

 \caption{Prediction of the simulation on an irregular mesh with the training performed on the regular mesh: (a) dataset A3; (b) dataset A4.}
 \label{fig:irreg}
\end{figure}

In Figure \ref{fig:irreg} we can see that the error value in this case is higher compared to the result of the training with the regular mesh. However, the original maximum relative error in the node was decreased from 16\% to 3\% for the dataset A3 and from 16\% to 7\% for the dataset A4.

For the training of all our GNNs, we decided to select mean aggregation based on the experiment with different aggregation functions for the models trained on 50\% of samples in the datasets A1 and A2. The mean aggregation during hidden feature update introduces a normalization on the number of the neighbors of each node. Consequently, this should improve the model generalization to the geometries with different mesh connectivities. In our case, the regular mesh has 4 neighbors for the majority of the nodes, which is not true for the irregular mesh.

\begin{figure}[ht!]
 \begin{subfigure}{0.5\textwidth}
     \includegraphics[width=\textwidth]{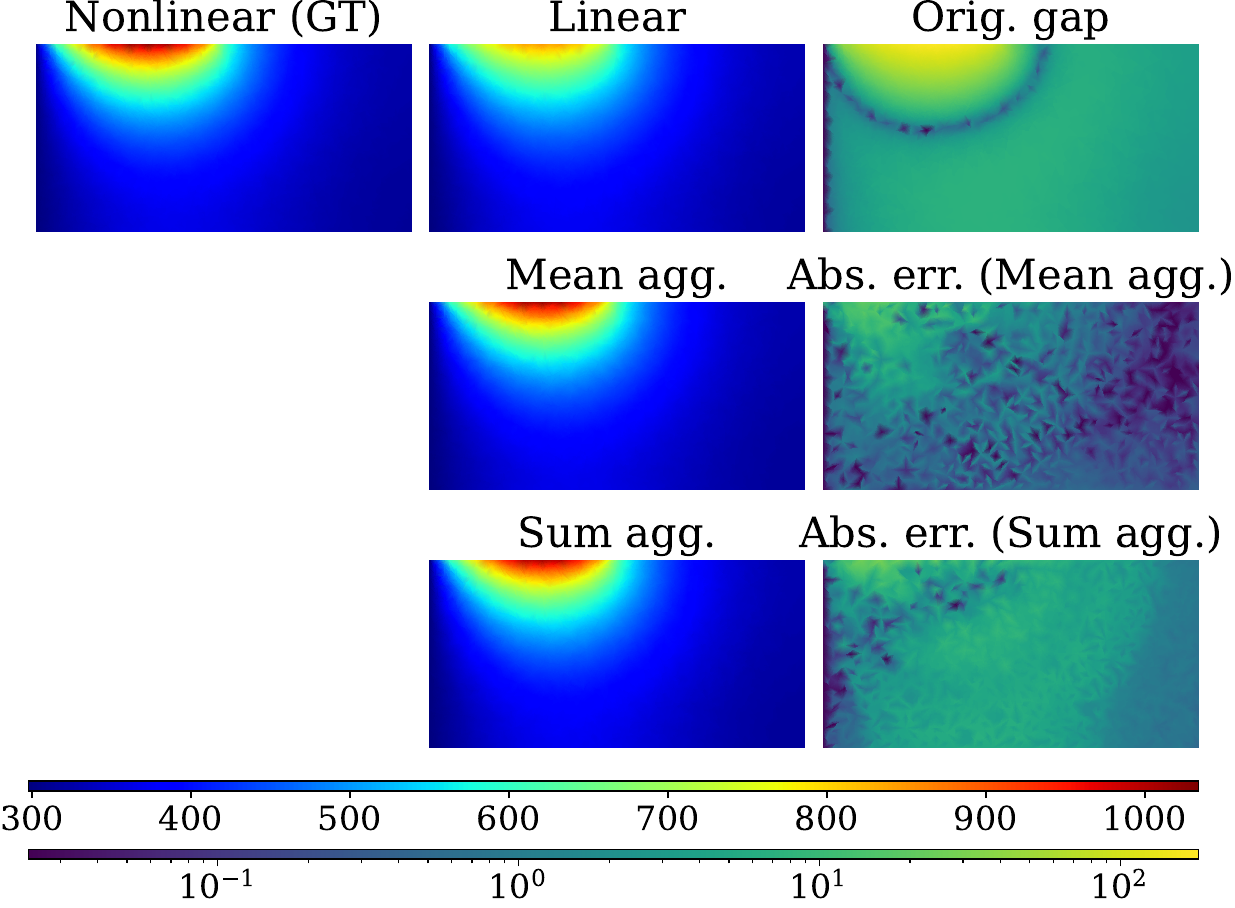}
     \caption{}
     \label{fig:half_irreg_sum}
 \end{subfigure}
 \hfill
 \begin{subfigure}{0.49\textwidth}
     \includegraphics[width=\textwidth]{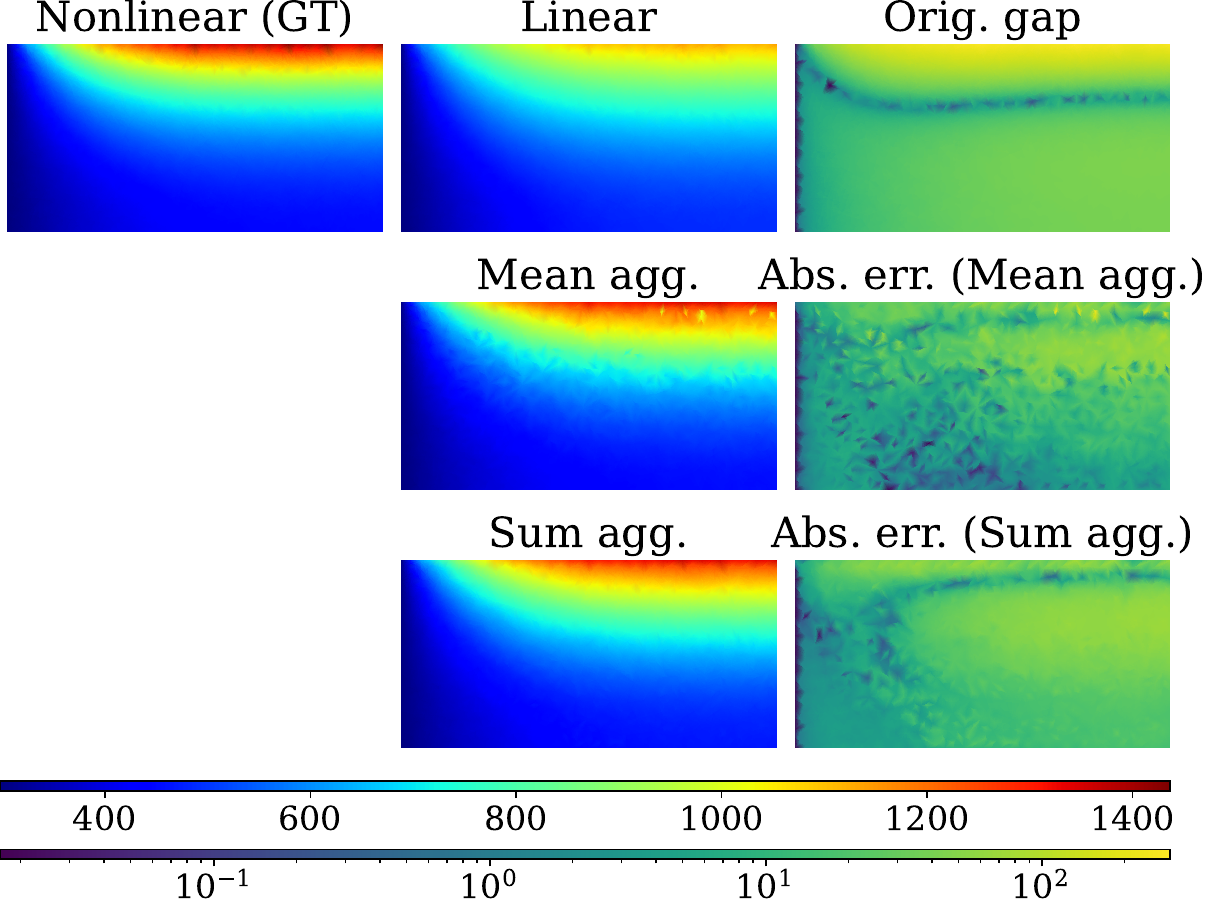}
     \caption{}
     \label{fig:all_irreg_sum_}
 \end{subfigure} 

 \caption{Comparison of the result of sum and mean aggregation for the prediction of the simulation on an irregular mesh: (a) dataset A3; (b) dataset A4.}
 \label{fig:irreg_sum}
\end{figure}

In Figure \ref{fig:irreg_sum} we compare the result of mean and sum aggregation function in the architecture for the case where the model was trained on 50\% of data. We can see that on average the absolute error for sum aggregation is higher that the one of mean aggregation. These results are summarized in Table \ref{tab:mean_sum}, although the discrepancy is not drastically high.

\begin{table}[ht!]
\begin{center}
\begin{tabular}{p{2.3cm}llll}
\toprule
Model\linebreak aggregation & Dataset & MAE, $\text{K}$ & MAPE, \%  \\
\midrule
Mean & \multirow{2}{*}{A1 (50 \%)} & $(72.37 \pm 0.80) \cdot 10^{-2}$ & $(15.16 \pm 0.12) \cdot 10^{-2}$  \\
Sum &  & $1.52 \pm 0.22$ & $0.36 \pm 0.05$ \\
\midrule
Mean & \multirow{2}{*}{A2 (50 \%)} & $7.31 \pm 1.25$ & $1.25 \pm 0.27$ \\
Sum &  & $7.72 \pm 1.44$ & $1.38 \pm 0.31$ \\
\bottomrule
\end{tabular}
\caption{Comparison of a performance on an irregular dataset of the models trained wit sum and mean aggregation function}
\label{tab:mean_sum}    
\end{center}
\end{table}

\subsection{Training with a scarce number of spatial nodes}

In the next use case, we extract a submesh of an irregular mesh and see if the model is capable of reaching the same performance if trained on a limited number of mesh nodes (datasets A5-A8). 

\begin{figure}[ht!]
 \begin{subfigure}{0.5\textwidth}
     \includegraphics[width=\textwidth]{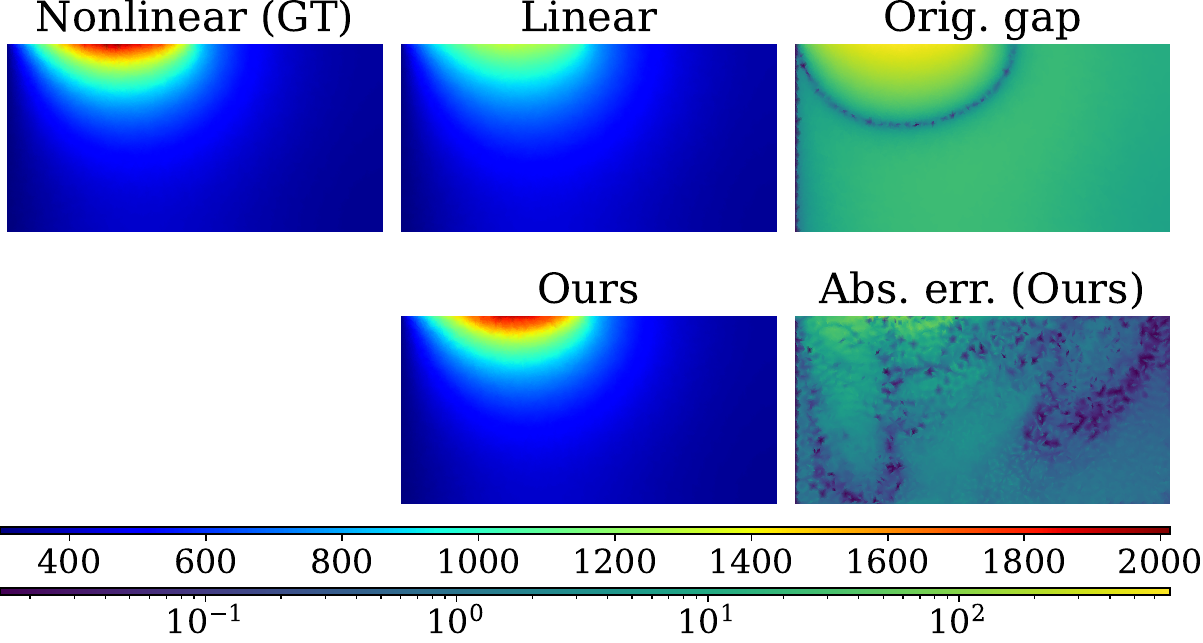}
     \caption{}
     \label{fig:half_submesh}
 \end{subfigure}
 \hfill
 \begin{subfigure}{0.5\textwidth}
     \includegraphics[width=\textwidth]{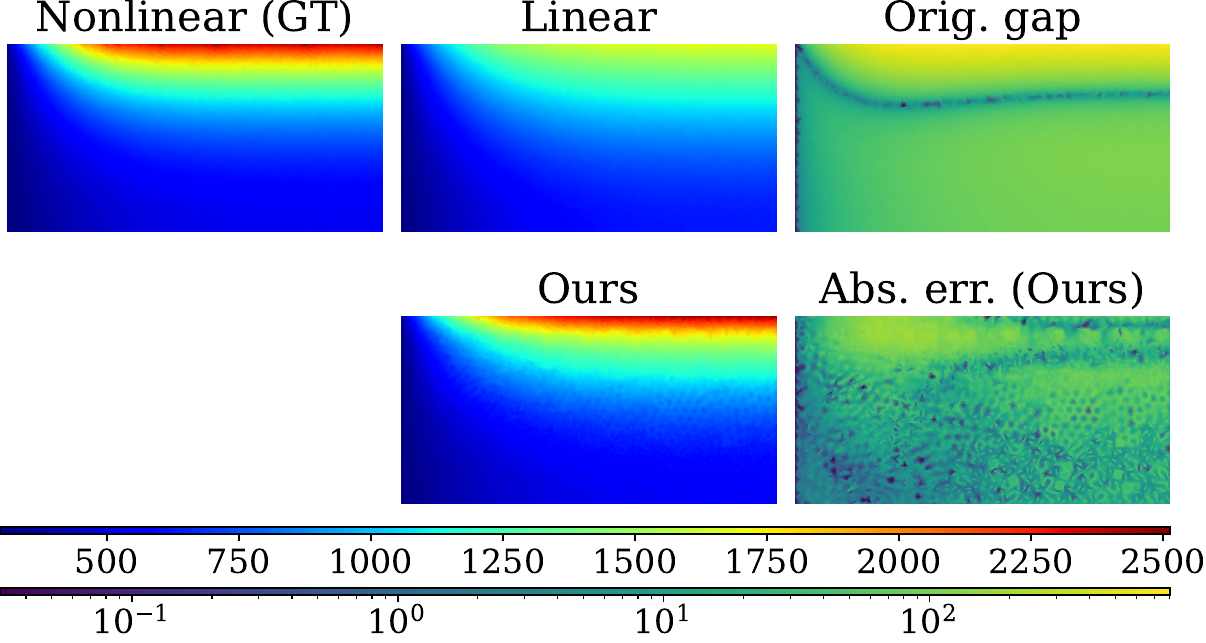}
     \caption{}
     \label{fig:all_submesh}
 \end{subfigure} 
 \caption{The prediction of the simulation on an original mesh with the corresponding training on the submesh: (a) dataset A5 (submesh A7); (b) dataset A6 (submesh A8).}
 \label{fig:res_submesh}
\end{figure}

In Figure \ref{fig:res_submesh} we can see the result of the evaluation. The original maximum error in the node between linear and nonlinear frames is 16\% for dataset A5 and 20\% for dataset A6. After applying the ignorance model, the maximum relative error in the node decreases to 5\% and 13\% respectively.

This result can be useful in scenarios when the ground truth data of the process comes from a setting with a scarce number of sensors across the domain. In this case, the model should be capable of learning the gap in the entire domain using the data from these sensors.

The results of the two mesh generalization use cases are summarized in Table \ref{tab:summary_2}. As can be seen, the use of only a few sensors can be sufficient for learning an ignorance model given the generalization capabilities of GNNs.

\begin{table}[ht!]
\begin{center}
\begin{tabular}{p{2cm}llll}
\toprule
Model & Dataset & MAE, $\text{K}$ & MAPE, \%\\
\midrule
\multirow{2}{*}{Hybrid twin} & A3 & $(72.37 \pm 0.80) \cdot 10^{-2}$ & $(15.16 \pm 0.12) \cdot 10^{-2}$  \\
 & A4 & $7.31 \pm 1.25$ & $1.25 \pm 0.27$  \\
\midrule
\multirow{2}{*}{Hybrid twin} & A5 & $2.13 \pm 0.93$ & $0.37 \pm 0.19$  \\
 & A6 & $15.90 \pm 1.06$ & $1.96 \pm 0.19$  \\
\bottomrule
\end{tabular}    
\vspace{0.2cm}
\caption{Results of the mesh generalization test cases}
\label{tab:summary_2}
\end{center}
\end{table}

\subsection{Generalization over load positions}\label{sec:gen_load}

In the following use case, we demonstrate that the model is capable of generalizing between different load positions. We train an ignorance model on the dataset B1 with a normally distributed heat source load. Each simulation has a different position of the Gaussian center on a plate. 40 simulations are used for the training, and the remaining 10 simulations are used for testing.

\begin{figure}[ht!]
 \begin{subfigure}{0.5\textwidth}
     \includegraphics[width=\textwidth]{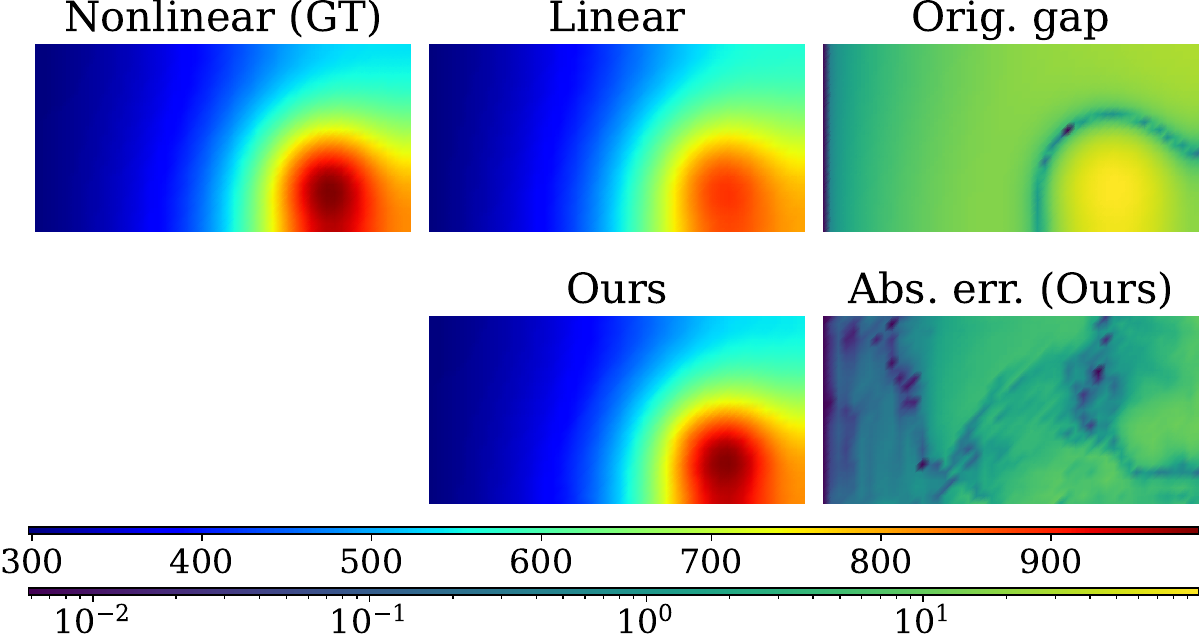}
     \caption{}
     \label{fig:gauss_d2}
 \end{subfigure}
 \hfill
 \begin{subfigure}{0.5\textwidth}
     \includegraphics[width=\textwidth]{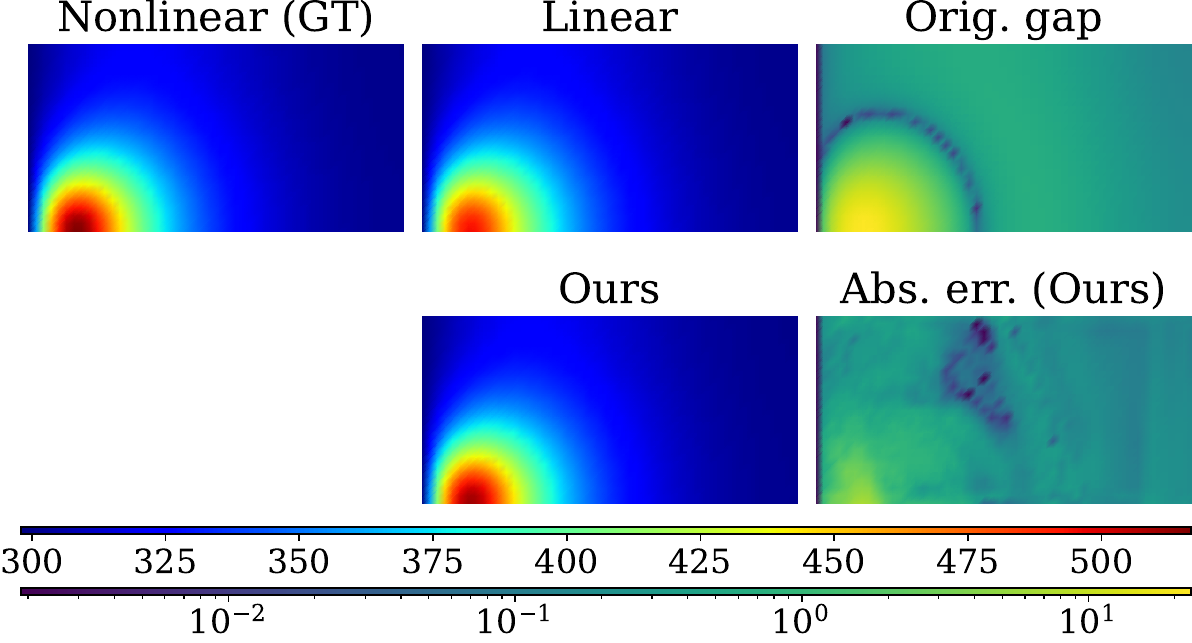}
     \caption{}
     \label{fig:gauss_d5}
 \end{subfigure} 
 \caption{Predicted temperature field for data with a Gaussian heat source load on 2 distribution center coordinates that were unseen during training. Our approach is able to generalize over different load positions.}
 \label{fig:gauss_unseen}
\end{figure}

In Figure \ref{fig:gauss_unseen} we can see the inference for two test load positions (the results for other load positions are described in Appendix \ref{appendix:gauss_all}).

The original maximum relative error in the node equal to $10\%$ was reduced to under $4\%$ for each unseen load position. This use case shows that our model is also able to generalize in cases when we vary the location of the heat source. The result for the rest of the unseen load positions can be found in  \ref{appendix:gauss_all}.

\subsection{Generalization over domain shapes}

In this use case, we check the model generalization over different domain geometries, L-Shaped,  included in dataset B2. This verification implies the evaluation of the twin to generalize to different geometries and meshes simultaneously. The ignorance model learns from 4 geometries, and it is tested on 4 previously unseen geometries. The maximum of the original relative difference between linear and nonlinear simulations in the node is equal to $30\%$ in each of the following cases.

\begin{figure}[ht!]
\begin{tabular}{c>{\centering\arraybackslash}p{2.5cm}>{\centering\arraybackslash}p{2.5cm}>{\centering\arraybackslash}p{2.5cm}>{\centering\arraybackslash}p{2.7cm}}
& \multicolumn{4}{c}{Unseen designs for domain shape generalization} \\ \hline
& $a = b = 0.8$ & $a = b = 0.5$ & $a = 0.4, b = 1.2$ & $a = 1.2, b = 0.4$ \\ \hline\hline

\rotatebox[origin=c]{90}{\parbox[c]{-2cm}{Linear}} & 
{\includegraphics[width=0.23\textwidth]{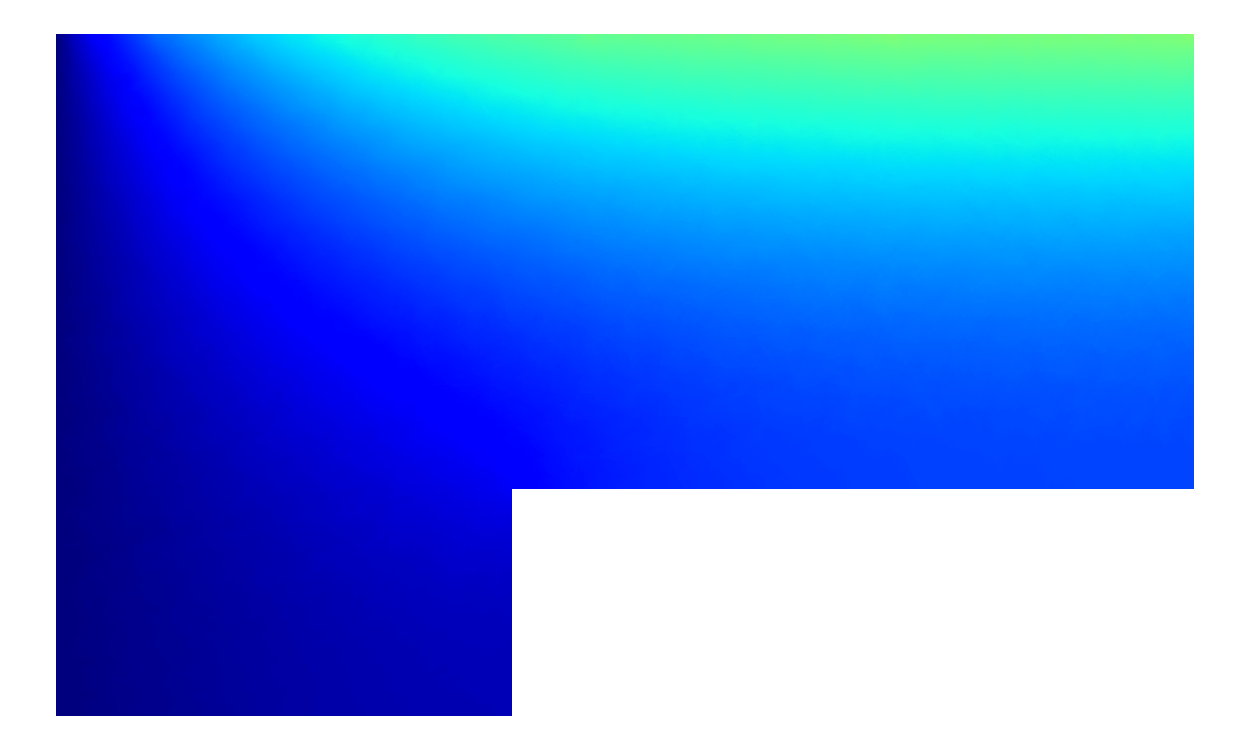}} & {\includegraphics[width=0.23\textwidth]{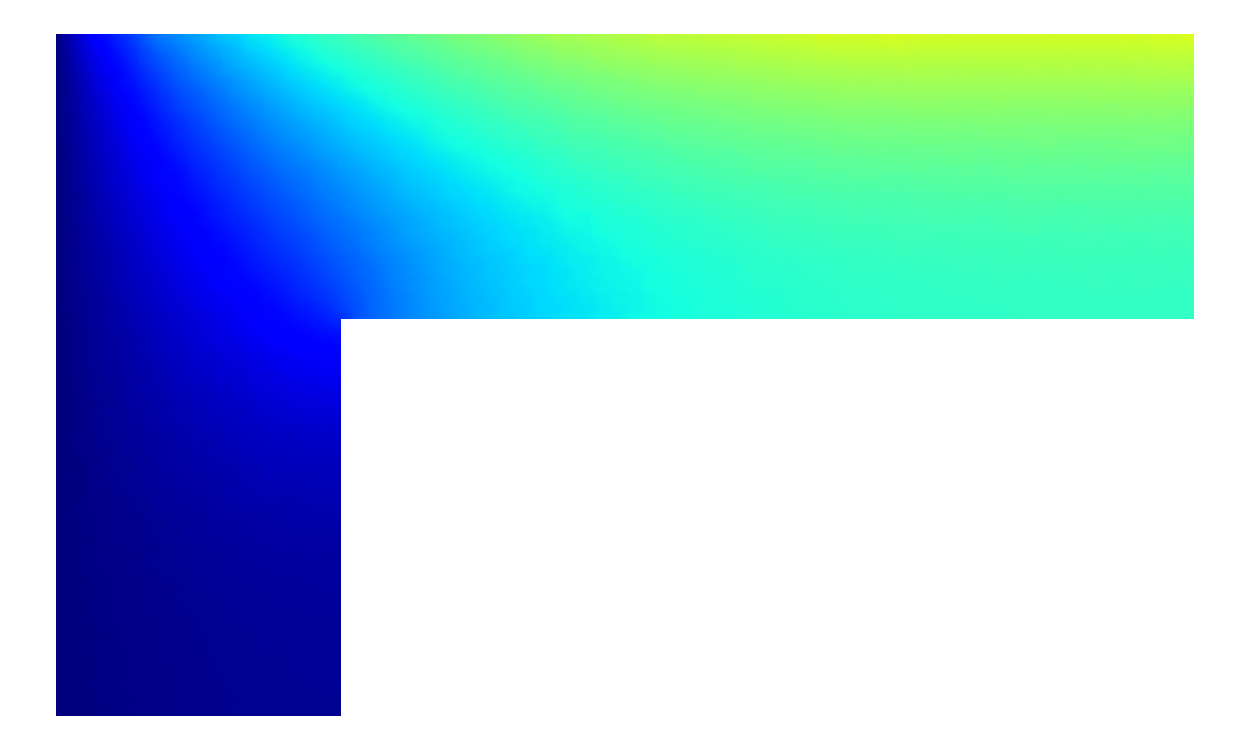}} & {\includegraphics[width=0.23\textwidth]{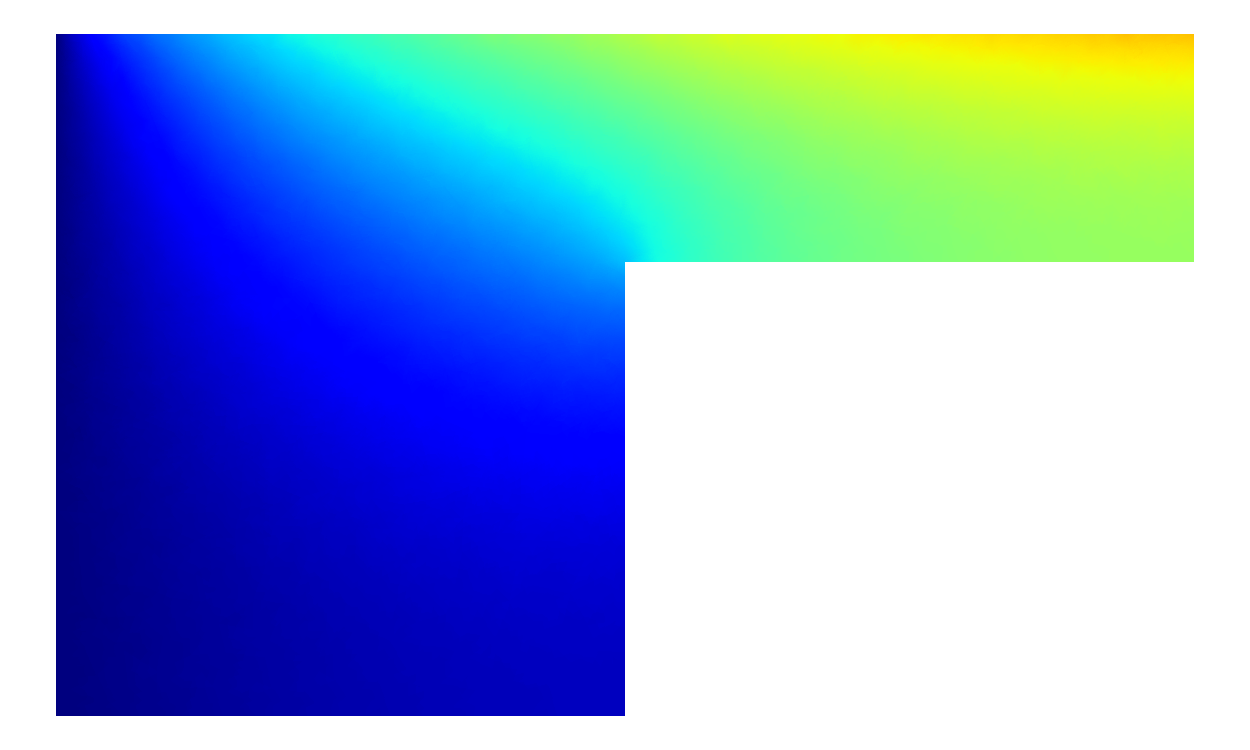}} & {\includegraphics[width=0.23\textwidth]{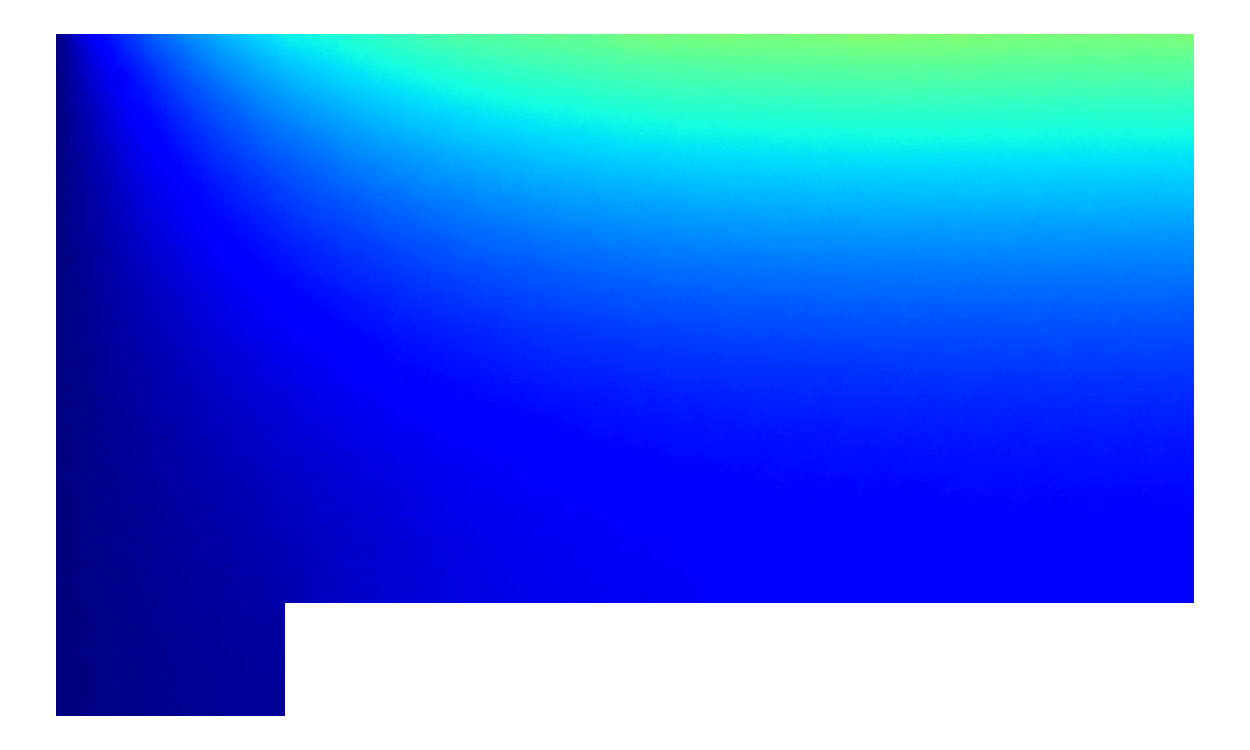}} \\
\rotatebox[origin=c]{90}{\parbox[c]{-2cm}{GT}} & {\includegraphics[width=0.23\textwidth]{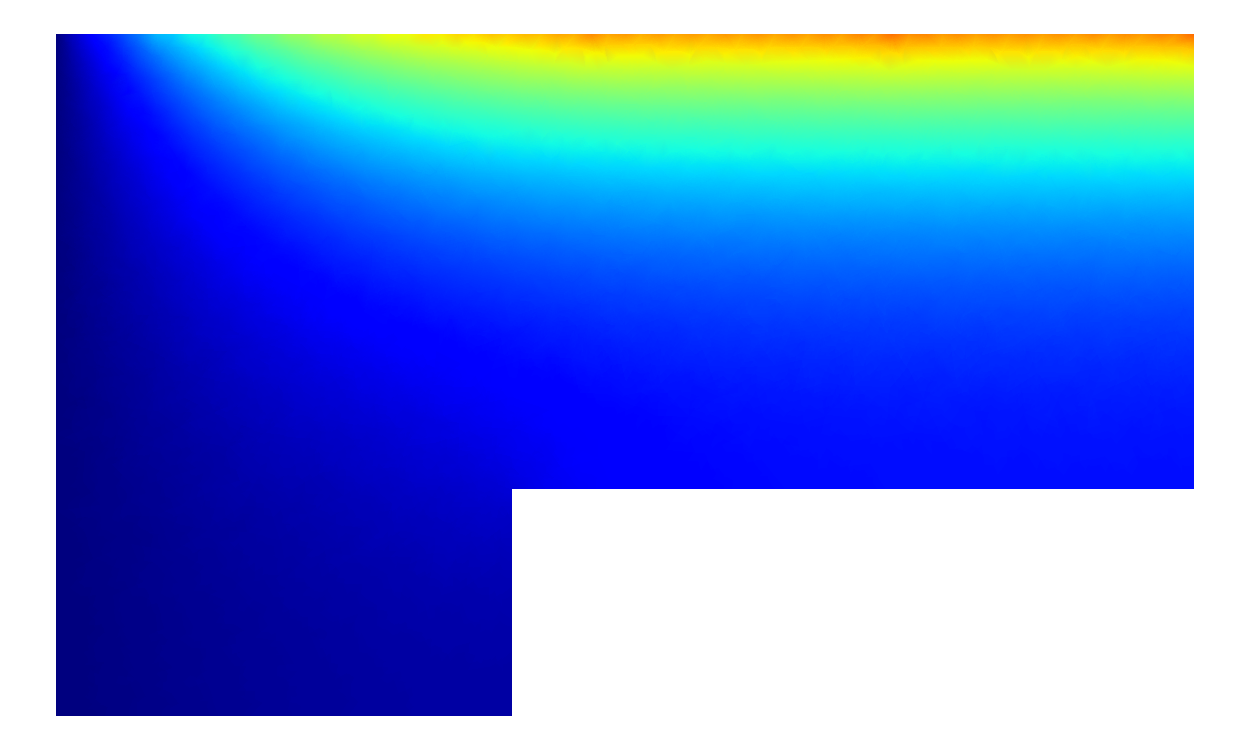}} &
{\includegraphics[width=0.23\textwidth]{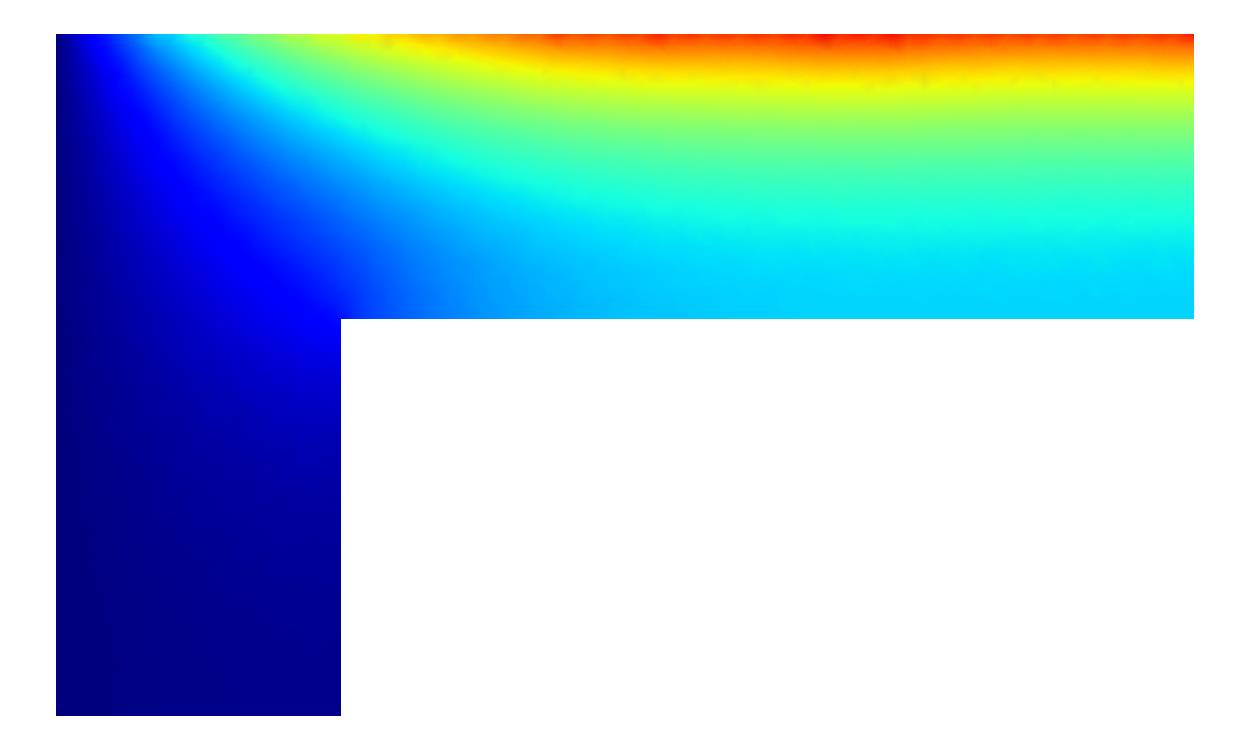}} & {\includegraphics[width=0.23\textwidth]{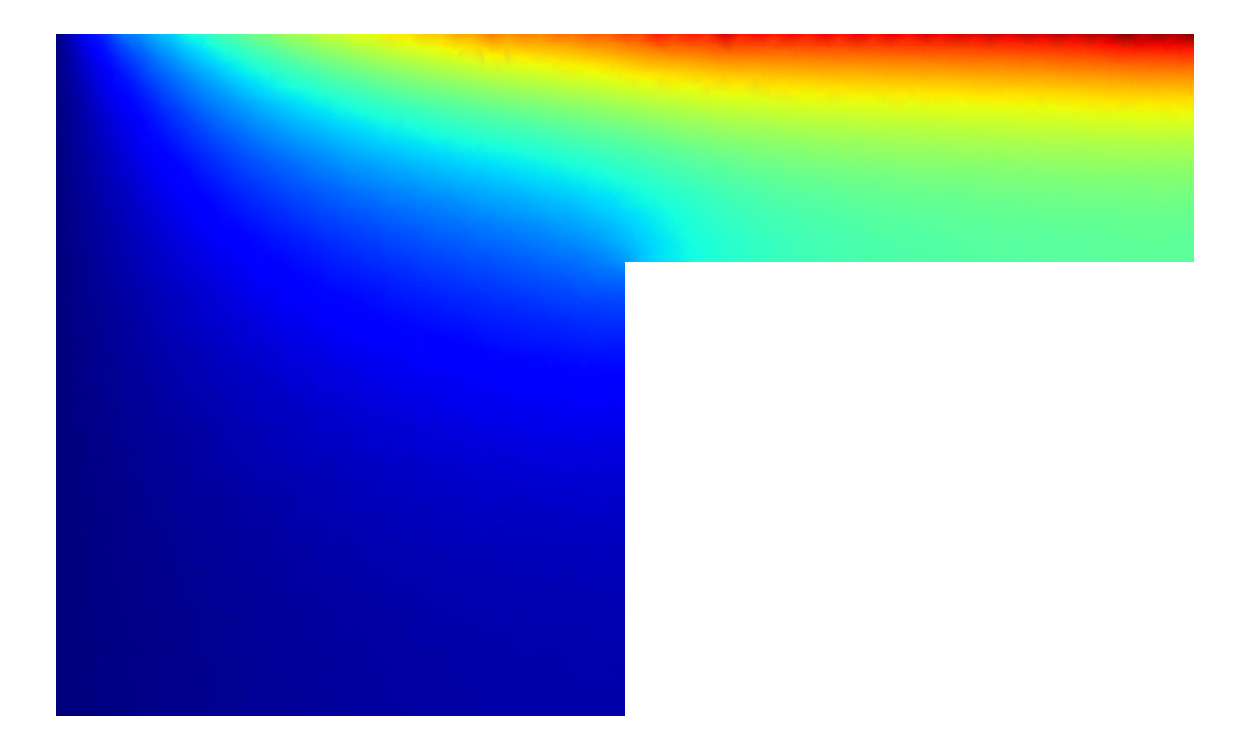}} & {\includegraphics[width=0.23\textwidth]{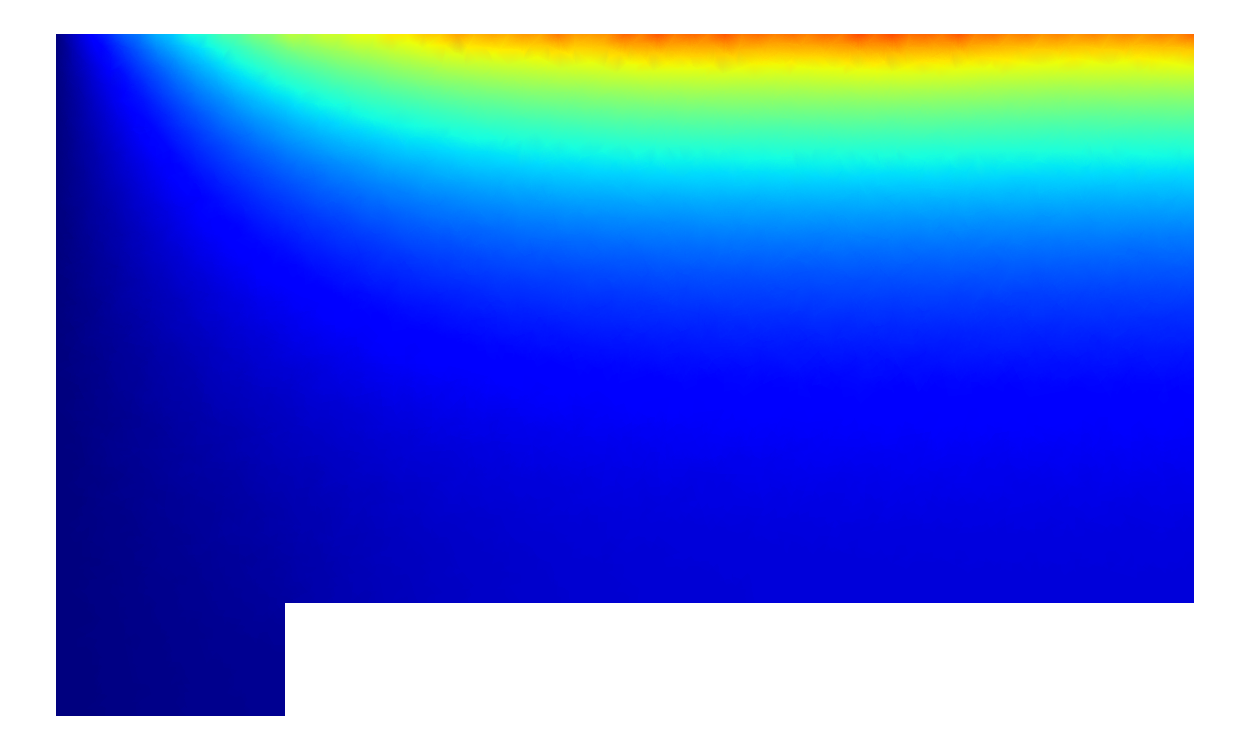}} \\
\rotatebox[origin=c]{90}{\parbox[c]{-2cm}{HT}} & {\includegraphics[width=0.23\textwidth]{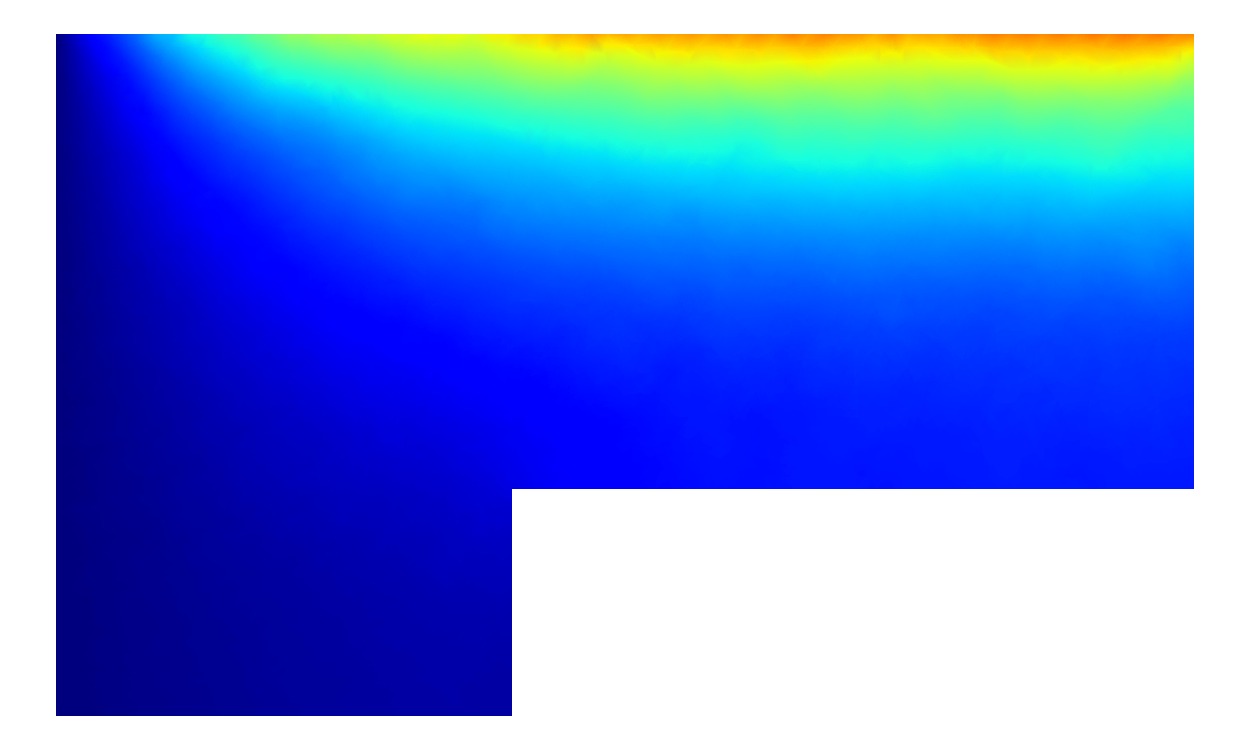}} & {\includegraphics[width=0.23\textwidth]{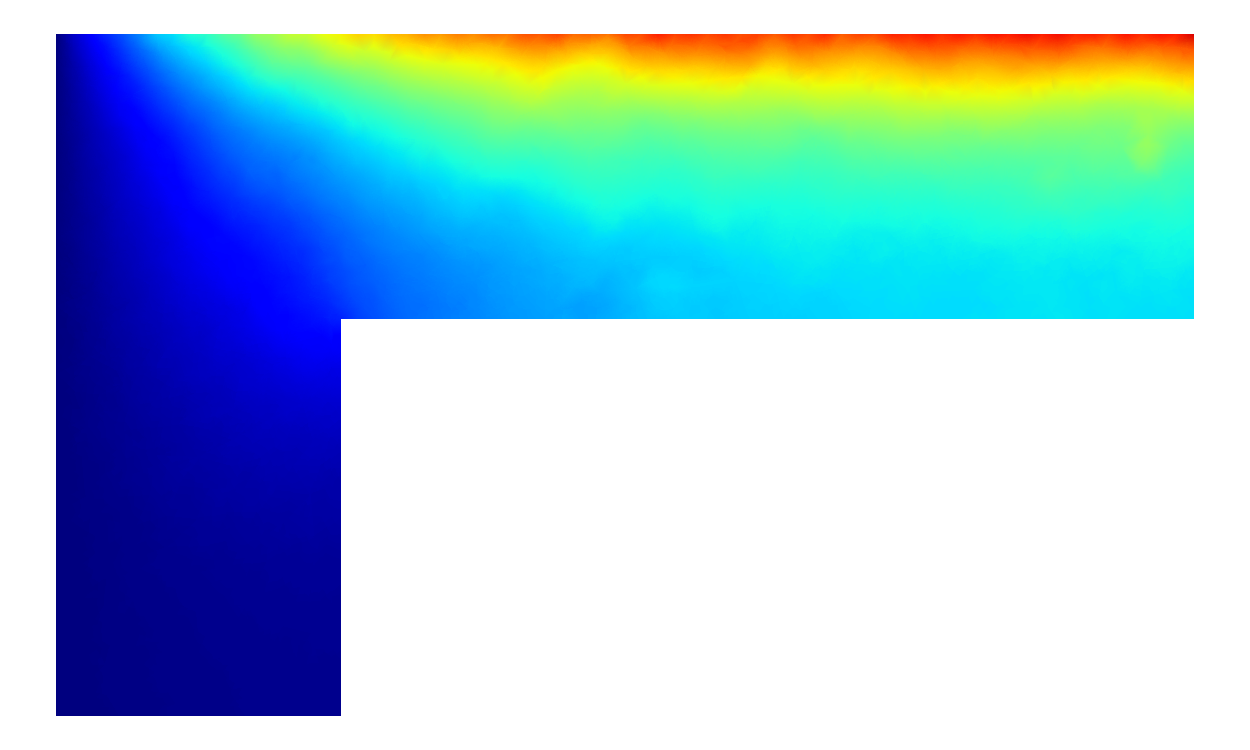}} & {\includegraphics[width=0.23\textwidth]{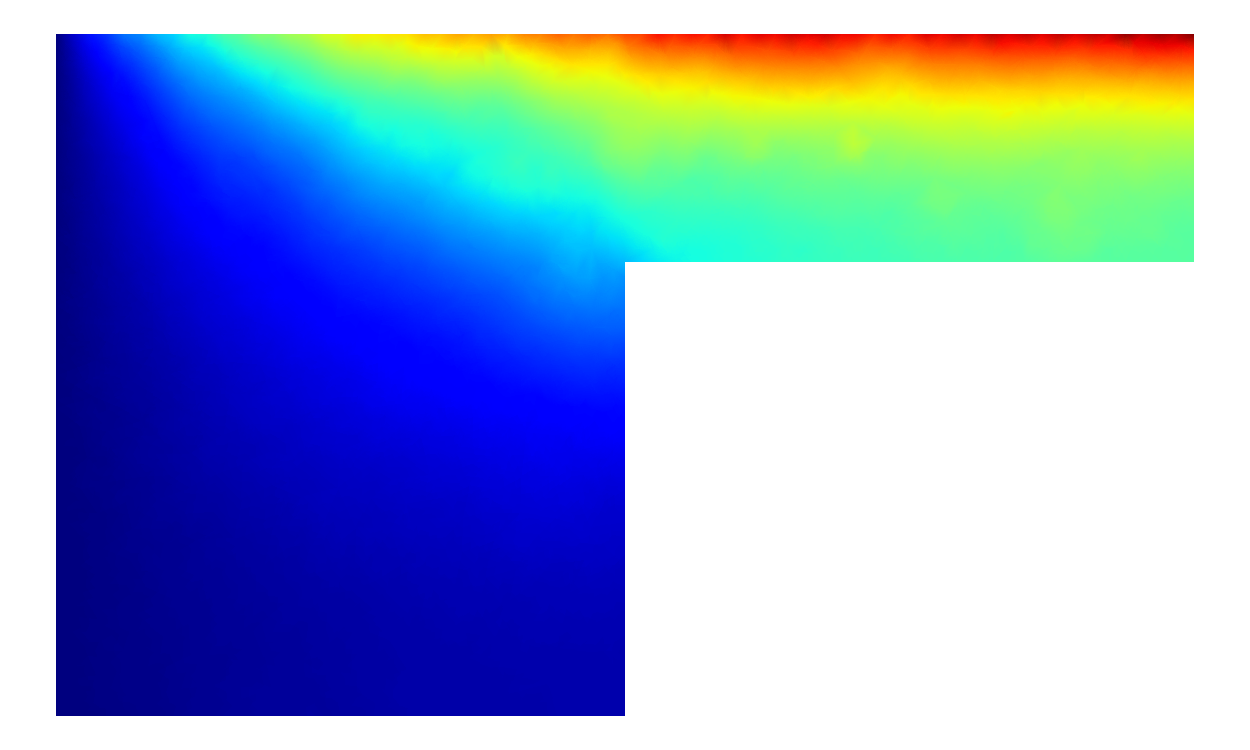}} & {\includegraphics[width=0.23\textwidth]{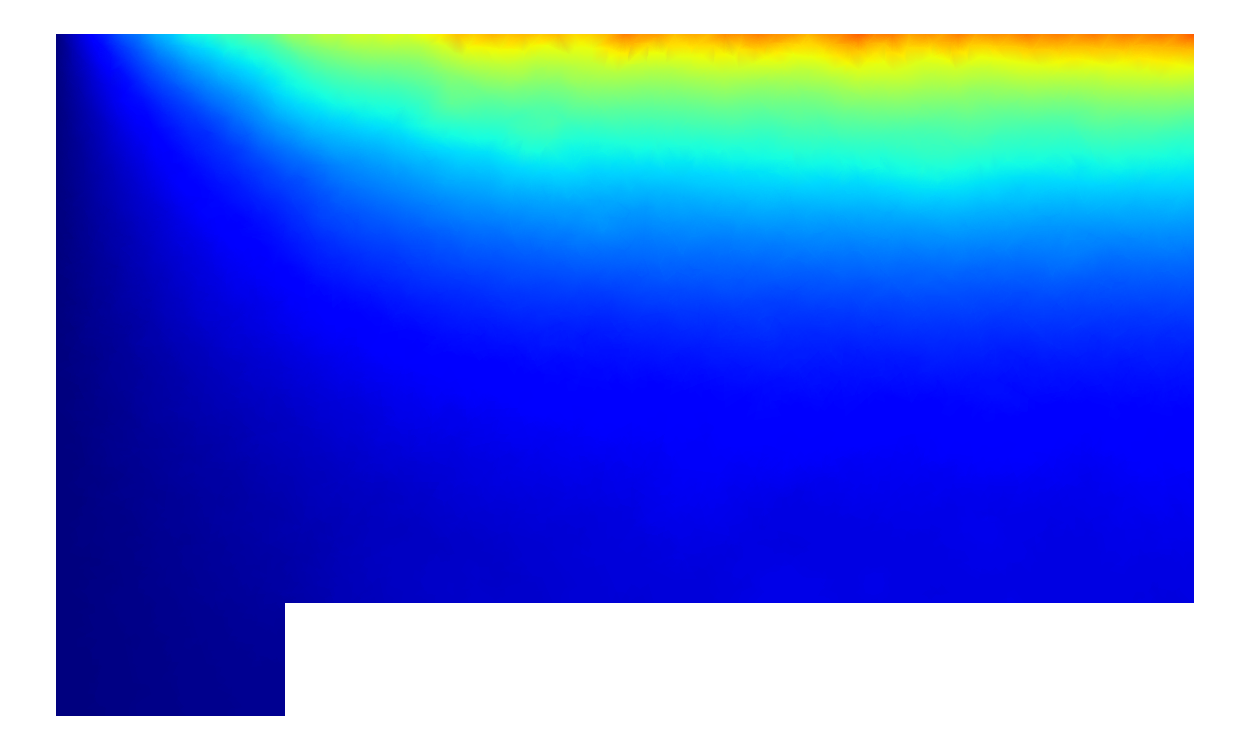}} \\
\rotatebox[origin=c]{90}{\parbox[c]{-2cm}{Orig. gap}} & {\includegraphics[width=0.23\textwidth]{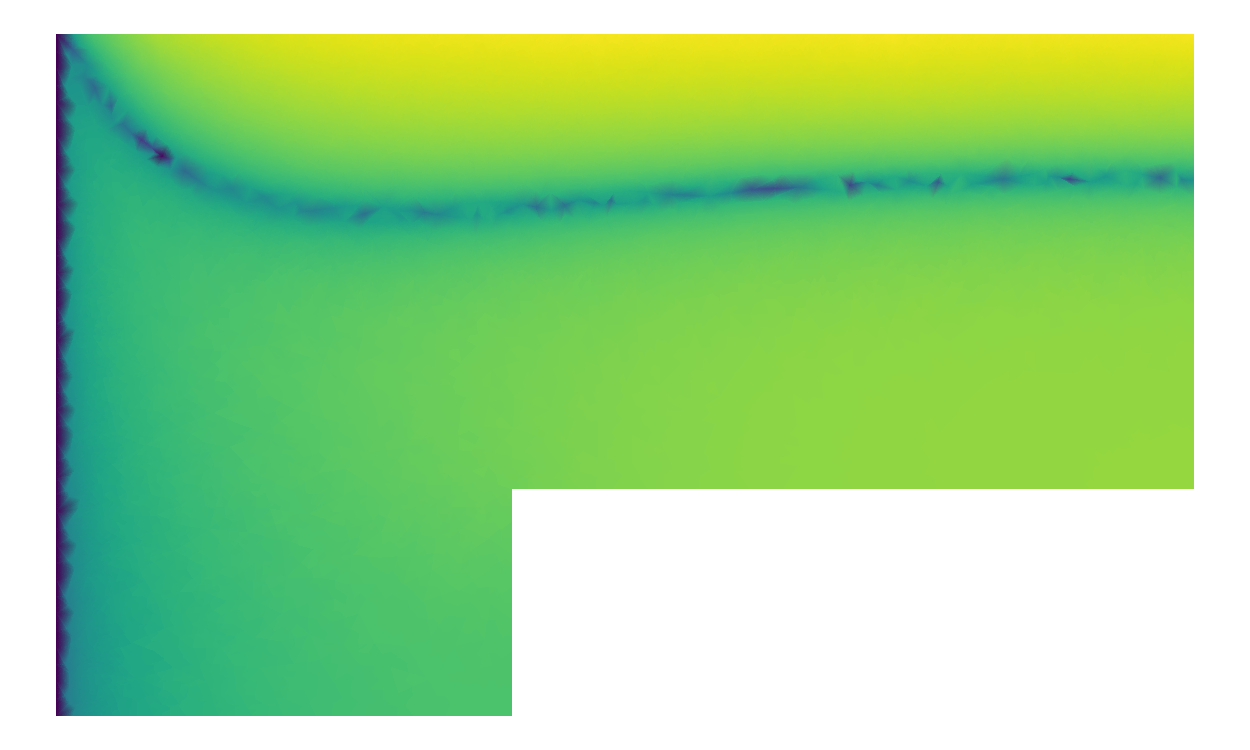}} & {\includegraphics[width=0.23\textwidth]{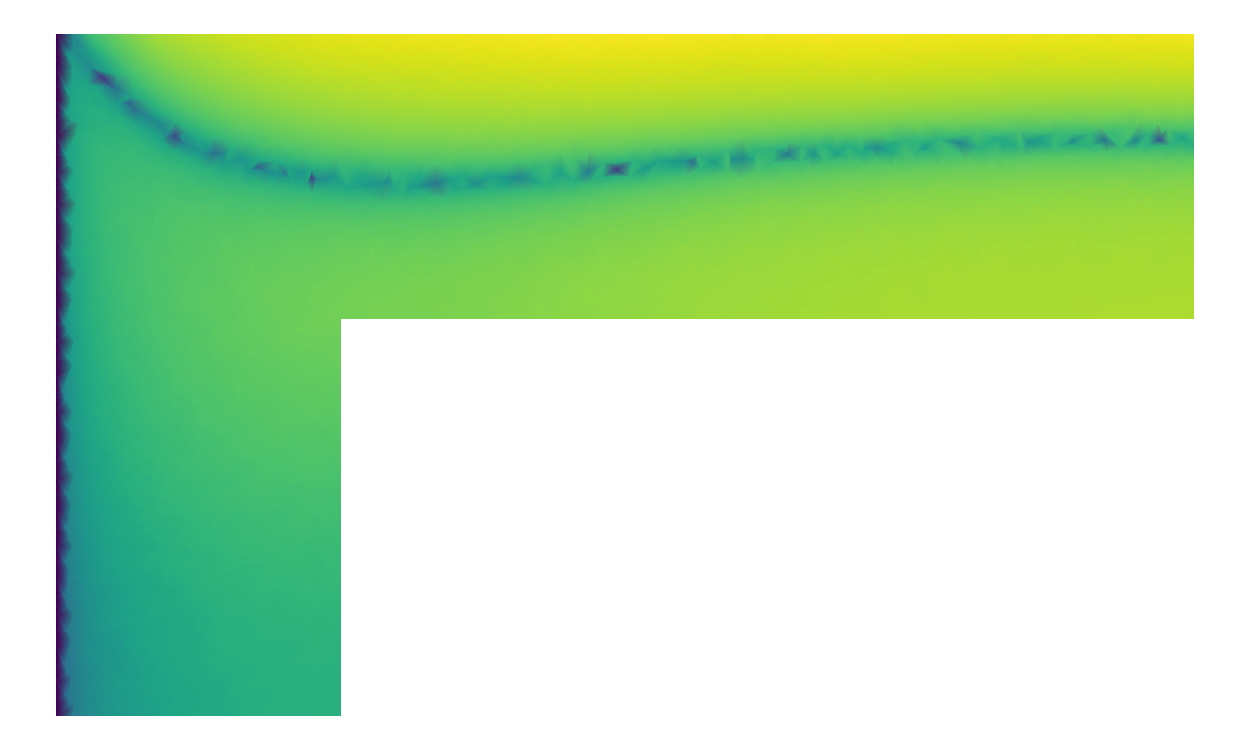}} & {\includegraphics[width=0.23\textwidth]{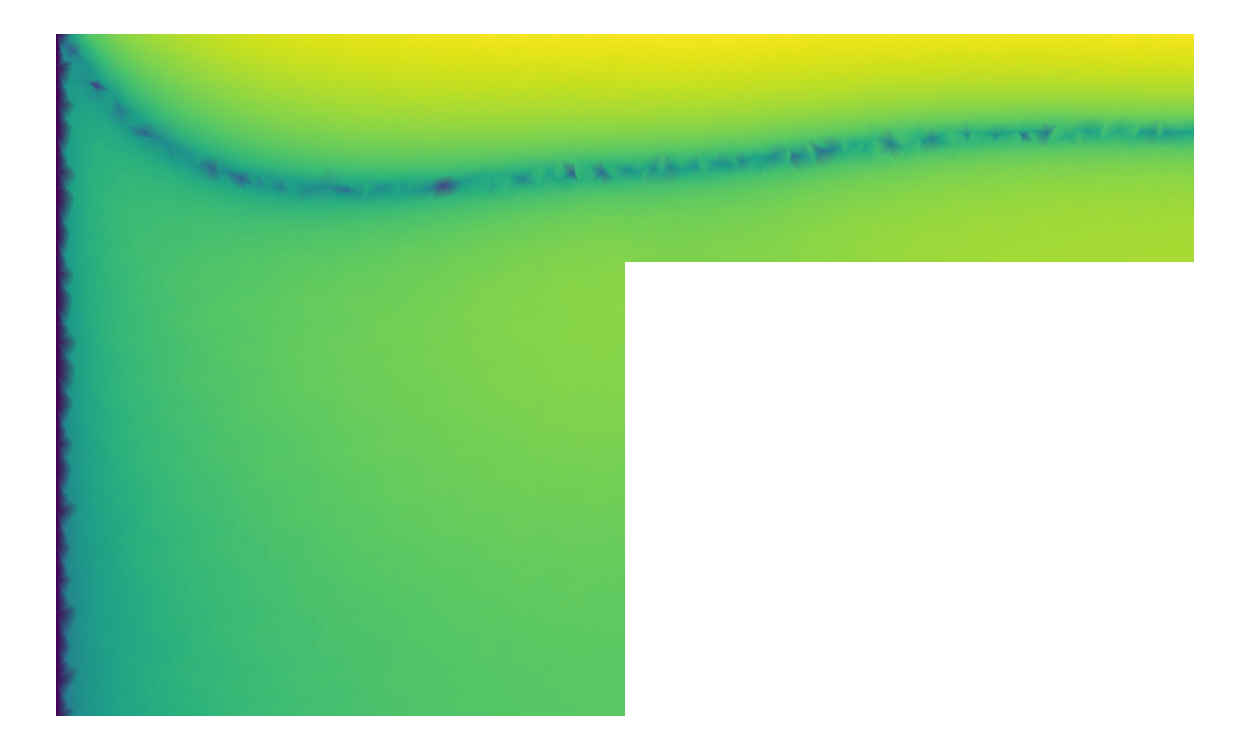}} & {\includegraphics[width=0.23\textwidth]{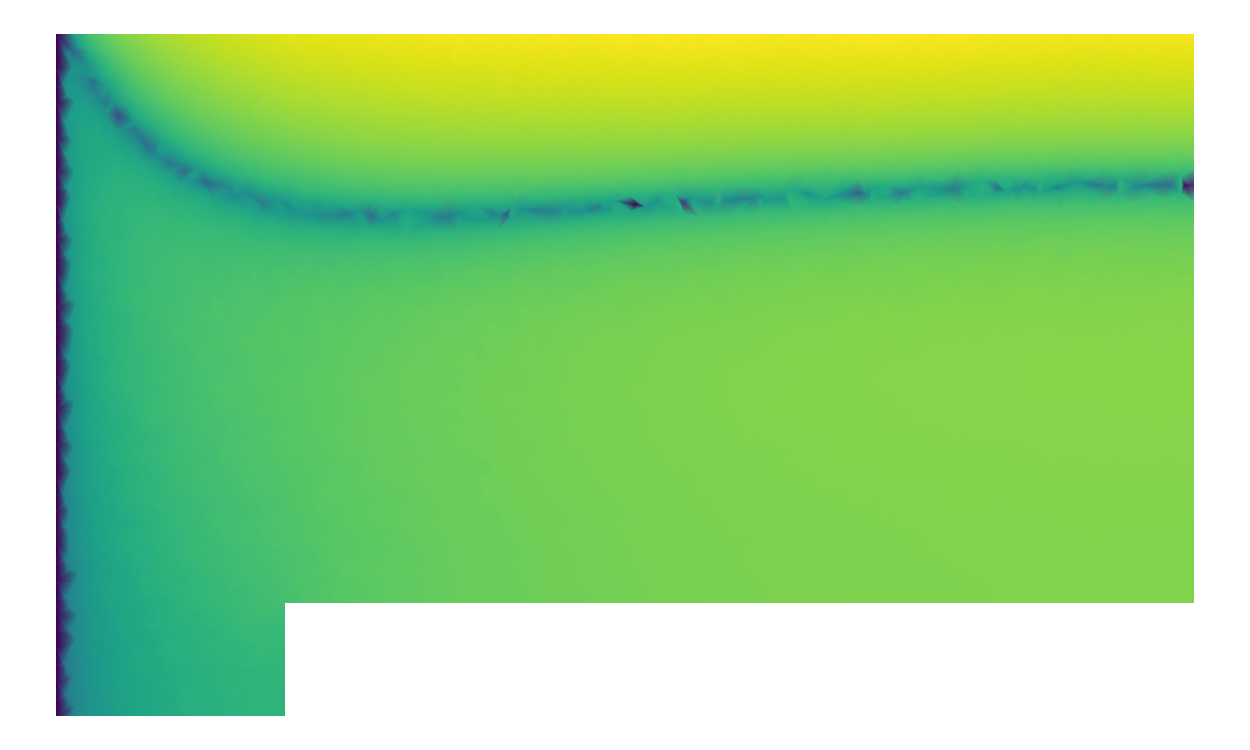}} \\
\rotatebox[origin=c]{90}{\parbox[c]{-2cm}{Absolute\\ error}} & 
{\includegraphics[width=0.23\textwidth]{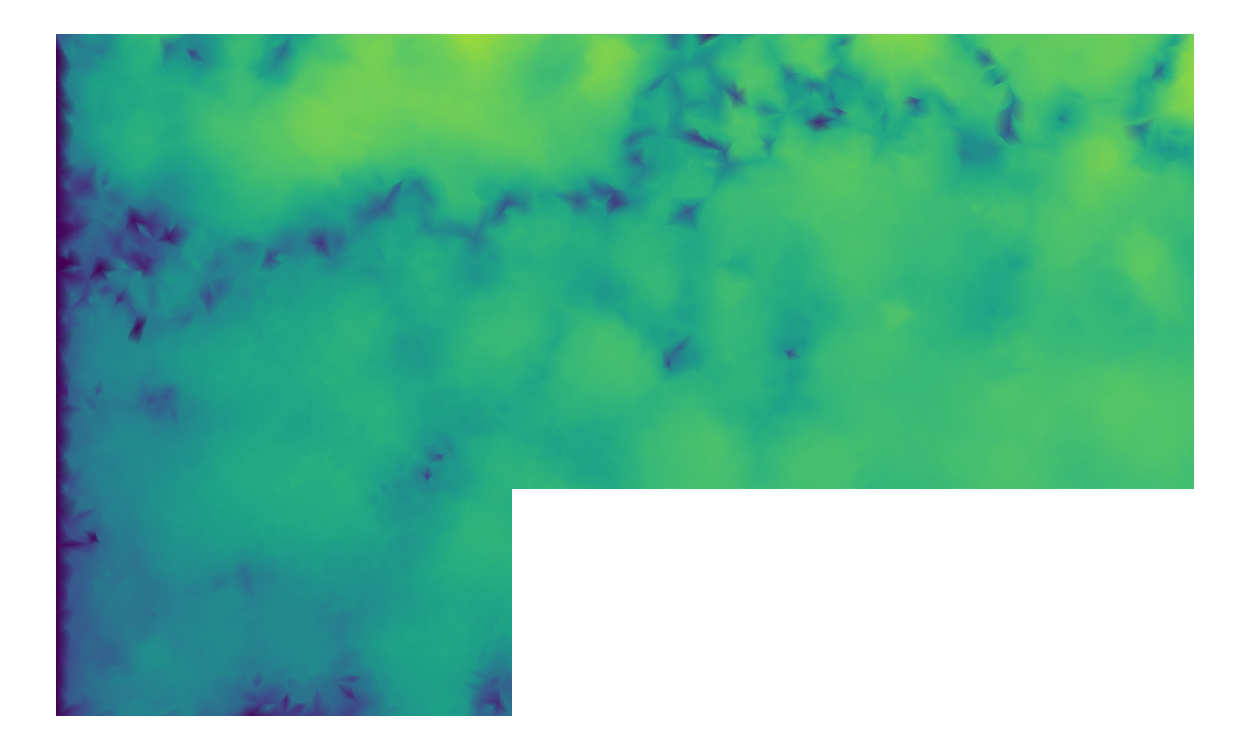}} & {\includegraphics[width=0.23\textwidth]{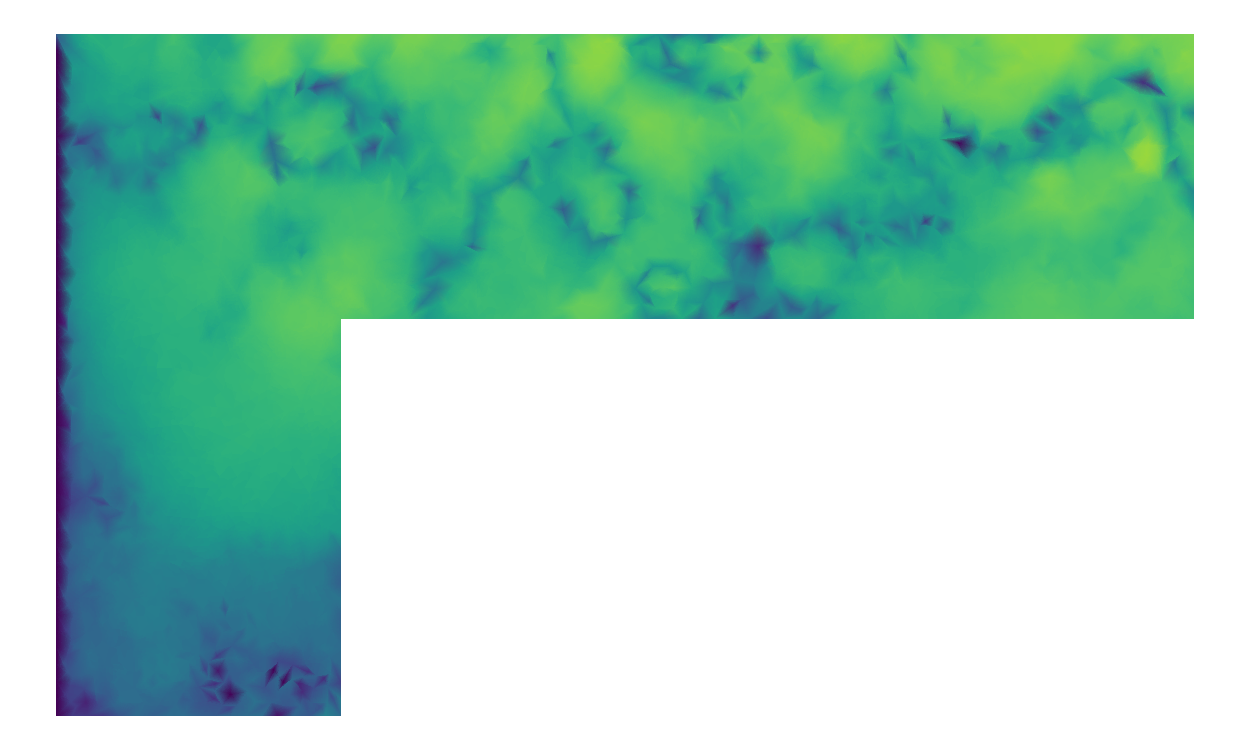}} & {\includegraphics[width=0.23\textwidth]{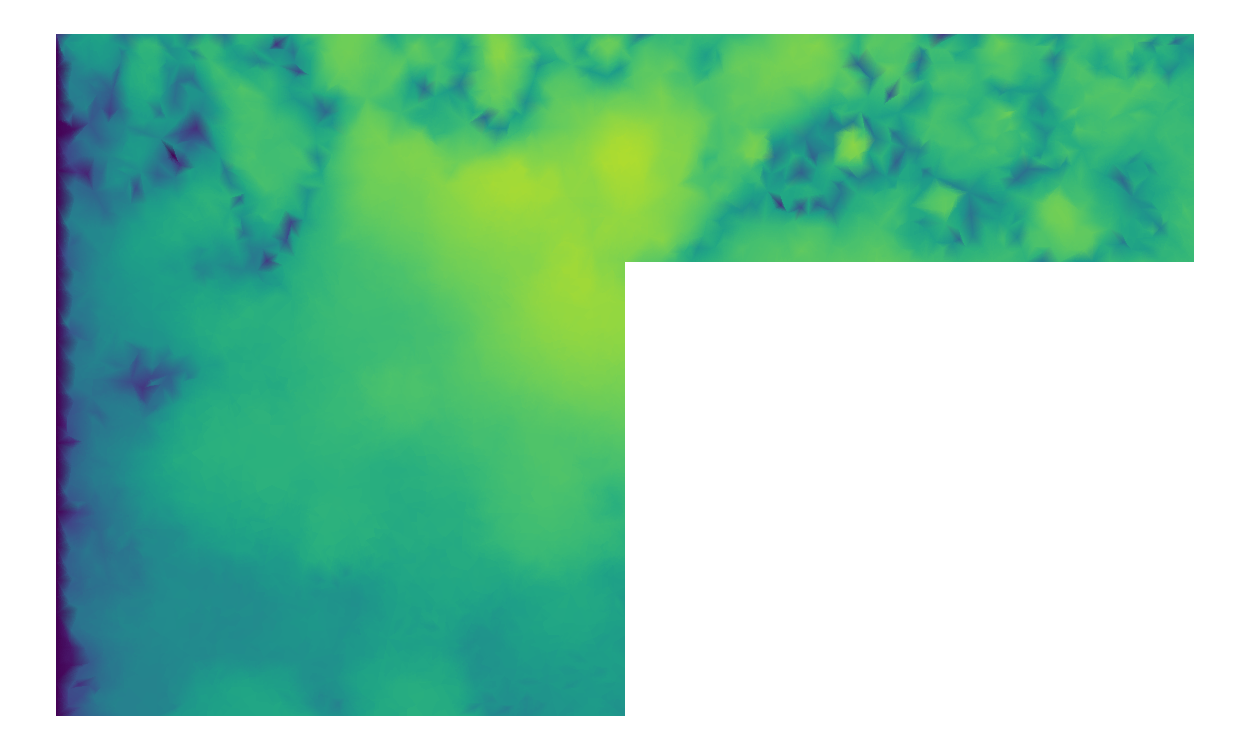}} & {\includegraphics[width=0.23\textwidth]{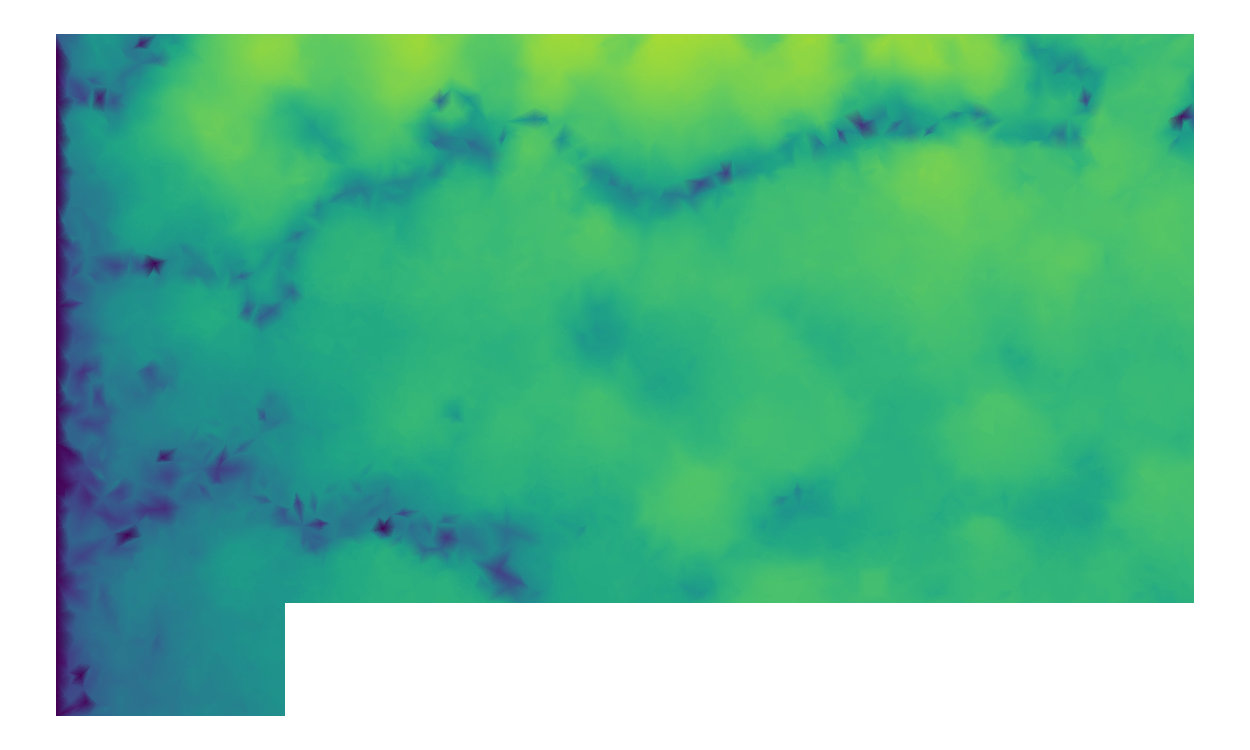}}
\end{tabular}
 \begin{subfigure}{\textwidth}
     \includegraphics[width=0.95\textwidth]{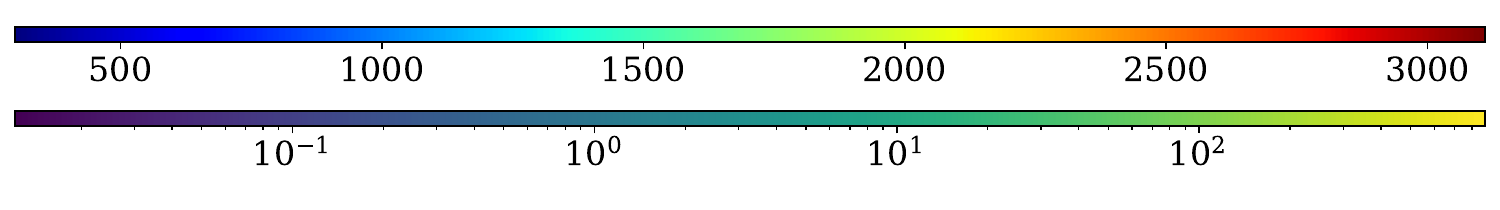}
     \caption*{}
     \label{fig:cbar}
 \end{subfigure} 
\vspace{-1cm}
\caption{Predicted temperature field for unseen designs for the domain shape generalization dataset B2. Our model is able to generalize over different domain shapes and corresponding meshes.}
\label{fig:varshape_unseen}
\end{figure}

In Figure \ref{fig:varshape_unseen}, for the case where $a = b = 0.8$, the maximum relative error in the node of the prediction is less than $10\%$, which is lower than that of the original relative difference between linear and nonlinear simulations.

In the case where $a = 1.2, b = 0.4$, the maximum of error is under $15\%$. In the case where $a = 0.4, b = 1.2$, the error reaches $20\%$, which is still lower than the original difference. This increase in the error can be explained by the shape of the domain affecting the temperature distribution,  which made it significantly different from the temperature distributions in the training dataset. It is also worth noting that this error reaches the maximum at a very localised part of the domain and an average error across all the nodes in a frame is $10\%$.

Eventually, we can state that our hybrid twin is capable of generalizing across different domain geometries with different meshes. The results are in Table \ref{tab:summary_3}.

\begin{table}[ht!]
\begin{center}
\begin{tabular}{p{2.2cm}p{2.1cm}p{2cm}p{1.6cm}}
\toprule
Model & Dataset & MAE, $\text{K}$ & MAPE, \%  \\
\midrule
\multirow{2}{*}{Hybrid twin} & B1 & $0.59 \pm 0.05$ & $0.12 \pm 0.01$  \\
 & B2 & $9.62 \pm 1.54$ & $0.99 \pm 0.15$  \\
\bottomrule
\end{tabular}
\vspace{0.2cm}
\caption{Results of the geometry generalization use cases}
\label{tab:summary_3}

\end{center}

\end{table}

The results of all the use cases together with their RMSE value are summarized in Appendix \ref{appendix:summary_table}.

\subsection{Analysis and Interpretation}

The primary distinction between our work and other approaches employing Graph Neural Networks (GNNs) for physical modeling lies in the proposed hybrid digital twin framework. Specifically, we combine an existing physics-based model with a GNN-based enrichment component. Rather than learning the full governing physics from first principles, the GNN is trained solely to capture the physical effects that are not represented in the underlying linear model. In this way, the neural network acts as a corrective mechanism, augmenting—rather than replacing—the baseline physics formulation.

In contrast to strategies that guide the network to enforce physical consistency by introducing additional constraints into the learning process, as proposed in \cite{toshev2024neural, dupuy2023modeling, quattromini2025mean} for CFD problems or \cite{firouzi2025graph, zhou2024graph,zhai2025stress} for solid mechanics simulations, our approach does not embed such constraints within the training objective. Instead, we adopt a hybrid modeling paradigm in which the existing physics-based model provides the dominant governing behavior, while the neural network component is tasked exclusively with learning the residual physics—that is, the unmodeled or imperfectly captured effects. Residual modeling is a well-established strategy in hybrid machine learning, where the prediction of a physics-based model is complemented by a data-driven correction term that learns the discrepancy between model outputs and observed responses \cite{chinesta2020virtual, rodriguez2023hybrid, ghnatios2024hybrid}.

This framework has been shown to enhance generalization and robustness, particularly in scenarios characterized by limited data availability or incomplete physical descriptions. In such contexts, advanced sampling strategies — such as active learning — may further improve performance \cite{quattromini2023active}. However, the implementation of active learning is beyond the scope of the present work. Instead, a Latin Hypercube Sampling (LHS) strategy was employed for the design space exploration (see Section \ref{sec:gen_load}), ensuring a structured and space-filling distribution of training samples.

\section{Conclusion and perspectives}\label{disc}

In this work, we propose a hybrid twin approach based on the combination of a physics-based model represented by a FEM solver, combined with a data-driven GNN ignorance model to predict discrepancy gaps. We fit the ground truth perceived from synthetic measurements, replicating scenarios where the mathematical approximations do not perfectly fit the reality seen. In this case, we focused on learning nonlinearity terms not considered a priori in the FEM model for heat transfer problems. We show that our hybrid approach is applicable to cases with a restricted amount of data for training (up to 10\% of the dataset samples available). Also, it is able to generalize to different  domain geometries and load locations with partial transfer in case of significantly different mesh connectivities, outperforming baseline data-driven models that approximate the whole phenomenon. Additionally, the inference time of the proposed neural network can be compared with other existing approaches such as pure FEM or other baselines. 

As future perspectives, this work can be extended to other complex problems or applied to real settings with real measurements to test it on real-world benchmarks. To improve the performance of the ignorance model based on GNNs, attention mechanisms on the nodes and edges of the graph could be used. Attention between any two nodes in a graph lets the model to capture long-range dependencies that local message passing struggles to capture \cite{chen2022structure}, while attention across edges captures local geometrical characteristics \cite{brody2021attentive}. Another possibility is the use of temporal attention, which can mitigate error accumulation when forecasting the response over time \cite{han2022predicting}.

\section*{Acknowledgements}

The authors also acknowledge the support of Valeo to perform this research.

\bibliographystyle{splncs04.bst}
\bibliography{bibliography.bib}

@article{moya50post,
  title={Post-earthquake rapid assessment of interconnected electrical equipment based on hybrid modelling},
  author={Moya, Beatriz and Liang, Huangbin and Chinesta, Francisco and Chatzi, Eleni},
  journal={Materials Research Proceedings},
  volume={50},
  pages={52--60},
  year={2025},
  publisher={Materials Research Forum}
}

@article{sancarlos2021learning,
  title={Learning stable reduced-order models for hybrid twins},
  author={Sancarlos, Abel and Cameron, Morgan and Le Peuvedic, Jean-Marc and Groulier, Juliette and Duval, Jean-Louis and Cueto, Elias and Chinesta, Francisco},
  journal={Data-Centric Engineering},
  volume={2},
  pages={e10},
  year={2021},
  publisher={Cambridge University Press}
}

@inproceedings{rodriguez2023hybrid,
  title={Hybrid twin applied to structural health monitoring},
  author={Rodriguez, Sebastian and Lorenzo, DD and Chinesta, Francisco and Monteiro, Eric and Rebillat, Marc and Mechbal, Nazih},
  booktitle={Proceedings of the 10th ECCOMAS Thematic Conference on Smart Structures and Materials (SMART 2023), Patras, Greece},
  pages={3--5},
  year={2023}
}

@article{di2024damage,
  title={Damage identification technique by model enrichment for structural elastodynamic problems},
  author={Di Lorenzo, Daniele and Rodriguez, Sebastian and Champaney, V and Germoso, Claudia and Beringhier, Marianne and Chinesta, F},
  journal={Results in engineering},
  volume={23},
  pages={102389},
  year={2024},
  publisher={Elsevier}
}

@article{ghnatios2023learning,
  title={Learning data-driven stable corrections of dynamical systems—application to the simulation of the top-oil temperature evolution of a power transformer},
  author={Ghnatios, Chady and Kestelyn, Xavier and Denis, Guillaume and Champaney, Victor and Chinesta, Francisco},
  journal={Energies},
  volume={16},
  number={15},
  pages={5790},
  year={2023},
  publisher={MDPI}
}

@article{rodriguez2023hybridscarse,
  title={Hybrid twin of RTM process at the scarce data limit},
  author={Rodriguez, Sebastian and Monteiro, Eric and Mechbal, Nazih and Rebillat, Marc and Chinesta, Francisco},
  journal={International Journal of Material Forming},
  volume={16},
  number={4},
  pages={40},
  year={2023},
  publisher={Springer}
}

@book{taylor2013finite,
  title={The finite element method},
  author={Taylor, Robert Leroy and Zienkiewicz, Olgierd Cecil},
  year={2013},
  publisher={Butterworth-Heinemann Oxford, UK:}
}

@article{mitusch2021hybrid,
  title={Hybrid FEM-NN models: Combining artificial neural networks with the finite element method},
  author={Mitusch, Sebastian K and Funke, Simon W and Kuchta, Miroslav},
  journal={Journal of Computational Physics},
  volume={446},
  pages={110651},
  year={2021},
  publisher={Elsevier}
}

@article{feng2024graph,
  title={Graph convolution network-based surrogate model for natural convection in annuli},
  author={Feng, Feng and Li, Yu-Bai and Chen, Zhi-Hua and Wu, Wei-Tao and Peng, Jiang-Zhou and Mei, Mei},
  journal={Case Studies in Thermal Engineering},
  volume={57},
  pages={104330},
  year={2024},
  publisher={Elsevier}
}

@article{toshev2024neural,
  title={Neural SPH: Improved Neural Modeling of Lagrangian Fluid Dynamics},
  author={Toshev, Artur P and Erbesdobler, Jonas A and Adams, Nikolaus A and Brandstetter, Johannes},
  journal={arXiv preprint arXiv:2402.06275},
  year={2024}
}

@article{barwey2023multiscale,
  title={Multiscale graph neural network autoencoders for interpretable scientific machine learning},
  author={Barwey, Shivam and Shankar, Varun and Viswanathan, Venkatasubramanian and Maulik, Romit},
  journal={Journal of Computational Physics},
  volume={495},
  pages={112537},
  year={2023},
  publisher={Elsevier}
}

@article{li2023finite,
  title={Finite Volume Graph Network (FVGN): Predicting unsteady incompressible fluid dynamics with finite volume informed neural network},
  author={Li, Tianyu and Zou, Shufan and Chang, Xinghua and Zhang, Laiping and Deng, Xiaogang},
  journal={arXiv preprint arXiv:2309.10050},
  year={2023}
}

@article{shao2023pignn,
  title={PIGNN-CFD: A physics-informed graph neural network for rapid predicting urban wind field defined on unstructured mesh},
  author={Shao, Xuqiang and Liu, Zhijian and Zhang, Siqi and Zhao, Zijia and Hu, Chenxing},
  journal={Building and Environment},
  volume={232},
  pages={110056},
  year={2023},
  publisher={Elsevier}
}

@article{hoang2023graph,
  title={Graph representation learning and its applications: a survey},
  author={Hoang, Van Thuy and Jeon, Hyeon-Ju and You, Eun-Soon and Yoon, Yoewon and Jung, Sungyeop and Lee, O-Joun},
  journal={Sensors},
  volume={23},
  number={8},
  pages={4168},
  year={2023},
  publisher={MDPI}
}

@article{wang2021learning,
  title={Learning the solution operator of parametric partial differential equations with physics-informed DeepONets},
  author={Wang, Sifan and Wang, Hanwen and Perdikaris, Paris},
  journal={Science advances},
  volume={7},
  number={40},
  pages={eabi8605},
  year={2021},
  publisher={American Association for the Advancement of Science}
}

@article{karpatne2017theory,
  title={Theory-guided data science: A new paradigm for scientific discovery from data},
  author={Karpatne, Anuj and Atluri, Gowtham and Faghmous, James H and Steinbach, Michael and Banerjee, Arindam and Ganguly, Auroop and Shekhar, Shashi and Samatova, Nagiza and Kumar, Vipin},
  journal={IEEE Transactions on knowledge and data engineering},
  volume={29},
  number={10},
  pages={2318--2331},
  year={2017},
  publisher={IEEE}
}

@article{pfaff2020learning,
  title={Learning mesh-based simulation with graph networks},
  author={Pfaff, Tobias and Fortunato, Meire and Sanchez-Gonzalez, Alvaro and Battaglia, Peter W},
  journal={arXiv preprint arXiv:2010.03409},
  year={2020}
}

@article{haghighat2021sciann,
  title={SciANN: A Keras/TensorFlow wrapper for scientific computations and physics-informed deep learning using artificial neural networks},
  author={Haghighat, Ehsan and Juanes, Ruben},
  journal={Computer Methods in Applied Mechanics and Engineering},
  volume={373},
  pages={113552},
  year={2021},
  publisher={Elsevier}
}

@inproceedings{hennigh2021nvidia,
  title={NVIDIA SimNet™: An AI-accelerated multi-physics simulation framework},
  author={Hennigh, Oliver and Narasimhan, Susheela and Nabian, Mohammad Amin and Subramaniam, Akshay and Tangsali, Kaustubh and Fang, Zhiwei and Rietmann, Max and Byeon, Wonmin and Choudhry, Sanjay},
  booktitle={International conference on computational science},
  pages={447--461},
  year={2021},
  organization={Springer}
}

@article{chinesta2020virtual,
  title={Virtual, digital and hybrid twins: a new paradigm in data-based engineering and engineered data},
  author={Chinesta, Francisco and Cueto, Elias and Abisset-Chavanne, Emmanuelle and Duval, Jean Louis and Khaldi, Fouad El},
  journal={Archives of computational methods in engineering},
  volume={27},
  pages={105--134},
  year={2020},
  publisher={Springer}
}

@book{oberkampf2010verification,
  title={Verification and validation in scientific computing},
  author={Oberkampf, William L and Roy, Christopher J},
  year={2010},
  publisher={Cambridge university press}
}

@article{quarteroni2025combining,
  title={Combining physics-based and data-driven models: advancing the frontiers of research with scientific machine learning},
  author={Quarteroni, Alfio and Gervasio, Paola and Regazzoni, Francesco},
  journal={arXiv preprint arXiv:2501.18708},
  year={2025}
}

@article{wu2024physics,
  title={Physics-informed machine learning: A comprehensive review on applications in anomaly detection and condition monitoring},
  author={Wu, Yuandi and Sicard, Brett and Gadsden, Stephen Andrew},
  journal={Expert Systems with Applications},
  volume={255},
  pages={124678},
  year={2024},
  publisher={Elsevier}
}

@book{strikwerda2004finite,
  title={Finite difference schemes and partial differential equations},
  author={Strikwerda, John C},
  year={2004},
  publisher={SIAM}
}

@article{moya2022digital,
  title={Digital twins that learn and correct themselves},
  author={Moya, Beatriz and Bad{\'\i}as, Alberto and Alfaro, Ic{\'\i}ar and Chinesta, Francisco and Cueto, El{\'\i}as},
  journal={International Journal for Numerical Methods in Engineering},
  volume={123},
  number={13},
  pages={3034--3044},
  year={2022},
  publisher={Wiley Online Library}
}

@article{amini2022physics,
  title={Physics-informed neural network solution of thermo--hydro--mechanical processes in porous media},
  author={Amini, Danial and Haghighat, Ehsan and Juanes, Ruben},
  journal={Journal of Engineering Mechanics},
  volume={148},
  number={11},
  pages={04022070},
  year={2022},
  publisher={American Society of Civil Engineers}
}

@article{xu2023physics,
  title={Physics-informed neural networks for studying heat transfer in porous media},
  author={Xu, Jiaxuan and Wei, Han and Bao, Hua},
  journal={International Journal of Heat and Mass Transfer},
  volume={217},
  pages={124671},
  year={2023},
  publisher={Elsevier}
}

@article{eivazi2024physics,
  title={Physics-informed deep-learning applications to experimental fluid mechanics},
  author={Eivazi, Hamidreza and Wang, Yuning and Vinuesa, Ricardo},
  journal={Measurement science and technology},
  volume={35},
  number={7},
  pages={075303},
  year={2024},
  publisher={IOP Publishing}
}

@article{zou2024correcting,
  title={Correcting model misspecification in physics-informed neural networks (PINNs)},
  author={Zou, Zongren and Meng, Xuhui and Karniadakis, George Em},
  journal={Journal of Computational Physics},
  volume={505},
  pages={112918},
  year={2024},
  publisher={Elsevier}
}

@article{kurz2022hybrid,
  title={Hybrid modeling: towards the next level of scientific computing in engineering},
  author={Kurz, Stefan and De Gersem, Herbert and Galetzka, Armin and Klaedtke, Andreas and Liebsch, Melvin and Loukrezis, Dimitrios and Russenschuck, Stephan and Schmidt, Manuel},
  journal={Journal of Mathematics in Industry},
  volume={12},
  number={1},
  pages={8},
  year={2022},
  publisher={Springer}
}

@article{greydanus2019hamiltonian,
  title={Hamiltonian neural networks},
  author={Greydanus, Samuel and Dzamba, Misko and Yosinski, Jason},
  journal={Advances in neural information processing systems},
  volume={32},
  year={2019}
}

@article{tong2021symplectic,
  title={Symplectic neural networks in Taylor series form for Hamiltonian systems},
  author={Tong, Yunjin and Xiong, Shiying and He, Xingzhe and Pan, Guanghan and Zhu, Bo},
  journal={Journal of Computational Physics},
  volume={437},
  pages={110325},
  year={2021},
  publisher={Elsevier}
}

@article{di2022data,
  title={Data completion, model correction and enrichment based on sparse identification and data assimilation},
  author={Di Lorenzo, Daniele and Champaney, Victor and Germoso, Claudia and Cueto, Elias and Chinesta, Francisco},
  journal={Applied Sciences},
  volume={12},
  number={15},
  pages={7458},
  year={2022},
  publisher={MDPI}
}

@article{luo2023hybrid,
  title={A hybrid model for modelling arbitrary cracks in isotropic plate structures},
  author={Luo, Yulin and Featherston, Carol A and Kennedy, David},
  journal={Thin-Walled Structures},
  volume={183},
  pages={110345},
  year={2023},
  publisher={Elsevier}
}

@article{ghnatios2024hybrid,
  title={A hybrid twin based on machine learning enhanced reduced order model for real-time simulation of magnetic bearings},
  author={Ghnatios, Chady and Rodriguez, Sebastian and Tomezyk, Jerome and Dupuis, Yves and Mouterde, Joel and Da Silva, Joaquim and Chinesta, Francisco},
  journal={Advanced Modeling and Simulation in Engineering Sciences},
  volume={11},
  number={1},
  pages={3},
  year={2024},
  publisher={Springer}
}

@article{daby2025finite,
  title={Finite element neural network interpolation: Part II—hybridisation with the proper generalised decomposition for non-linear surrogate modelling},
  author={Daby-Seesaram, Alexandre and {\v{S}}kardov{\'a}, Kate{\v{r}}ina and Genet, Martin},
  journal={Computational Mechanics},
  pages={1--26},
  year={2025},
  publisher={Springer}
}

@article{liang2025harnessing,
  title={Harnessing hybrid digital twinning for decision-support in smart infrastructures},
  author={Liang, Huangbin and Moya, Beatriz and Seah, Eugene and Weng, Ashley Ng Kwok and Baillargeat, Dominique and Joerin, Jonas and Zhang, Xiaozheng and Chinesta, Francisco and Chatzi, Eleni},
  journal={Data-Centric Engineering},
  volume={6},
  pages={e43},
  year={2025},
  publisher={Cambridge University Press}
}

@article{sancarlos2020rom,
  title={From ROM of electrochemistry to AI-based battery digital and hybrid twin},
  author={Sancarlos, Abel and Cameron, Morgan and Abel, Andreas and Cueto, Elias and Duval, Jean-Louis and Chinesta, Francisco},
  journal={Archives of Computational Methods in Engineering},
  volume={28},
  number={3},
  pages={979--1015},
  year={2020}
}

@article{yun2022novel,
  title={A novel digital twin architecture with similarity-based hybrid modeling for supporting dependable disaster management systems},
  author={Yun, Seong-Jin and Kwon, Jin-Woo and Kim, Won-Tae},
  journal={Sensors},
  volume={22},
  number={13},
  pages={4774},
  year={2022},
  publisher={MDPI}
}

@article{sun2022physinet,
  title={PhysiNet: A combination of physics-based model and neural network model for digital twins},
  author={Sun, Chao and Shi, Victor G},
  journal={International journal of intelligent systems},
  volume={37},
  number={8},
  pages={5443--5456},
  year={2022},
  publisher={Wiley Online Library}
}

@article{holt2024automatically,
  title={Automatically learning hybrid digital twins of dynamical systems},
  author={Holt, Samuel and Liu, Tennison and van der Schaar, Mihaela},
  journal={Advances in Neural Information Processing Systems},
  volume={37},
  pages={72170--72218},
  year={2024}
}

@article{gonzalez2019learning,
  title={Learning corrections for hyperelastic models from data},
  author={Gonz{\'a}lez, David and Chinesta, Francisco and Cueto, El{\'\i}as},
  journal={Frontiers in Materials},
  volume={6},
  pages={14},
  year={2019},
  publisher={Frontiers Media SA}
}

@article{bonavita2020machine,
  title={Machine learning for model error inference and correction},
  author={Bonavita, Massimo and Laloyaux, Patrick},
  journal={Journal of Advances in Modeling Earth Systems},
  volume={12},
  number={12},
  pages={e2020MS002232},
  year={2020},
  publisher={Wiley Online Library}
}

@article{montans2019data,
  title={Data-driven modeling and learning in science and engineering},
  author={Mont{\'a}ns, Francisco J and Chinesta, Francisco and G{\'o}mez-Bombarelli, Rafael and Kutz, J Nathan},
  journal={Comptes Rendus M{\'e}canique},
  volume={347},
  number={11},
  pages={845--855},
  year={2019},
  publisher={Elsevier}
}

@article{willard2020integrating,
  title={Integrating physics-based modeling with machine learning: A survey},
  author={Willard, Jared and Jia, Xiaowei and Xu, Shaoming and Steinbach, Michael and Kumar, Vipin},
  journal={arXiv preprint arXiv:2003.04919},
  volume={1},
  number={1},
  pages={1--34},
  year={2020}
}

@article{tierz2025feasibility,
  title={On the feasibility of foundational models for the simulation of physical phenomena},
  author={Tierz, Alicia and Iparraguirre, Mikel M and Alfaro, Ic{\'\i}ar and Gonz{\'a}lez, David and Chinesta, Francisco and Cueto, El{\'\i}as},
  journal={International Journal for Numerical Methods in Engineering},
  volume={126},
  number={6},
  pages={e70027},
  year={2025},
  publisher={Wiley Online Library}
}

@article{champaney2022engineering,
  title={Engineering empowered by physics-based and data-driven hybrid models: A methodological overview},
  author={Champaney, Victor and Chinesta, Francisco and Cueto, Elias},
  journal={International Journal of Material Forming},
  volume={15},
  number={3},
  pages={31},
  year={2022},
  publisher={Springer}
}

@article{idrissi2024multiscale,
  title={Multiscale Thermodynamics-Informed Neural Networks (MuTINN) towards fast and frugal inelastic computation of woven composite structures},
  author={Idrissi, M El Fallaki},
  journal={Journal of the Mechanics and Physics of Solids},
  volume={186},
  pages={105604},
  year={2024},
  publisher={Elsevier}
}

@article{cuomo2022scientific,
  title={Scientific machine learning through physics--informed neural networks: Where we are and what’s next},
  author={Cuomo, Salvatore and Di Cola, Vincenzo Schiano and Giampaolo, Fabio and Rozza, Gianluigi and Raissi, Maziar and Piccialli, Francesco},
  journal={Journal of Scientific Computing},
  volume={92},
  number={3},
  pages={88},
  year={2022},
  publisher={Springer}
}

@article{bronstein2021geometric,
  title={Geometric deep learning: Grids, groups, graphs, geodesics, and gauges},
  author={Bronstein, Michael M and Bruna, Joan and Cohen, Taco and Veli{\v{c}}kovi{\'c}, Petar},
  journal={arXiv preprint arXiv:2104.13478},
  year={2021}
}

@article{zhang2020deep,
  title={Deep learning on graphs: A survey},
  author={Zhang, Ziwei and Cui, Peng and Zhu, Wenwu},
  journal={IEEE Transactions on Knowledge and Data Engineering},
  volume={34},
  number={1},
  pages={249--270},
  year={2020},
  publisher={IEEE}
}

@article{thangamuthu2022unravelling,
  title={Unravelling the performance of physics-informed graph neural networks for dynamical systems},
  author={Thangamuthu, Abishek and Kumar, Gunjan and Bishnoi, Suresh and Bhattoo, Ravinder and Krishnan, NM and Ranu, Sayan},
  journal={Advances in Neural Information Processing Systems},
  volume={35},
  pages={3691--3702},
  year={2022}
}

@article{bermejo2025meshgraphnets,
  title={Meshgraphnets informed locally by thermodynamics for the simulation of flows around arbitrarily shaped objects},
  author={Bermejo, Carlos and Bad{\'\i}as, Alberto and Gonz{\'a}lez, David and Cueto, El{\'\i}as},
  journal={Advanced Modeling and Simulation in Engineering Sciences},
  volume={12},
  number={1},
  pages={27},
  year={2025},
  publisher={Springer}
}

@article{battaglia2018relational,
  title={Relational inductive biases, deep learning, and graph networks},
  author={Battaglia, Peter W and Hamrick, Jessica B and Bapst, Victor and Sanchez-Gonzalez, Alvaro and Zambaldi, Vinicius and Malinowski, Mateusz and Tacchetti, Andrea and Raposo, David and Santoro, Adam and Faulkner, Ryan and others},
  journal={arXiv preprint arXiv:1806.01261},
  year={2018}
}

@article{velivckovic2017graph,
  title={Graph attention networks},
  author={Veli{\v{c}}kovi{\'c}, Petar and Cucurull, Guillem and Casanova, Arantxa and Romero, Adriana and Lio, Pietro and Bengio, Yoshua},
  journal={arXiv preprint arXiv:1710.10903},
  year={2017}
}

@article{hamilton2017inductive,
  title={Inductive representation learning on large graphs},
  author={Hamilton, Will and Ying, Zhitao and Leskovec, Jure},
  journal={Advances in neural information processing systems},
  volume={30},
  year={2017}
}

@article{raissi2017physics,
  title={Physics informed deep learning (part i): Data-driven solutions of nonlinear partial differential equations},
  author={Raissi, Maziar and Perdikaris, Paris and Karniadakis, George Em},
  journal={arXiv preprint arXiv:1711.10561},
  year={2017}
}

@article{kipf2016semi,
  title={Semi-supervised classification with graph convolutional networks},
  author={Kipf, Thomas N and Welling, Max},
  journal={arXiv preprint arXiv:1609.02907},
  year={2016}
}

@article{han2022predicting,
  title={Predicting physics in mesh-reduced space with temporal attention},
  author={Han, Xu and Gao, Han and Pfaff, Tobias and Wang, Jian-Xun and Liu, Li-Ping},
  journal={arXiv preprint arXiv:2201.09113},
  year={2022}
}

@inproceedings{sanchez2020learning,
  title={Learning to simulate complex physics with graph networks},
  author={Sanchez-Gonzalez, Alvaro and Godwin, Jonathan and Pfaff, Tobias and Ying, Rex and Leskovec, Jure and Battaglia, Peter},
  booktitle={International conference on machine learning},
  pages={8459--8468},
  year={2020},
  organization={PMLR}
}

@article{eslamlou2025hybrid,
  title={A hybrid data-physics framework with conformal GNN for enhanced damage identification},
  author={Eslamlou, Armin Dadras and Ghasemlou, Arshia and Barros, Brais and Riveiro, Bel{\'e}n},
  journal={Advanced Engineering Informatics},
  volume={68},
  pages={103718},
  year={2025},
  publisher={Elsevier}
}

@inproceedings{he2016deep,
  title={Deep residual learning for image recognition},
  author={He, Kaiming and Zhang, Xiangyu and Ren, Shaoqing and Sun, Jian},
  booktitle={Proceedings of the IEEE conference on computer vision and pattern recognition},
  pages={770--778},
  year={2016}
}

@article{brody2021attentive,
  title={How attentive are graph attention networks?},
  author={Brody, Shaked and Alon, Uri and Yahav, Eran},
  journal={arXiv preprint arXiv:2105.14491},
  year={2021}
}

@inproceedings{chen2022structure,
  title={Structure-aware transformer for graph representation learning},
  author={Chen, Dexiong and O’Bray, Leslie and Borgwardt, Karsten},
  booktitle={International conference on machine learning},
  pages={3469--3489},
  year={2022},
  organization={PMLR}
}

@inproceedings{toshev2023learning,
  title={Learning lagrangian fluid mechanics with e (3)-equivariant graph neural networks},
  author={Toshev, Artur P and Galletti, Gianluca and Brandstetter, Johannes and Adami, Stefan and Adams, Nikolaus A},
  booktitle={International Conference on Geometric Science of Information},
  pages={332--341},
  year={2023},
  organization={Springer}
}

@article{dupuy2023modeling,
  title={Modeling the wall shear stress in large-eddy simulation using graph neural networks},
  author={Dupuy, Dorian and Odier, Nicolas and Lapeyre, Corentin and Papadogiannis, Dimitrios},
  journal={Data-Centric Engineering},
  volume={4},
  pages={e7},
  year={2023},
  publisher={Cambridge University Press}
}

@article{quattromini2023active,
  title={Active learning of data-assimilation closures using Graph Neural Networks},
  author={Quattromini, Michele and Bucci, Michele Alessandro and Cherubini, Stefania and Semeraro, Onofrio},
  journal={arXiv preprint arXiv:2303.03806},
  year={2023}
}

@article{quattromini2025mean,
  title={Mean flow data assimilation using physics-constrained graph neural networks},
  author={Quattromini, Michele and Bucci, Michele Alessandro and Cherubini, Stefania and Semeraro, Onofrio},
  journal={Data-Centric Engineering},
  volume={6},
  pages={e48},
  year={2025},
  publisher={Cambridge University Press}
}

@article{bhattoo2022learning,
  title={Learning articulated rigid body dynamics with lagrangian graph neural network},
  author={Bhattoo, Ravinder and Ranu, Sayan and Krishnan, NM},
  journal={Advances in Neural Information Processing Systems},
  volume={35},
  pages={29789--29800},
  year={2022}
}

@article{firouzi2025graph,
  title={Graph-FEM/ML framework for inverse load identification in thick-walled hyperelastic pressure vessels},
  author={Firouzi, Nasser and Hafez, Ramy M and Salloomi, Kareem N and Abdelkawy, Mohamed A and Hussain, Raja Rizwan},
  journal={Symmetry},
  volume={17},
  number={12},
  pages={2021},
  year={2025},
  publisher={MDPI}
}

@article{zhou2024graph,
  title={Graph neural networks to simulate flexible pavement responses using three-dimensional finite element analysis data},
  author={Zhou, Qingwen and Al-Qadi, Imad L},
  journal={Transportation Research Record},
  volume={2678},
  number={11},
  pages={1111--1127},
  year={2024},
  publisher={SAGE Publications Sage CA: Los Angeles, CA}
}

@article{zhai2025stress,
  title={Stress predictions in polycrystal plasticity using graph neural networks with subgraph training},
  author={Zhai, Hanfeng},
  journal={Computational Mechanics},
  volume={76},
  number={2},
  pages={387--408},
  year={2025},
  publisher={Springer}
}

@article{zaheer2017deep,
  title={Deep sets},
  author={Zaheer, Manzil and Kottur, Satwik and Ravanbakhsh, Siamak and Poczos, Barnabas and Salakhutdinov, Russ R and Smola, Alexander J},
  journal={Advances in neural information processing systems},
  volume={30},
  year={2017}
}

@article{xu2018powerful,
  title={How powerful are graph neural networks?},
  author={Xu, Keyulu and Hu, Weihua and Leskovec, Jure and Jegelka, Stefanie},
  journal={arXiv preprint arXiv:1810.00826},
  year={2018}
}

@article{bronstein2017geometric,
  title={Geometric deep learning: going beyond euclidean data},
  author={Bronstein, Michael M and Bruna, Joan and LeCun, Yann and Szlam, Arthur and Vandergheynst, Pierre},
  journal={IEEE Signal Processing Magazine},
  volume={34},
  number={4},
  pages={18--42},
  year={2017},
  publisher={IEEE}
}

\section*{Appendix}
\appendix

\section{Baseline architecture}\label{appendix:base_arch}
The architecture of the MGN network replicates the Encoder-Processor-Decoder structure described in Section \ref{method}. The difference between it and our hybrid twin is the input and output of the network.

For MGN applied to thermal simulations, the input node features include:

\begin{equation*}
    \bz_{i}(t) = T^\text{GT}_i(t).
\end{equation*}

\noindent
And the output node features are:

\begin{equation*}
    \by_{i}(t) = \frac{\bz_{i}(t+\Delta t) - \bz_{i}(t)}{\Delta t}
\end{equation*}

\noindent
or more specifically:

\begin{equation*}
    y_{i}(t) = \frac{T_{i}^\text{GT}(t+\Delta t) - T_{i}^\text{GT}(t)}{\Delta t}.
\end{equation*}

Thus, the network learns the temperature increment, and as a result, the whole model predicts the state of the system in the next timestep $t = t+\Delta t$ by using the increment $y_{i}(t)$ and the previous state $T_{i}^\text{GT}(t)$. During inference, the model autoregressively predicts the state of the system at each timestep.

\section{Hyperparameters}

Here we list key hyperparameters used during training. We use Adam optimizer with the constant learning rate equal to $10^{-3}$, the number of message passing steps is equal to $10$, and the batch size is equal to 13 frames. The NI standard deviation is set to 10; this value is selected based on the RMSE of the one-step MGN model \cite{sanchez2020learning}.

\section{Additional use cases} \label{appendix:add_exp}

Here we mention the results for the use cases omitted in the main text of the paper for brevity.

\subsection{Geometry generalization: additional figures}\label{appendix:gauss_all}

\begin{figure}[ht!]
 \begin{subfigure}{0.482\textwidth}
     \includegraphics[width=\textwidth]{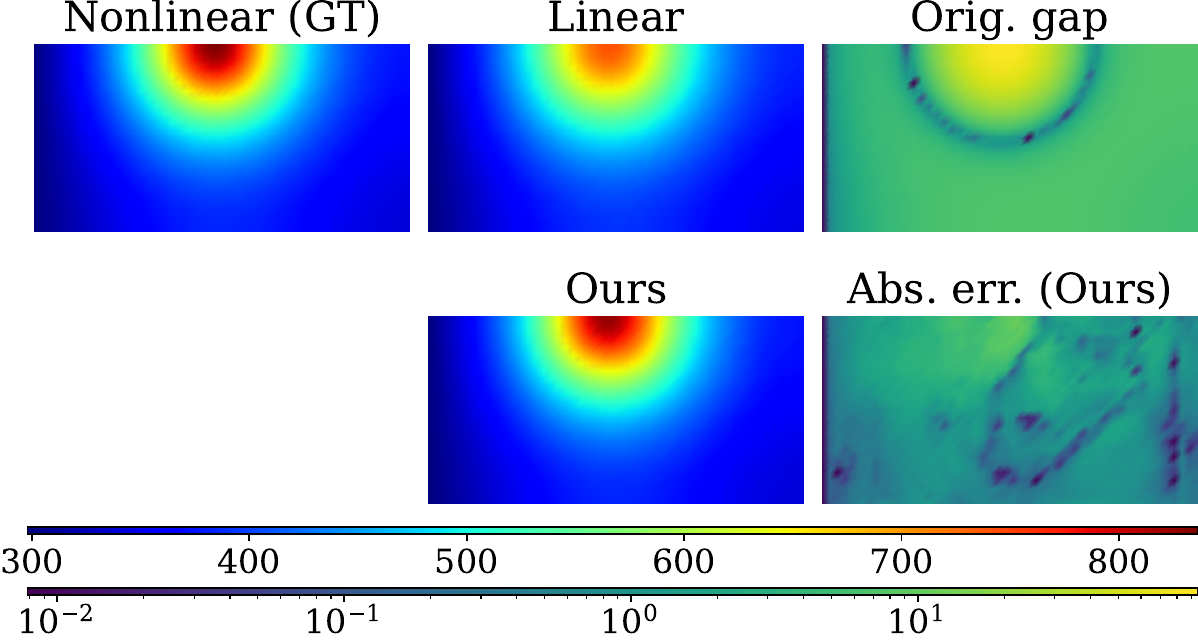}
     \caption{}
     \label{fig:gauss_d1}
 \end{subfigure}
\hspace{0.28cm}
\vspace{0.2cm}
 \begin{subfigure}{0.5\textwidth}
     \includegraphics[width=\textwidth]{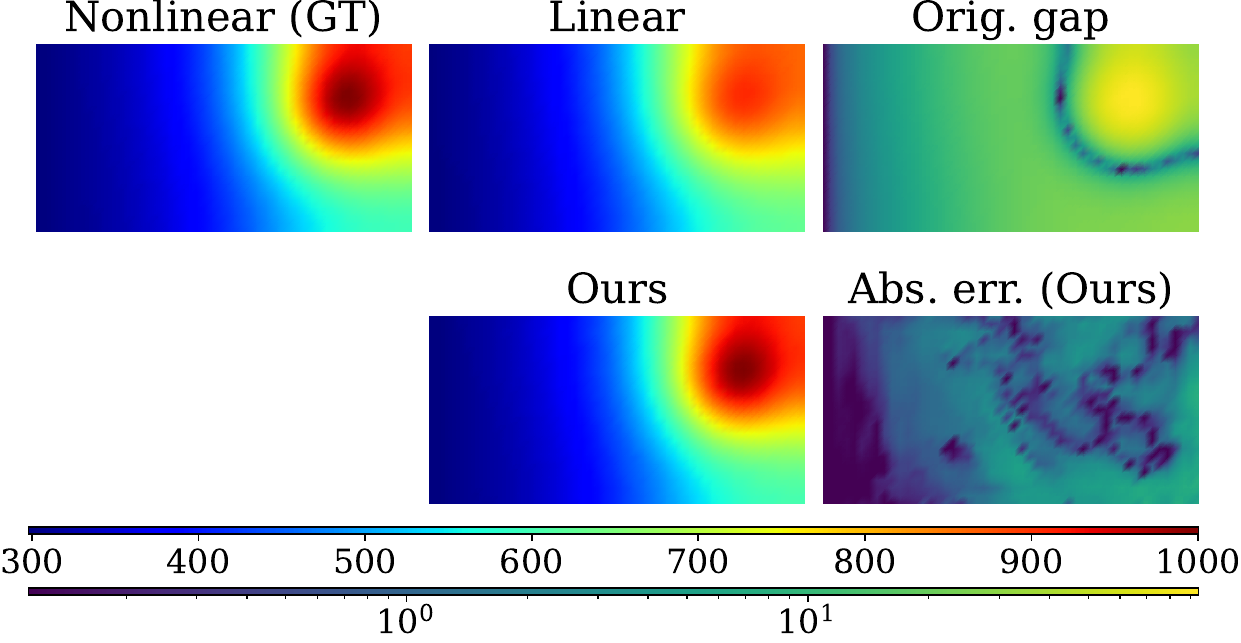}
     \caption{}
     \label{fig:gauss_d3}
 \end{subfigure}
 \medskip
 \begin{subfigure}{0.49\textwidth}
     \includegraphics[width=\textwidth]{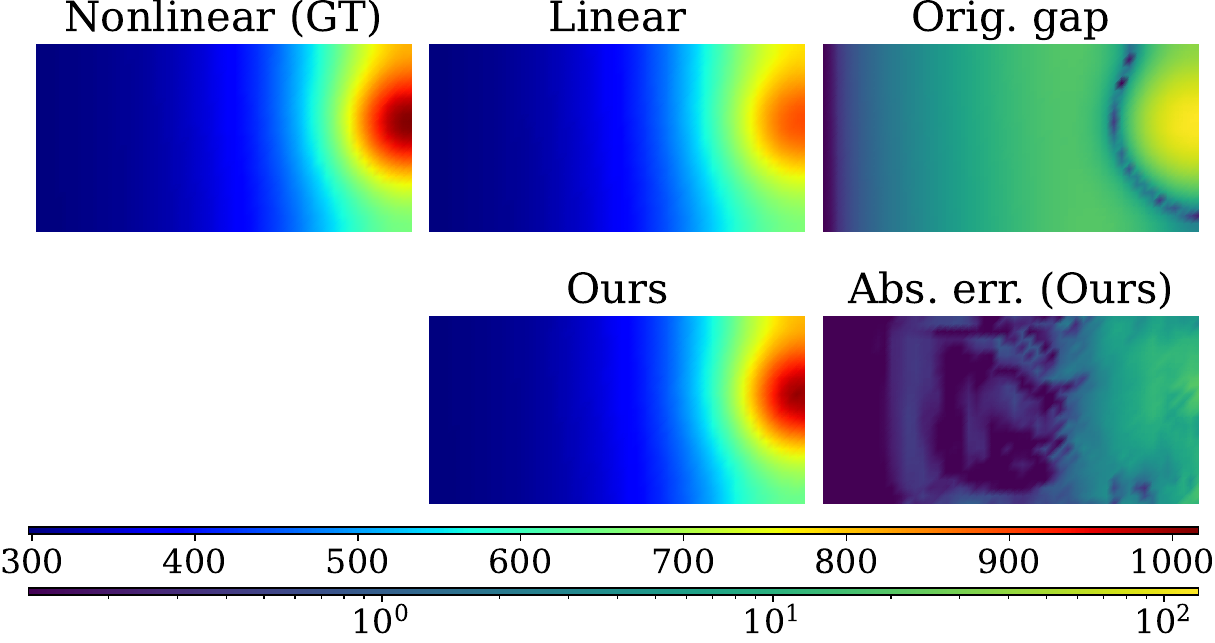}
     \caption{}
     \label{fig:gauss_d4}
 \end{subfigure}
 \begin{subfigure}{0.482\textwidth}
     \includegraphics[width=\textwidth]{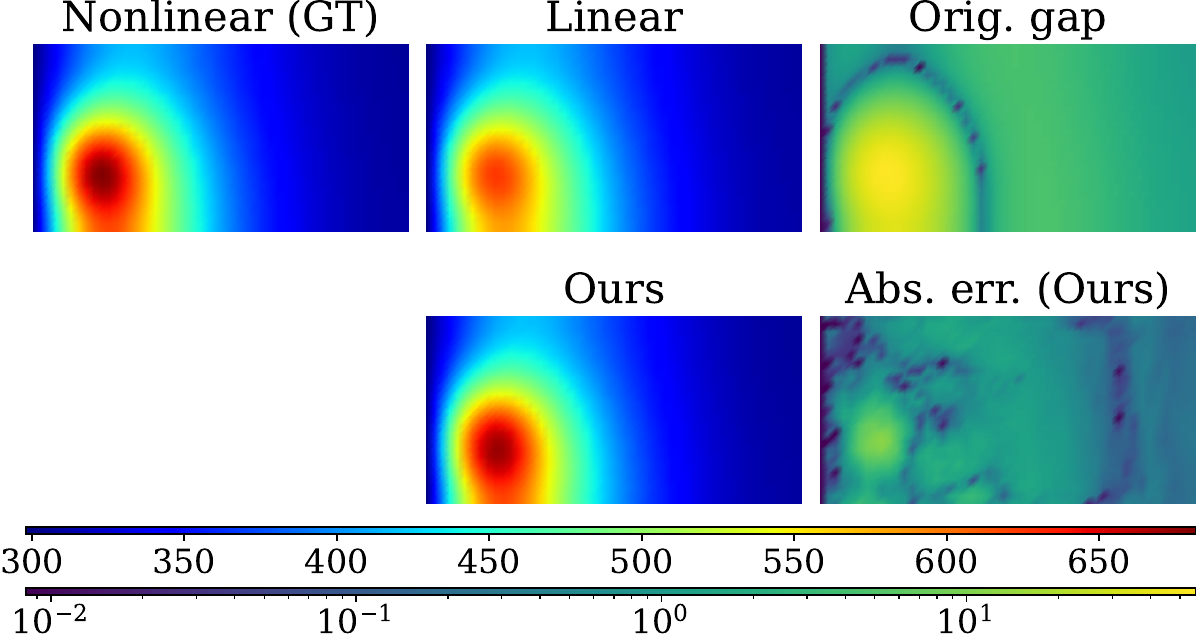}
     \caption{}
     \label{fig:gauss_d6}
 \end{subfigure}
 \medskip
 \begin{subfigure}{0.482\textwidth}
     \includegraphics[width=\textwidth]{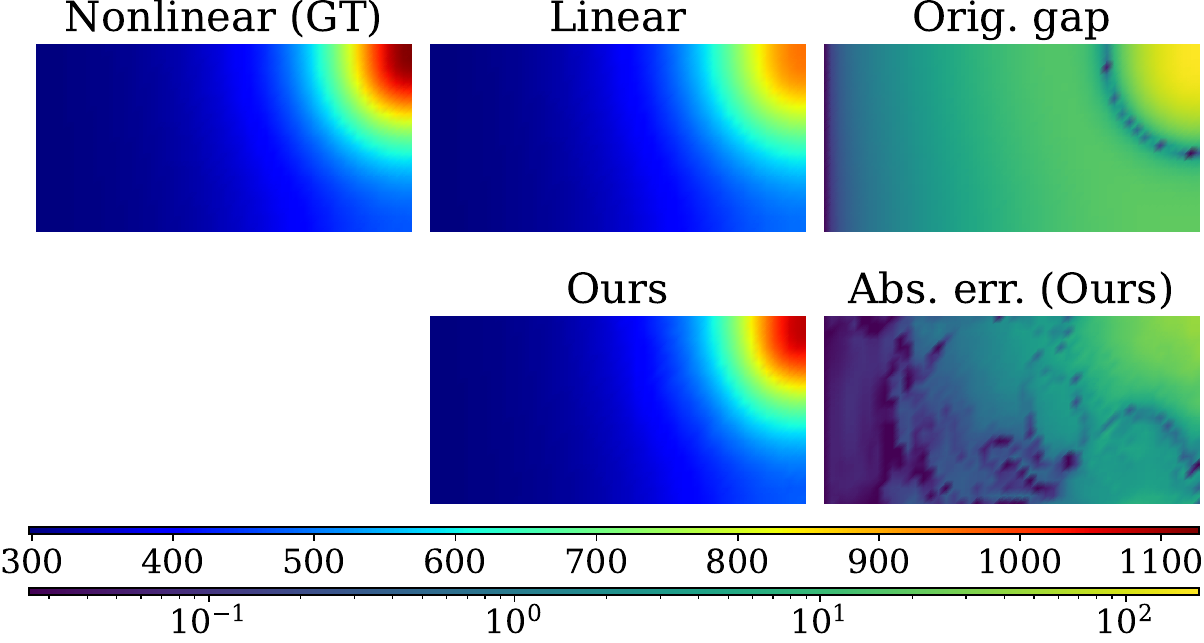}
     \caption{}
     \label{fig:gauss_d7}
 \end{subfigure}
 \begin{subfigure}{0.482\textwidth}
     \includegraphics[width=\textwidth]{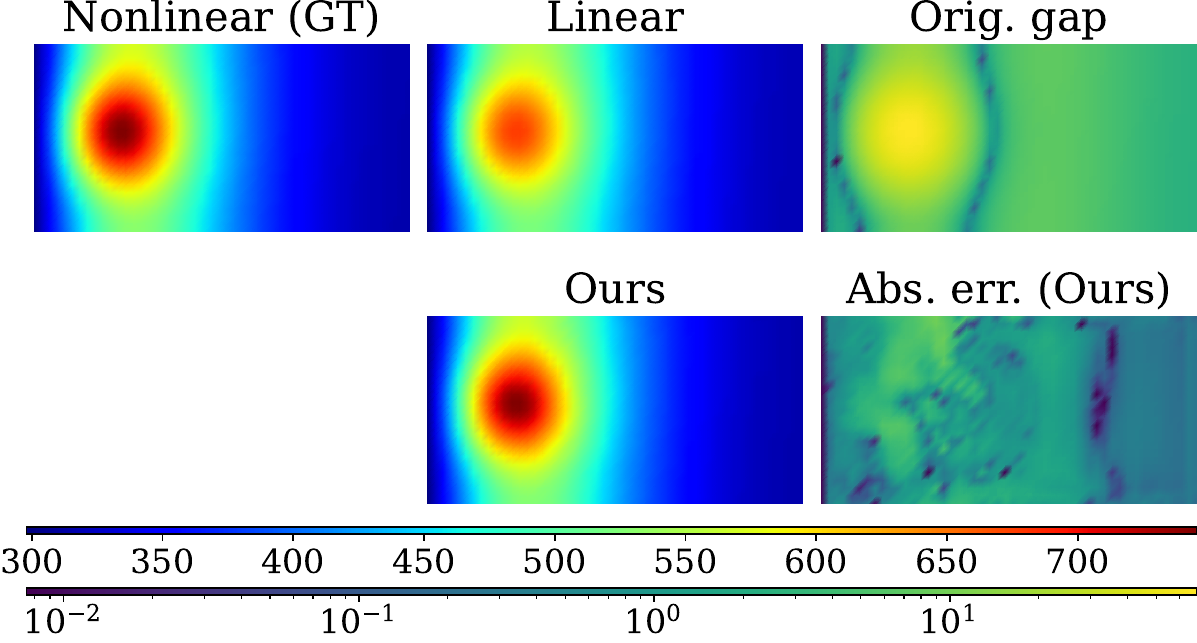}
     \caption{}
     \label{fig:gauss_d8}
 \end{subfigure}
 \medskip
 \begin{subfigure}{0.482\textwidth}
     \includegraphics[width=\textwidth]{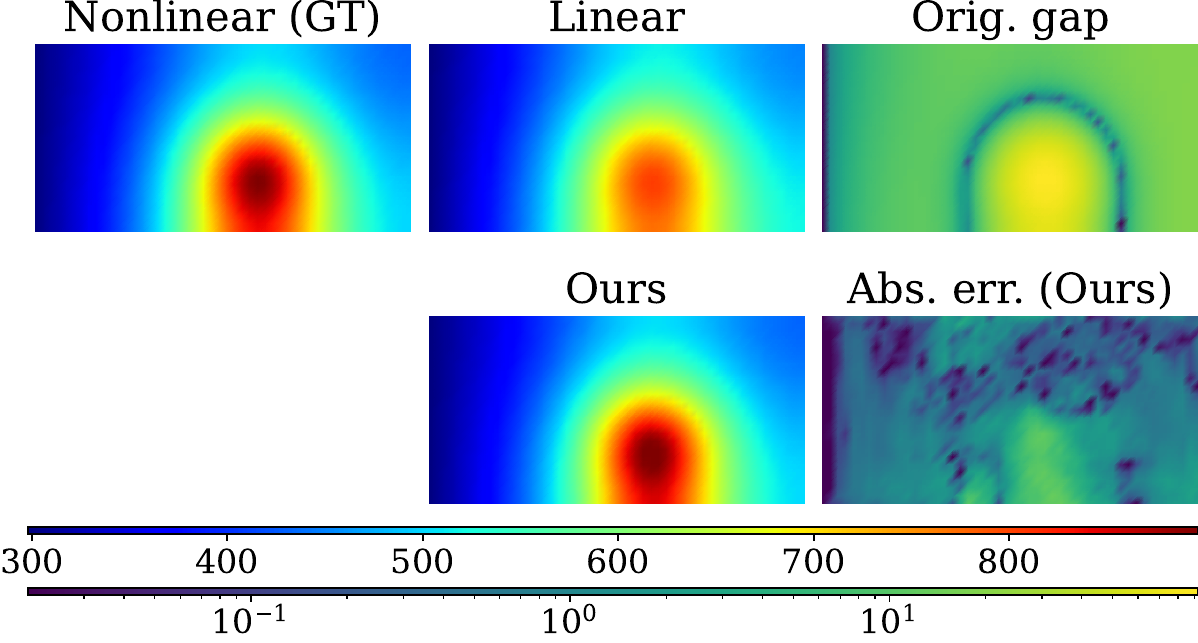}
     \caption{}
     \label{fig:gauss_d9}
 \end{subfigure}
\hspace{0.28cm}
 \begin{subfigure}{0.482\textwidth}
     \includegraphics[width=\textwidth]{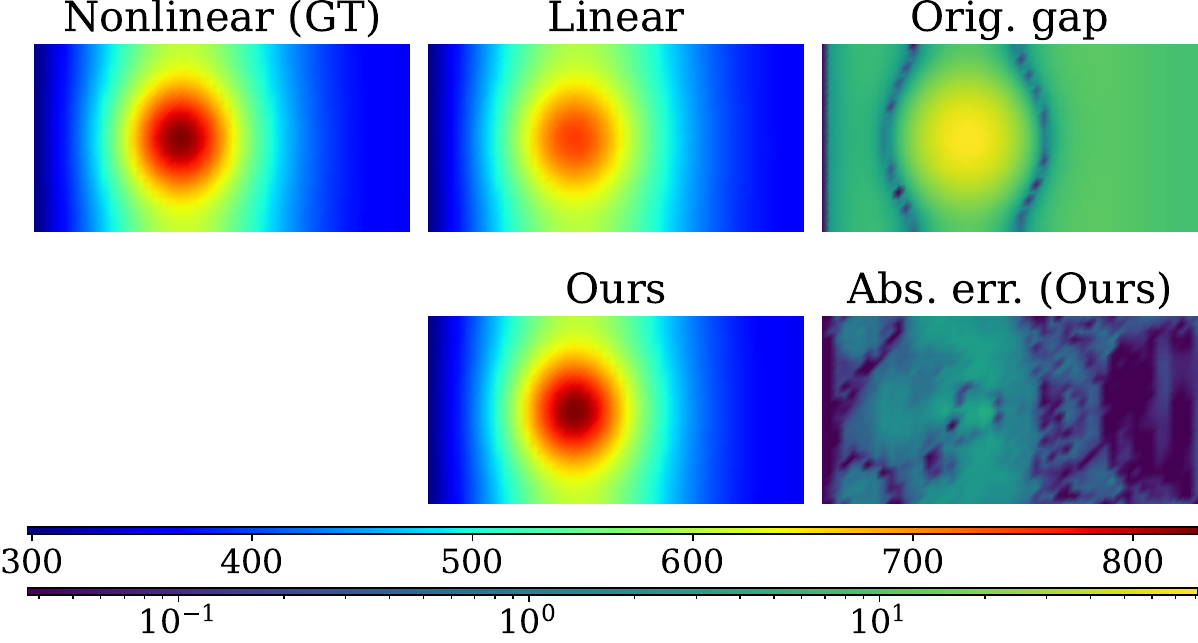}
     \caption{}
     \label{fig:gauss_d10}
 \end{subfigure}
 \caption{Predicted temperature field for data with Gaussian heat source load on the additional load positions, unseen during training.}
 \label{fig:gauss_unseen2}
\end{figure}

In this section, we demonstrate additional unseen designs for the training on the dataset B1 (Figure \ref{fig:gauss_unseen2}).

We can see that the maximum error in the node for each design does not exceed $10$ degrees.

\subsection{Noisy input}

The goal of this use case is to test the robustness of the model with regards to the possible measurement noise that usually is present in the real world problems. To do this we introduce a synthetic noise $\varepsilon$ in our dataset.

\begin{equation*}
    \varepsilon \sim \mathcal{N}(0, 10). 
\end{equation*}

The standard deviation is chosen to be equal to the RMSE loss during the training so that the  noise would not disrupt the temperature field pattern.

\begin{figure}[ht!]
 \begin{subfigure}{0.5\textwidth}
     \includegraphics[width=\textwidth]{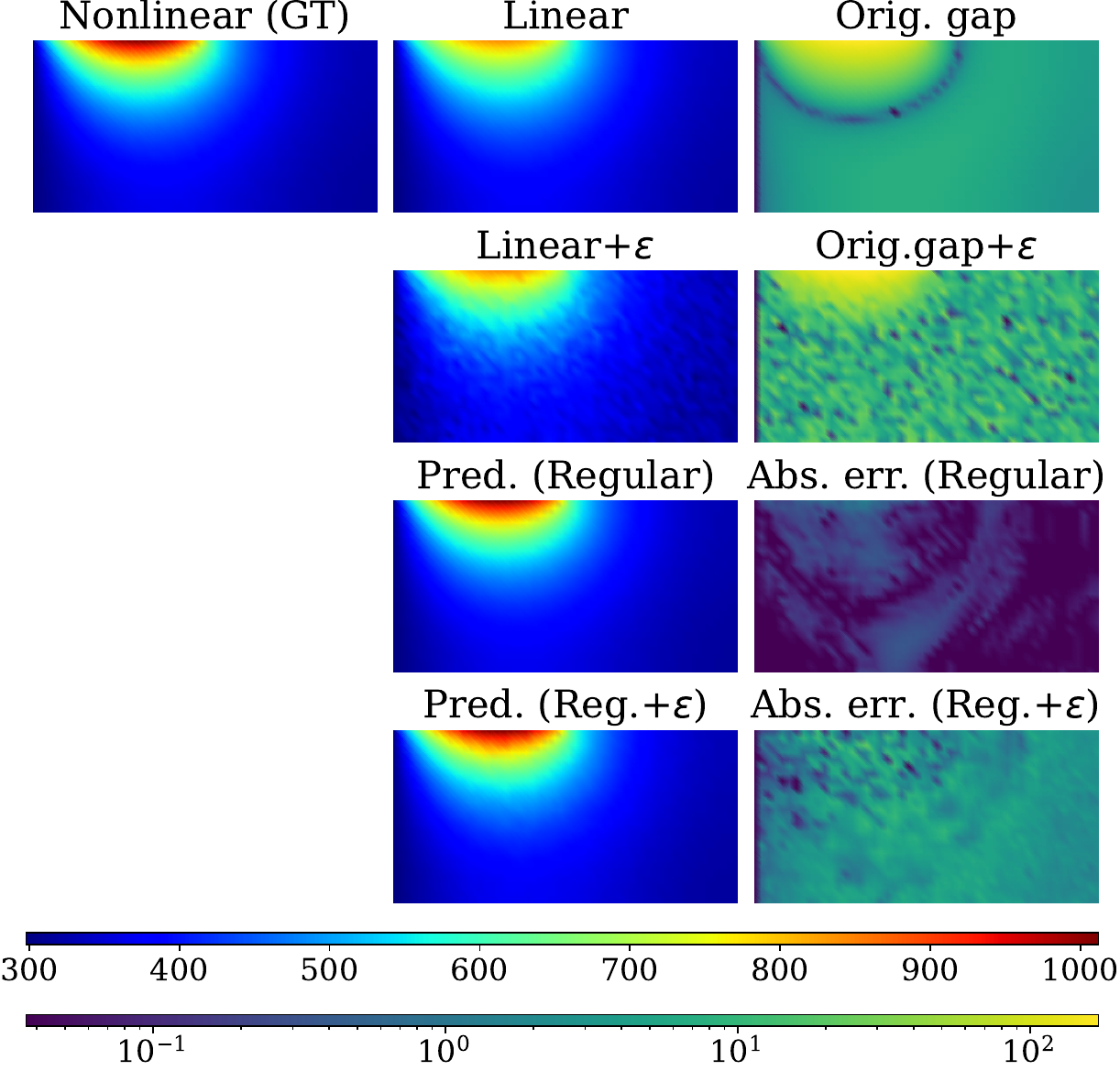}
     \caption{}
     \label{fig:half_noise}
 \end{subfigure}
 \hfill
 \begin{subfigure}{0.5\textwidth}
     \includegraphics[width=\textwidth]{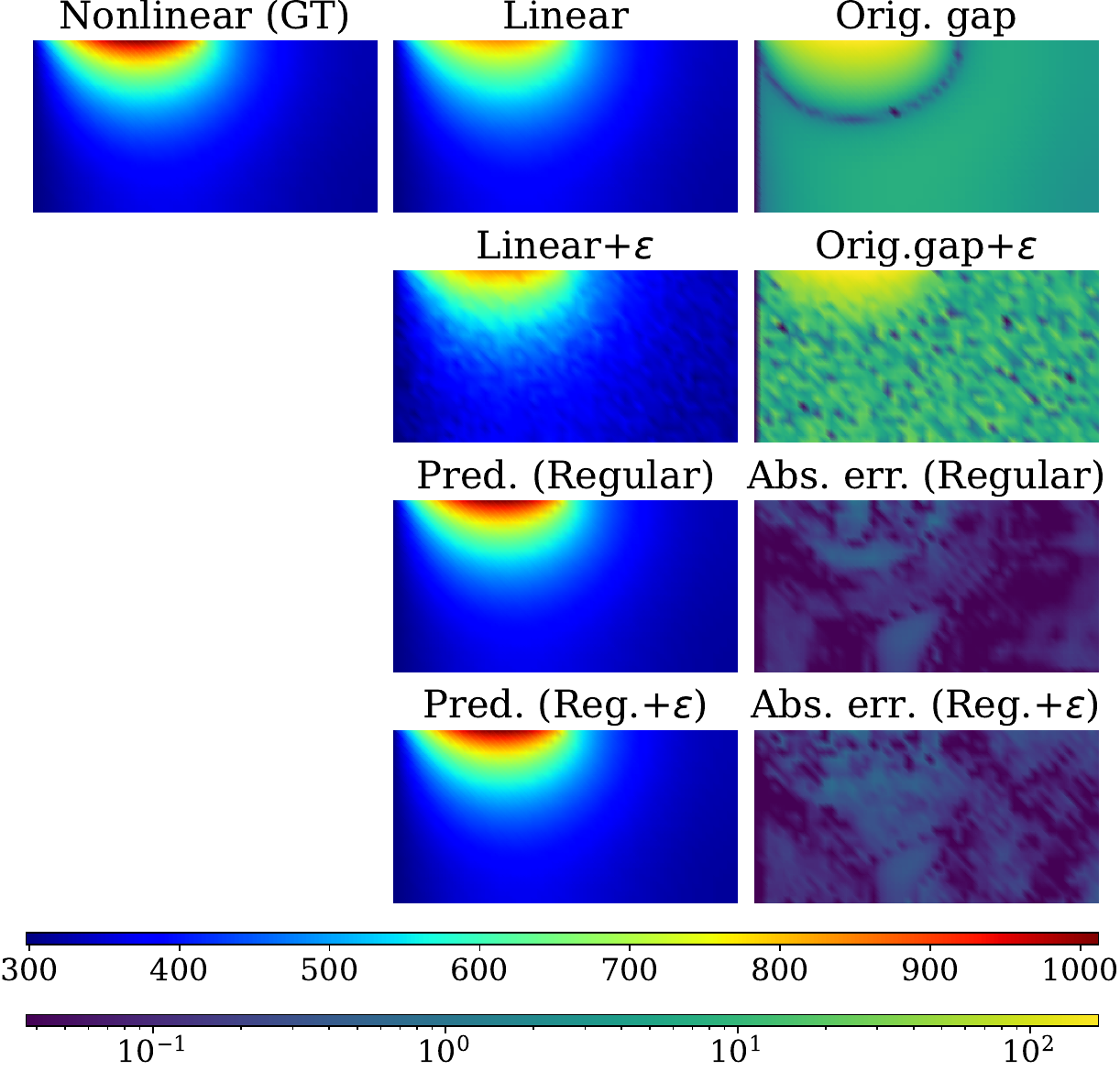}
     \caption{}
     \label{fig:half_noise_ni}
 \end{subfigure} 
 \medskip
 
  \begin{subfigure}{0.5\textwidth}
     \includegraphics[width=\textwidth]{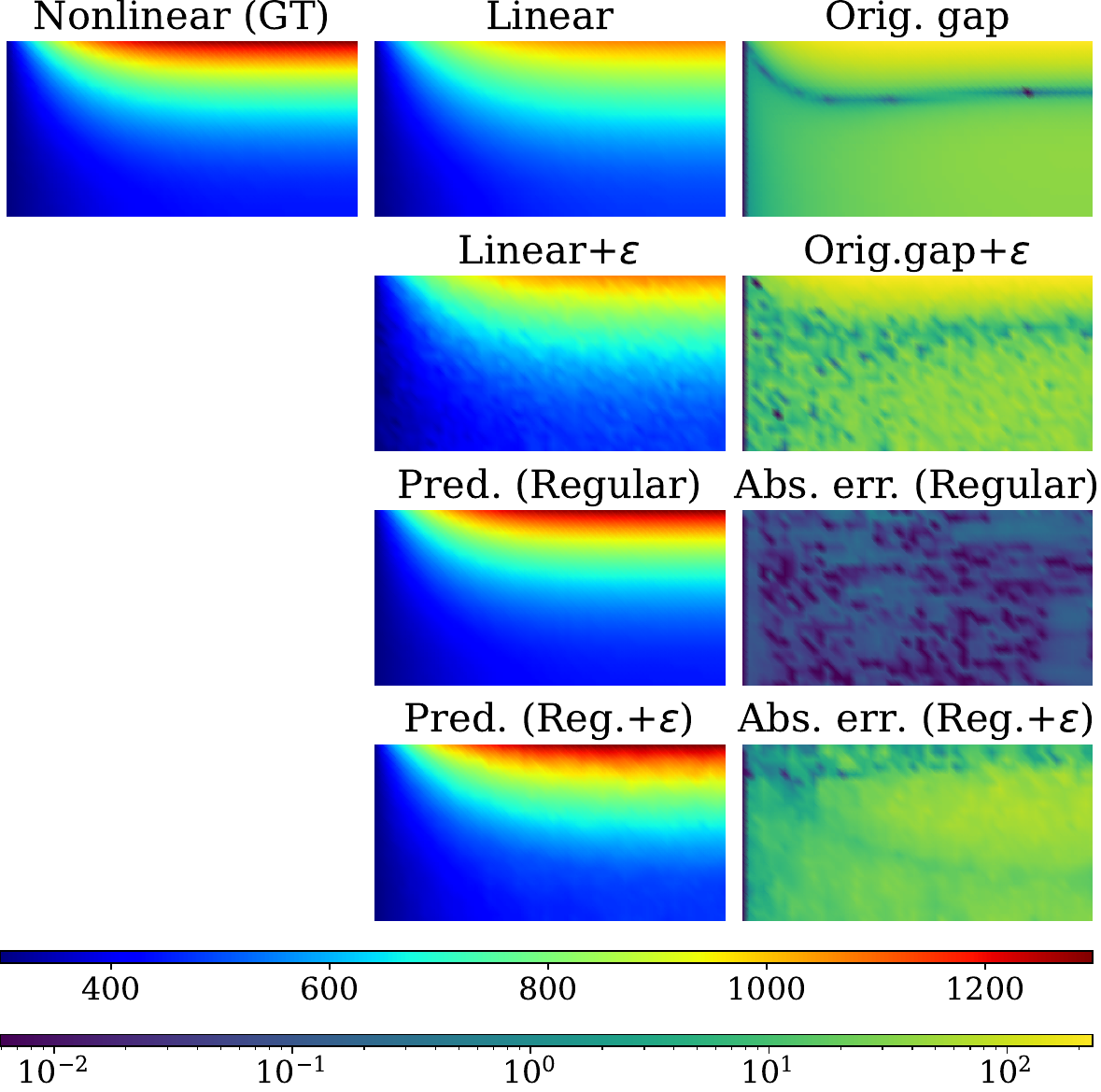}
     \caption{}
     \label{fig:all_noise}
 \end{subfigure}
 \hfill
 \begin{subfigure}{0.5\textwidth}
     \includegraphics[width=\textwidth]{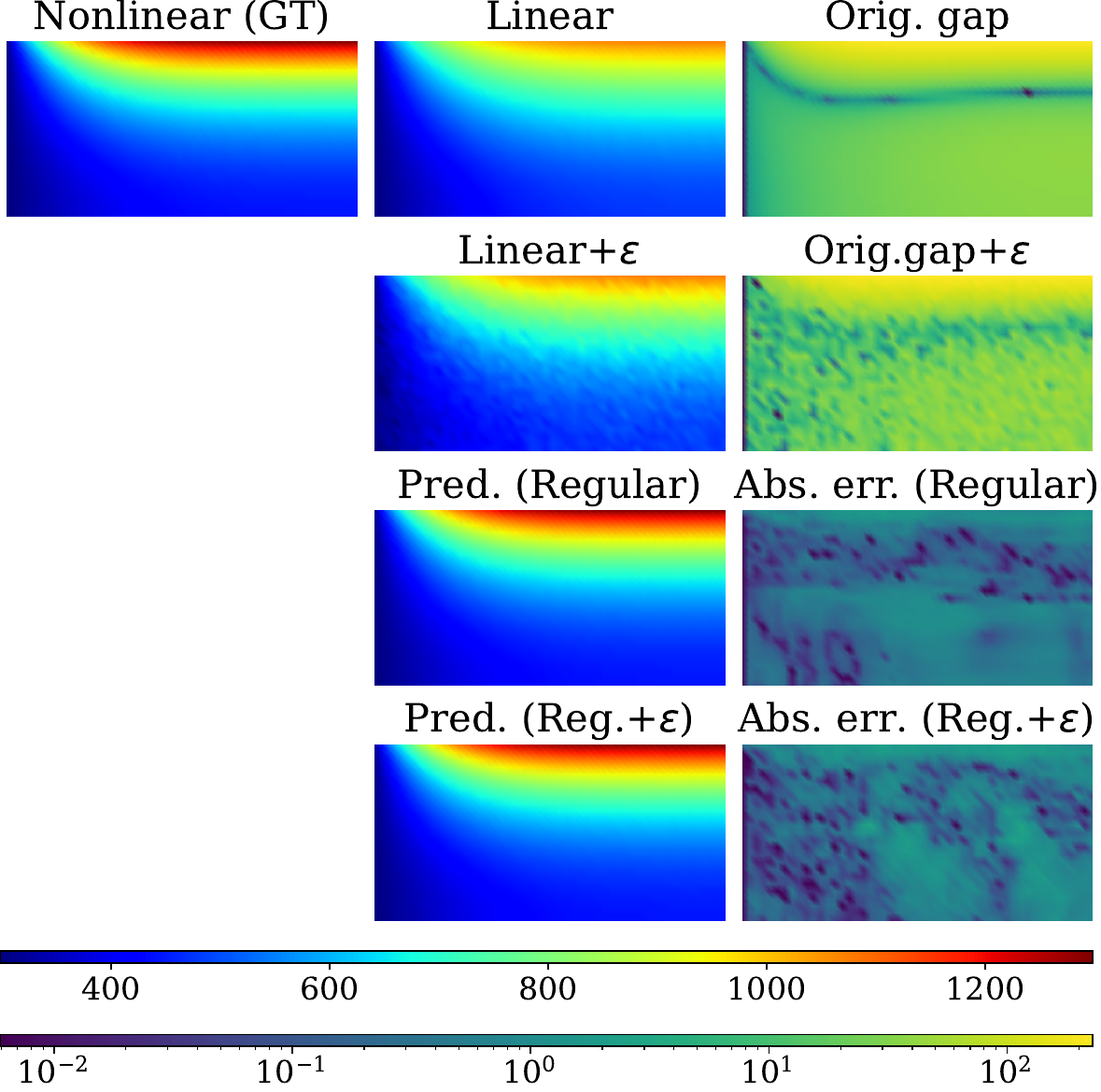}
     \caption{}
     \label{fig:all_noise_ni}
 \end{subfigure} 
 \caption{Comparison of the model  robustness to the  synthetic measurement noise in the model input in cases: (a) training on A1 dataset without data augmentation; (b) training on A1 dataset with noise augmentation; (c) training on A2 dataset without data augmentation; (d) training on A2 dataset with noise augmentation.}
 \label{fig:noisy_input}
\end{figure}

In Figure \ref{fig:noisy_input} we can see the comparison between models trained on the datasets A1 and A2 with 10\% of samples. One set of runs was performed on the original data, and another one on the data with noise augmentation. 

As we can see in Figures \ref{fig:half_noise} and \ref{fig:all_noise}, the model trained without any data augmentation exhibits higher prediction error on noisy samples. However, we can still see the lower error compared to the original one between linear and nonlinear simulations, especially in the regions with higher temperature.

In Figures \ref{fig:half_noise_ni} and \ref{fig:all_noise_ni} we can see that in the case of noise augmentation during training the performance of the model does not change in case of a noisy sample. 

\begin{table}[ht!]
\begin{tabular}{lp{2cm}ll}
\toprule
Augmentation & Inference\linebreak  dataset & MAE, $\text{K}$ & MAPE, \%  \\
\midrule
\multirow{2}{*}{---} & A1 & $(4.86 \pm 0.68) \cdot 10^{-2}$ & $(1.24 \pm 0.23) \cdot 10^{-2}$  \\
 & A1 (noisy) & $1.87 \pm 0.07$ & $0.51 \pm 0.02$  \\
\midrule
\multirow{2}{*}{Noise} & A1 & $(7.27 \pm 0.66) \cdot 10^{-2}$ & $(1.80 \pm 0.14) \cdot 10^{-2}$ \\
 & A1 (noisy) & $(9.94 \pm 0.38) \cdot 10^{-2}$ & $(25.66 \pm 0.75) \cdot 10^{-3}$ \\
\midrule
\multirow{2}{*}{---} & A2 & $(10.34 \pm 0.76) \cdot 10^{-2}$ & $(22.93 \pm 0.86) \cdot 10^{-3}$ \\
 & A2 (noisy) & $7.37 \pm 2.30$ & $1.51 \pm 0.51$ \\
\midrule
\multirow{2}{*}{Noise} & A2 & $0.23 \pm 0.02$ & $(4.87 \pm 0.55) \cdot 10^{-2}$ \\
 & A2 (noisy) & $0.33 \pm 0.03$ & $(7.21 \pm 0.76) \cdot 10^{-2}$ \\
\bottomrule
\end{tabular}
\caption{Results of the models trained on 10\% of data with and without noise augmentation, applied to the regular and noisy data.}
\label{tab:noisy_input}
\end{table}

In Table \ref{tab:noisy_input} we can see the same performance on average. Models trained with noise augmentation perform better on noisy samples.

In conclusion, we can say that the error on the noisy frames does not increase to the point of being equal or higher then the one between linear and nonlinear simulations.

\subsection{Summary table}\label{appendix:summary_table}

In this section, we include the summary table with the results for all use cases (Table \ref{tab:summary}).

\begin{sidewaystable}[ht!]
\centering
\begin{tabular}{lllll}
\toprule
Model & Dataset & MAE, $\text{K}$ & MAPE, \% & RMSE, $\times 10^{-3}$ \\
\midrule
Hybrid twin & \multirow{2}{*}{A1 (50\%)} & $\mathbf{(2.69 \pm 0.25) \cdot 10^{-2}}$ & $\mathbf{(6.83 \pm 0.33) \cdot 10^{-3}}$ & $(0.29 \pm 0.17) \cdot 10^{-4}$ \\
MGN+NI &  & $33.42 \pm 28.60$ & $6.89 \pm 5.87$ & $61.21 \pm 53.18$ \\
\midrule
Hybrid twin & \multirow{2}{*}{A2 (50\%)} & $\mathbf{(7.29 \pm 1.33) \cdot 10^{-2}}$ & $\mathbf{(1.58 \pm 0.23) \cdot 10^{-2}}$ & $\mathbf{(0.54 \pm 0.18) \cdot 10^{-4}}$ \\
MGN+NI &  & $0.42 \pm 0.12$ & $(9.19 \pm 2.63) \cdot 10^{-2}$ & $(1.64 \pm 0.91) \cdot 10^{-3}$ \\
\midrule
Hybrid twin & \multirow{2}{*}{A1 (10\%)} & $\mathbf{(4.86 \pm 0.68) \cdot 10^{-2}}$ & $\mathbf{(1.24 \pm 0.23) \cdot 10^{-2}}$ & $\mathbf{(0.59 \pm 0.42) \cdot 10^{-4}}$ \\
MGN+NI &  & $19.19 \pm 31.63$ & $4.64 \pm 7.66$ & $18.17 \pm 31.44$ \\
\midrule
Hybrid twin & \multirow{2}{*}{A2 (10\%)} & $\mathbf{(10.34 \pm 0.76) \cdot 10^{-2}}$ & $\mathbf{(22.93 \pm 0.86) \cdot 10^{-3}}$ & $\mathbf{(0.29 \pm 0.21) \cdot 10^{-3}}$ \\
MGN+NI &  & $1.01 \pm 0.15$ & $0.20 \pm 0.02$ & $(0.91 \pm 0.23) \cdot 10^{-2}$ \\
\midrule
Hybrid twin & A3 & $(72.37 \pm 0.80) \cdot 10^{-2}$ & $(15.16 \pm 0.12) \cdot 10^{-2}$ & $(10.11 \pm 0.79) \cdot 10^{-3}$ \\
Hybrid twin & A4 & $7.31 \pm 1.25$ & $1.25 \pm 0.27$ & $0.40 \pm 0.11$ \\
\midrule
Hybrid twin & A5 & $2.13 \pm 0.93$ & $0.37 \pm 0.19$ & $(4.61 \pm 3.54) \cdot 10^{-2}$ \\
Hybrid twin & A6 & $15.90 \pm 1.06$ & $1.96 \pm 0.19$ & $1.19 \pm 0.16$ \\
\midrule
Hybrid twin & B1 & $0.59 \pm 0.05$ & $0.12 \pm 0.01$ & $(6.50 \pm 0.52) \cdot 10^{-3}$ \\
\midrule
Hybrid twin & B2 & $9.62 \pm 1.54$ & $0.99 \pm 0.15$ & $0.33 \pm 0.08$ \\
\bottomrule
\end{tabular}
\caption{Results for all the use cases}
\label{tab:summary}
\end{sidewaystable}

\section{Train and validation curves}\label{appendix:train_val}

Here we display train and validation curves for all the models mentioned in the paper. For each model we compute mean and standard deviation over 3 runs with different seeds.

\begin{figure}[ht!]
 \begin{subfigure}{0.3\textwidth}
     \includegraphics[width=\textwidth]{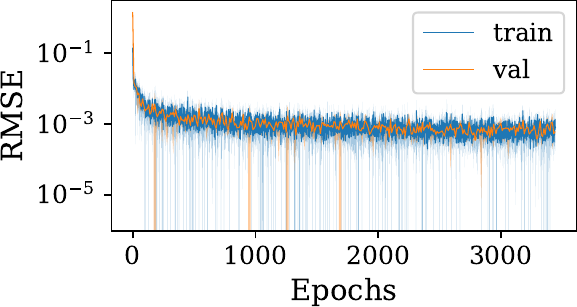}
     \caption{}
     \label{fig:v1}
 \end{subfigure}
\hspace{0.28cm}
\vspace{0.2cm}
 \begin{subfigure}{0.3\textwidth}
     \includegraphics[width=\textwidth]{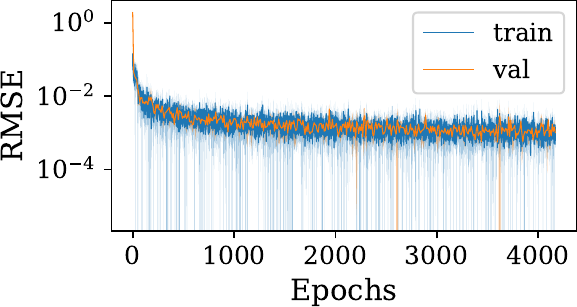}
     \caption{}
     \label{fig:v2}
 \end{subfigure}
\hspace{0.28cm}
\vspace{0.2cm}
  \begin{subfigure}{0.3\textwidth}
     \includegraphics[width=\textwidth]{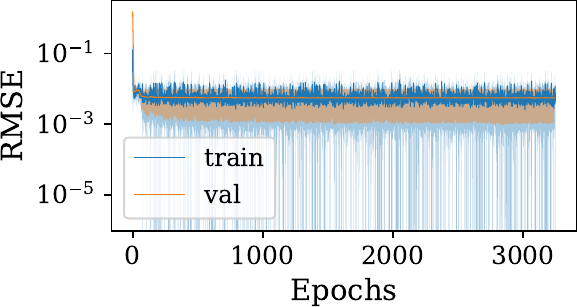}
     \caption{}
     \label{fig:v3}
 \end{subfigure}
 \medskip
 \begin{subfigure}{0.3\textwidth}
     \includegraphics[width=\textwidth]{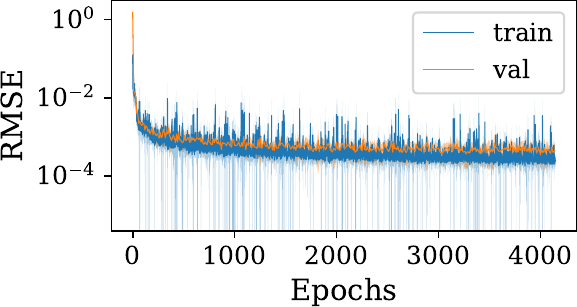}
     \caption{}
     \label{fig:v4}
 \end{subfigure}
\hspace{0.28cm}
\vspace{0.2cm}
  \begin{subfigure}{0.3\textwidth}
     \includegraphics[width=\textwidth]{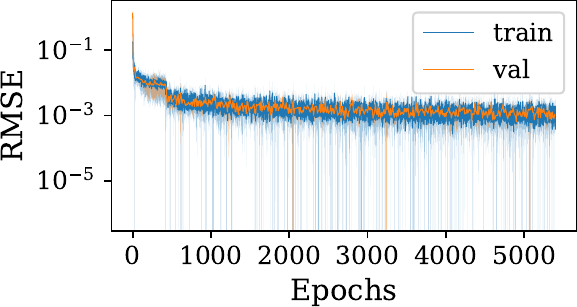}
     \caption{}
     \label{fig:v5}
 \end{subfigure}
\hspace{0.28cm}
\vspace{0.2cm}
 \begin{subfigure}{0.3\textwidth}
     \includegraphics[width=\textwidth]{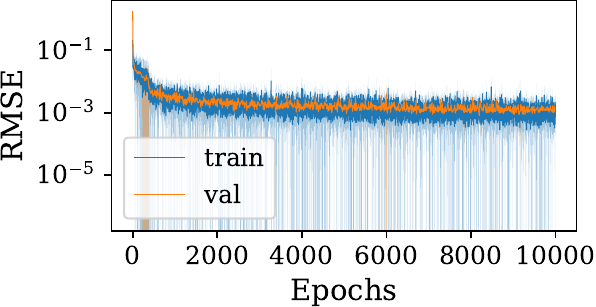}
     \caption{}
     \label{fig:v6}
 \end{subfigure}
 \medskip
   \begin{subfigure}{0.3\textwidth}
     \includegraphics[width=\textwidth]{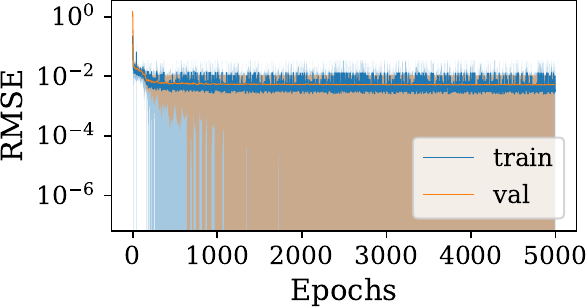}
     \caption{}
     \label{fig:base_ni_10_tr_v}
 \end{subfigure}
\hspace{0.28cm}
\vspace{0.2cm}
 \begin{subfigure}{0.3\textwidth}
     \includegraphics[width=\textwidth]{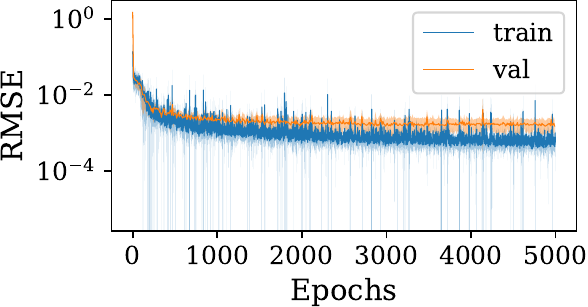}
     \caption{}
     \label{fig:v7}
 \end{subfigure}
\hspace{0.28cm}
\vspace{0.2cm}
   \begin{subfigure}{0.3\textwidth}
     \includegraphics[width=\textwidth]{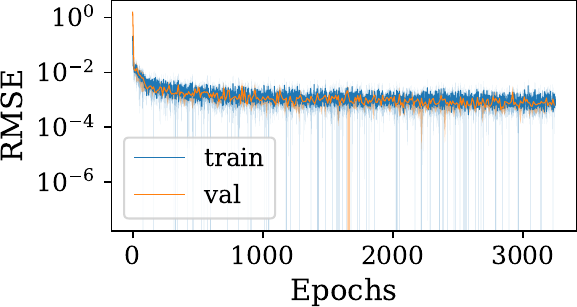}
     \caption{}
     \label{fig:v8}
 \end{subfigure}
 \medskip
 \begin{subfigure}{0.3\textwidth}
     \includegraphics[width=\textwidth]{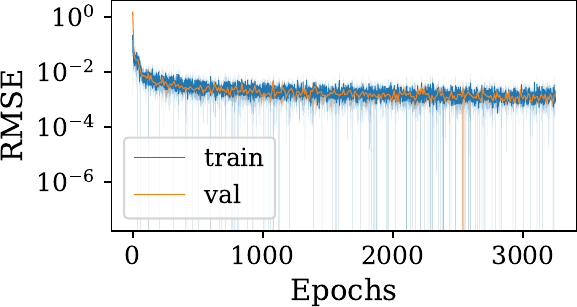}
     \caption{}
     \label{fig:v9}
 \end{subfigure}
\hspace{0.28cm}
\vspace{0.2cm}
   \begin{subfigure}{0.3\textwidth}
     \includegraphics[width=\textwidth]{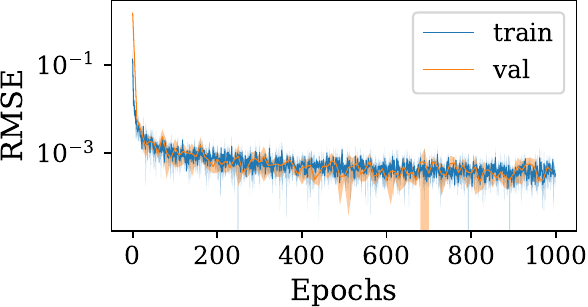}
     \caption{}
     \label{fig:v10}
 \end{subfigure}
\hspace{0.28cm}
\vspace{0.2cm}
 \begin{subfigure}{0.3\textwidth}
     \includegraphics[width=\textwidth]{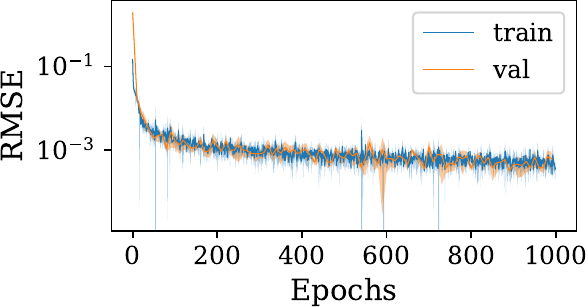}
     \caption{}
     \label{fig:v11}
 \end{subfigure}
 \medskip
   \begin{subfigure}{0.3\textwidth}
     \includegraphics[width=\textwidth]{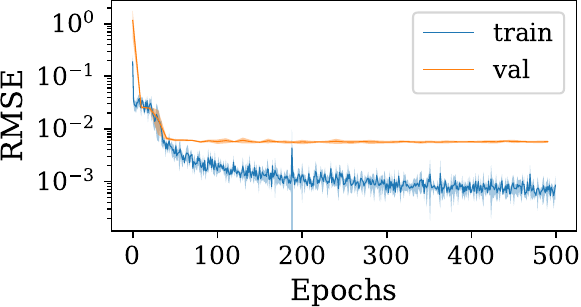}
     \caption{}
     \label{fig:b1}
 \end{subfigure}
\hspace{0.28cm}
\vspace{0.2cm}
 \begin{subfigure}{0.3\textwidth}
     \includegraphics[width=\textwidth]{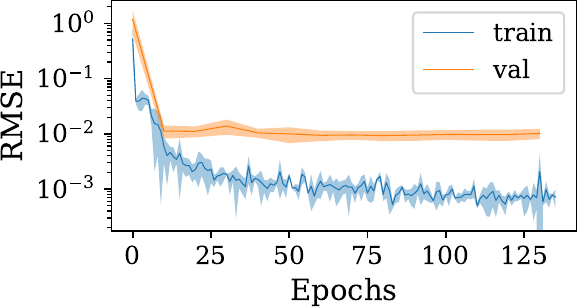}
     \caption{}
     \label{fig:b2}
 \end{subfigure} 
\caption{Train and validation curves given in log scale for all the models used in this work: (a) Hybrid twin, A1 (50\%); (b) Hybrid twin, A2 (50\%);  (c) MeshGraphNet+NI, A1 (50\%);  (d) MeshGraphNet+NI, A2 (50\%); (e) Hybrid twin, A1 (10\%); (f) Hybrid twin, A2 (10\%);  (g) MeshGraphNet+NI, A1 (10\%);  (h) MeshGraphNet+NI, A2 (10\%); (i) Hybrid twin (sum aggregation), A1 (50\%); (j) Hybrid twin (sum aggregation), A2 (50\%);  (k) Hybrid twin, A7;  (l) Hybrid twin, A8; (m) Hybrid twin, B1; (n) Hybrid twin, B2.
}
 \label{fig:train_val}
\end{figure}

In Figure \ref{fig:train_val} we can see that for every model our training converges. For models in Figures \ref{fig:b1} and \ref{fig:b2} we can notice a slight overfitting. These models were trained on datasets B1 and B2 that contain simulations of different designs. The overfitting here is an indicator of an insufficient number of designs in the training dataset. In Figure \ref{fig:base_ni_10_tr_v} we can see that the standard deviation is higher than in other cases because one out of 3 runs had a different training trajectory.

\end{document}